\documentclass[13pt]{article}
\usepackage{latexsym}
\usepackage{geometry}
\usepackage{graphicx}
\usepackage{amsmath, amssymb, amsthm}
\usepackage{booktabs}
\usepackage{algorithm}
\usepackage{algorithmic}

\usepackage{url}
\usepackage{natbib}
\usepackage{appendix}
\usepackage{amsthm}
\usepackage{amsmath}

\usepackage{amsfonts}
\usepackage{multirow}
\usepackage{multicol}

\usepackage{color}
\usepackage{algorithm}
\usepackage{algorithmic}

\usepackage{xcolor,colortbl}
\definecolor{LightCyan}{rgb}{0.88,1,1}

\usepackage{graphicx}
\usepackage{subfigure}
\usepackage{hyperref}
\usepackage{tcolorbox}

\newtheorem{theorem}{Theorem}
\newtheorem{lemma}{Lemma}

\newtheorem{assumption}{Assumption}
\newtheorem{remark}{Remark}

\begin{document}
\title{ Fast Adaptive Federated Bilevel Optimization }
\author{Feihu Huang\thanks{College of Computer Science and Technique, College of Artificial Intelligence, Nanjing University of Aeronautics and Astronautics, Nanjing, China. Email: huangfeihu2018@gmail.com} }

\date{}
\maketitle

\begin{abstract}
Bilevel optimization is a popular hierarchical model in machine learning, and
 has been widely applied to many machine learning tasks such as meta learning,
hyperparameter learning and policy optimization. Although many bilevel
optimization algorithms recently have been developed, few adaptive algorithm focuses on
the bilevel optimization under the distributed setting. It is well known that the adaptive gradient methods show superior performances on both distributed and non-distributed optimization.
In the paper, thus, we propose a novel adaptive federated bilevel optimization algorithm (i.e.,AdaFBiO) to solve the distributed bilevel optimization problems,
where the objective function of Upper-Level (UL) problem is possibly nonconvex, and that of Lower-Level (LL) problem is strongly convex.
Specifically, our AdaFBiO algorithm builds on the momentum-based variance reduced technique and local-SGD
to obtain the best known sample and communication complexities simultaneously.
In particular, our AdaFBiO algorithm uses the unified adaptive matrices to flexibly incorporate various adaptive learning rates to
update variables in both UL and LL problems.
Moreover, we provide a convergence analysis framework for our AdaFBiO algorithm, and prove it needs the sample complexity of $\tilde{O}(\epsilon^{-3})$
with communication complexity of $\tilde{O}(\epsilon^{-2})$ to obtain an $\epsilon$-stationary point.
 Experimental results on federated hyper-representation learning and federated data hyper-cleaning tasks verify efficiency of our algorithm.
\end{abstract}

\section{Introduction}
Federated Learning (FL)~\citep{mcmahan2017communication} is a popular distributed learning paradigm in machine learning for training a centralized model
based on decentralized data over a network of clients. Although many FL algorithms recently have been developed,
few FL algorithms focus on the bilevel optimization problems with hierarchical structures.
In the paper, we consider studying the following distributed stochastic bilevel optimization problem, defined as
\begin{align} \label{eq:1}
 \min_{x \in \mathbb{R}^d} & \ F(x):=\frac{1}{M}\sum_{m=1}^M\mathbb{E}_{\xi^m \sim \mathcal{D}^m}\big[f^m(x,y^*(x);\xi^m)\big]  & \mbox{(Upper-Level)}  \\
 \mbox{s.t.} & \ y^*(x) \in \arg\min_{y\in \mathbb{R}^p} \ \frac{1}{M}\sum_{m=1}^M\mathbb{E}_{\zeta^m \sim \mathcal{S}^m}\big[g^m(x,y;\zeta^m)\big], &  \mbox{(Lower-Level)} \nonumber
\end{align}
where for any $m\in[M]$, $f^m(x,y^*(x))=\mathbb{E}_{\xi^m}\big[f^m(x,y^*(x);\xi^m)\big]$ denotes the objective function of Upper-Level (UL) problem in
$m$-th client, which is a differentiable and possibly nonconvex function,
and $g^m(x,y)=\mathbb{E}_{\zeta^m}\big[g^m(x,y;\zeta^m)\big]$ denotes the objective function of Lower-Level (LL) problem in
$m$-th client, which is a differentiable and strongly convex
function. Here $\xi^m$ and $\zeta^m$ for any $m\in[M]$ are independent random variables follow unknown distributions $\mathcal{D}^m$
and $\mathcal{S}^m$ respectively, and for any $m,j\in [M]$ possibly $\mathcal{D}^m \neq \mathcal{D}^j$,
$\mathcal{S}^m \neq \mathcal{S}^j$  and $\mathcal{D}^m\neq \mathcal{S}^j$. Note that let $y^m\in \mathbb{R}^{p_m}$ for any $m\in [M]$,
$p=\sum_{m=1}^Mp_m$, $y=(y^1;\cdots;y^M)\in \mathbb{R}^p$,
$y^*=(y^{*(1)};\cdots;y^{*(M)})\in \mathbb{R}^p$,
$g^m(x,y;\zeta^m) = g^m(x,y^m;\zeta^m)$ and $f^m(x,y^*(x);\xi^m)=f^m(x,y^{*(m)}(x);\xi^m)$, the above problem \eqref{eq:1} will become the following
formulation studied in~\citep{guo2021randomized,li2022local},
\begin{align} \label{eq:2}
 \min_{x \in \mathbb{R}^d} & \ F(x)=\frac{1}{M}\sum_{m=1}^M\mathbb{E}_{\xi^m \sim \mathcal{D}^m}\big[f^m(x,y^{*(m)}(x);\xi^m)\big]    \\
 \mbox{s.t.} & \ y^{*(m)}(x) \in \arg\min_{y^m\in \mathbb{R}^{p_m}} \ \mathbb{E}_{\zeta^m \sim \mathcal{S}^m}\big[g^m(x,y^m;\zeta^m)\big], \ \forall m \in [M]. \nonumber
\end{align}
Thus, Problem \eqref{eq:2} is a specific case of Problem \eqref{eq:1}. Note that \cite{guo2021randomized} did not focus on the \emph{distributed} bilevel optimization.
Applications of the problem \eqref{eq:1} involves many machine learning problems with a hierarchical structure, which include meta learning \cite{franceschi2018bilevel},
hyper-parameter optimization \cite{franceschi2018bilevel}, reinforcement learning \cite{hong2020two} and neural network architecture search \cite{liu2018darts}.
Specifically, we provide two popular applications
that can be formulated as the bilevel problem \eqref{eq:1}.

\begin{table*}
  \centering
  \caption{ \textbf{Sample} and \textbf{Communication} complexities comparison of the representative \textbf{federated bilevel optimization} algorithms to find
  an $\epsilon$-stationary point of the bilevel problem \eqref{eq:1}, i.e., $\mathbb{E}\|\nabla F(x)\|\leq \epsilon$
  or its equivalent variants. \textbf{ALR} denotes adaptive learning rate.  }
  \label{tab:1}
   \resizebox{\textwidth}{!}{
\begin{tabular}{c|c|c|c|c}
  \hline
   \textbf{Algorithm} & \textbf{Reference} & \textbf{Sample Complexity} & \textbf{Communication Complexity}& \textbf{ALR}   \\ \hline
  FEDNEST  & \cite{tarzanagh2022fednest}  & $\tilde{O}(\epsilon^{-4})$ & $\tilde{O}(\epsilon^{-4})$ &\\  \hline
  FedBiOAcc  & \cite{li2022local} & $\tilde{O}(\epsilon^{-3})$ & $\tilde{O}(\epsilon^{-2})$ &  \\  \hline
  LocalBSGVRM  & \cite{gao2022convergence}  & $\tilde{O}(\epsilon^{-3})$ &$\tilde{O}(\epsilon^{-2})$ &  \\  \hline
  AdaFBiO  & Ours & $\tilde{O}(\epsilon^{-3})$ & $\tilde{O}(\epsilon^{-2})$ & $\checkmark$ \\  \hline
\end{tabular}
 }
\end{table*}

\textbf{1). Distributed Meta Learning.} The meta learning \citep{finn2017model,franceschi2018bilevel} aims to learn task specific parameters that generalize
to a diverse set of tasks. Suppose we have $M$ tasks $\{\mathcal{T}_m, \ m\in [M]\}$ and each task has a corresponding
loss function $L(x,y^m;\xi^m)$ with $\xi^m$ representing data sample for task $\mathcal{T}_m$, where $y^m\in \mathbb{R}^{p_m}$ denotes
the task specific parameters, and $x\in \mathbb{R}^{d}$ denotes the model parameters shared
among tasks. In the paper, we consider the mate learning under the distributed setting, where
each task is in each client. Then the goal of meta learning can be represented to solve the following bilevel optimization problem:
\begin{align} \label{eq:3}
 & \min_{x\in \mathbb{R}^d} \frac{1}{M} \sum_{m=1}^M \mathbb{E}_{\xi^m \sim \mathcal{D}_{te}^m}\Big[ L(x,y^{*(m)}(x);\xi^m)\Big]  \\
 & \mbox{s.t.} \ y^*(x)\in \arg\min_{y \in \mathbb{R}^p}\frac{1}{M} \sum_{m=1}^M \Big\{\mathbb{E}_{\zeta^m \sim \mathcal{D}_{tr}^m}\Big[ L(x,y^m;\zeta^m)\Big] + \mathcal{R}(y^m)\Big\}, \nonumber
\end{align}
where $y^m\in \mathbb{R}^{p_m}$ for any $m\in [M]$, $p=\sum_{m=1}^Mp_m$, $y=(y^1;\cdots;y^M)\in \mathbb{R}^p$, $y^*=(y^{*(1)};\cdots;y^{*(M)})\in \mathbb{R}^p$ and
$\mathcal{R}(y^m)$ is a strongly convex regularize for all $m\in [M]$. Here
$\mathcal{D}_{tr}^m$ and $\mathcal{D}_{te}^m$ are the training and testing datasets respectively for task $\mathcal{T}_m$ for all $m\in [M]$.
Let $f^m(x,y^*(x);\xi^m) = L(x,y^{*(m)}(x);\xi^m)$ and $g^m(x,y;\zeta^m) = L(x,y^m;\zeta^m) + R(y^m)$, so the above formulation \eqref{eq:3}
can be represented by the formulation of Problem \eqref{eq:1}.

\textbf{2).  Federated Data Hyper-Cleaning.}
 The data hyper-cleaning \citep{shaban2019truncated} is a hyperparameter optimization problem that
aims to train a classifier model with a dataset of randomly corrupted label.
In the paper, we design a robust federated learning framework by using data hyper-cleaning technique, which
is to solve the following distributed bilevel problem:
\begin{align} \label{eq:4}
 & \min_{x \in \mathbb{R}^d} \frac{1}{M} \sum_{m=1}^M \Big\{ \sum_{i \in \mathcal{D}_{te}^m}L\big(a_i^T y^*(x),b_i\big) \Big\}  \\
 & \mbox{s.t.} \ y^*(x) \in \arg\min_{y\in \mathbb{R}^p} \frac{1}{M} \sum_{m=1}^M\Big\{ \sum_{i \in \mathcal{D}_{tr}^m} \sigma(x_i)L(a_i^T y,b_i) + \nu\|y\|^2 \Big\}, \nonumber
\end{align}
where $\nu>0$, $L(\cdot,\cdot)$ is the loss function, $(a_i,b_i)$ denotes the $i$-th data point, $y\in \mathbb{R}^p$ is the model parameter,
and $x_i$ is the parameter that determines the weight for the $i$-th data.
Here $\sigma: \mathbb{R}\rightarrow \mathbb{R}^+$ denotes the weight function.
$\mathcal{D}^m_{tr}$ and $\mathcal{D}^m_{te}$ are validation and training sets on the $m$-th client, respectively.

Although many algorithms recently have been proposed to solve the bilevel optimization problems, few distributed algorithms focus on
 solving the distributed bilevel optimization problems. More recently, \cite{li2022local,tarzanagh2022fednest,gao2022convergence} proposed some federated bilevel algorithms for
 the distributed bilevel problems.
 Meanwhile, \cite{chen2022decentralized,lu2022decentralized} studied the bilevel optimization algorithms under the decentralized setting.
 It is well known that the adaptive gradient methods show superior
performance over the non-adaptive learning rate schedule, however, how to incorporate
these adaptive gradient methods efficiently to the federated setting is still an open problem.
Recently, only a few adaptive federated learning algorithms have been developed
for the single-level optimization problems.
Then there exists a natural question:
\begin{center}
\begin{tcolorbox}
\textbf{ Could we design an effective adaptive federated algorithm for solving the distributed bilevel
optimization problem \eqref{eq:1} ? }
\end{tcolorbox}
\end{center}

In the paper, we provide an affirmative answer to the above question and propose
an effective and
efficient adaptive federated bilevel optimization algorithm (i.e., AdaFBiO) to solve the problem \eqref{eq:1}.
Specifically, our AdaFBiO algorithm builds on the local-SGD \citep{stich2018local} and momentum-based variance reduced of STORM \citep{cutkosky2019momentum} techniques
to obtain the best known sample and communication complexities simultaneously.
Our main contributions are three-fold:
\begin{itemize}
\item[(1)] We propose an efficient adaptive federated bilevel optimization algorithm (i.e., AdaFBiO) for the distributed bilevel optimization
based on the local-SGD and momentum-based variance reduced techniques. In particular, our AdaFBiO algorithm uses the unified adaptive matrices to flexibly incorporate various adaptive learning rates to
update variables in both UL and LL problems.
\item[(2)] We provide a convergence analysis framework for our AdaFBiO algorithm under non-i.i.d. setting, and prove it reaches the best known sample complexity of $O(\epsilon^{-3})$
with communication complexity of $O(\epsilon^{-2})$ to obtain an $\epsilon$-stationary point (Please see Table \ref{tab:1}).
\item[(3)] Some experiments demonstrate the efficiency of our AdaFBiO algorithm on federated hyper-representation learning and federated data hyper-cleaning tasks.
\end{itemize}

\section{Related Works}
In this section, we overview some representative federated learning algorithms, bilevel optimization algorithms and adaptive gradient algorithms, respectively.

\subsection{ Federated Learning Algorithms }
Federated Learning (FL) is a distributed and privacy-preserving optimization method, which
learns a global model from a set of distributed located clients without sharing their private data under the coordination of a server.
 \cite{mcmahan2017communication} firstly proposed the FedAvg algorithm for FL based on local-SGD algorithms~\citep{stich2018local},
where each client conducts multiple steps of gradient descent with its local data
and then sends the learned parameter to the server for averaging.
Recently, local-SGD and FedAvg algorithms have been developed and studied in \citep{karimireddy2019error,deng2021local,li2019convergencefedvrg,glasgow2022sharp}.
For example, \cite{li2019convergencefedvrg} studied the convergence properties
of FedAvg or local-SGD algorithm under non-i.i.d setting.
More recently, \cite{glasgow2022sharp} provided a near optimal convergence rate
of the Local-SGD under the convex setting.
In parallel, many accelerated FL algorithms~\citep{yuan2020federated,karimireddy2020scaffold,karimireddy2020mime,chen2020fedcluster,khanduri2021stem} have been proposed to accelerate the vanilla local-SGD and FedAvg algorithms.
For example, \cite{karimireddy2020scaffold} proposed a stochastic controlled
averaging algorithm for FL by adopting the variance-reduced technique.
Meanwhile, \cite{karimireddy2020mime} presented a general framework for FL by using a
combination of control-variate and server-level optimizer state.
More recently, \cite{khanduri2021stem} proposed a momentum-based variance reduced algorithm for FL,
which reaches simultaneously near-optimal sample and communication complexities.

Recently, some adaptive gradient methods \citep{reddi2020adaptive,chen2020toward,li2022federated} are proposed for FL. For example, \cite{reddi2020adaptive} proposed a class of adaptive federated algorithms for FL
by using adaptive learning rates at the server side. Meanwhile, \cite{chen2020toward} proposed an efficient local-AMSGrad algorithm, where clients locally update variables by using
adaptive learning rates shared with all clients.
Subsequently, some communication-efficient adaptive federated algorithms~\citep{tang2020apmsqueeze,lu2022maximizing} have been developed based on the compression technique.

\subsection{ Bilevel Optimization Algorithms }
Bilevel optimization is widely applied in many machine learning problems with hierarchical structures such as meta learning and hyperparameter learning.
So its researches have become active in the machine learning community, and
some bilevel optimization methods recently have been proposed.
For example, one class of successful methods \cite{colson2007overview,kunapuli2008classification} are to reformulate the bilevel problem as a single-level problem by replacing
the inner problem by its optimality conditions. Another class of successful methods \cite{ghadimi2018approximation,hong2020two,ji2020bilevel,chen2021single,khanduri2021near,
li2021fully,guo2021randomized}
for bilevel optimization
are to iteratively approximate the (stochastic) gradient of the outer problem either
in forward or backward. Moreover, the non-asymptotic analysis of these
bilevel optimization methods has been recently studied.
For example, \cite{ghadimi2018approximation} firstly studied the sample complexity of $O(\epsilon^{-6})$ of the proposed double-loop algorithm
for the bilevel problem \eqref{eq:1}. Subsequently, \cite{ji2020bilevel} proposed an accelerated double-loop algorithm that
reaches the sample complexity of $O(\epsilon^{-4})$ relying on large batches.
At the same time, \cite{hong2020two} studied a single-loop algorithm that
reaches the sample complexity of $O(\epsilon^{-5})$ without relying on large batches.
Moreover, \cite{yang2021provably,khanduri2021near,guo2021randomized,huang2021biadam} proposed a class of accelerated methods by using variance-reduced technique,
which achieve the best known sample complexity of $O(\epsilon^{-3})$. More recently, \cite{huang2022enhanced} proposed a class of accelerated
bilevel optimization algorithms by using Bregman distances.

More recently, \cite{li2022local,tarzanagh2022fednest,chen2022decentralized,lu2022decentralized,gao2022convergence} studied the bilevel optimization under the distributed setting. Specifically, \cite{li2022local,tarzanagh2022fednest,gao2022convergence} studied the federated learning framework for bilevel optimization. \cite{chen2022decentralized,lu2022decentralized} studied the bilevel optimization under the decentralized learning setting.

\subsection{ Adaptive Gradient Algorithms }
Adaptive gradient methods are a class of successful optimization tools to solve large-scale machine learning problems. Recently, the adaptive gradient methods~\citep{duchi2011adaptive,kingma2014adam,loshchilov2018decoupled} have been widely studied in machine learning community.
Adam \citep{kingma2014adam} is one of popular adaptive gradient methods by using a coordinate-wise adaptive learning rate
and momentum technique to accelerate algorithm, which is the default optimization tool
 for training deep neural networks (DNNs).
Subsequently, some variants of Adam algorithm \citep{zaheer2018adaptive,reddi2019convergence,chen2019convergence,guo2021novel}
have been presented to obtain a convergence guarantee under the nonconvex setting.
Due to using coordinate-wise adaptive learning rate, Adam frequently shows a bad generalization performance in training DNNs.
To improve the generalization performance of Adam, recently some adaptive gradient methods such as AdamW \citep{loshchilov2018decoupled} and AdaBelief \citep{zhuang2020adabelief} have been proposed.
More recently, some accelerated adaptive gradient methods \citep{cutkosky2019momentum,huang2021super} have been proposed based on the variance-reduced techniques.

\section{Preliminaries}

\subsection{Notations}
$\mathcal{U}\{1,2,\cdots,K\}$ denotes a uniform distribution over a discrete set $\{1,2,\cdots,K\}$.
$[M]$ denotes the set $\{1,2,\cdots,M\}$.
$\|\cdot\|$ denotes the $\ell_2$ norm for vectors and spectral norm for matrices. 
$\nabla f(x)$ denotes the gradient of function $f(x)$. 
$\langle x,y\rangle$ denotes the inner product of two vectors $x$ and $y$. For vectors $x$ and $y$, $x^r \ (r>0)$ denotes the element-wise
power operation, $x/y$ denotes the element-wise division and $\max(x,y)$ denotes the element-wise maximum. $I_{d}$ denotes a $d$-dimensional identity matrix. Matrix $A\succ 0$ is a positive definite.
Given $f(x,y)$, $f(x,\cdot)$ denotes  function \emph{w.r.t.} the second variable with fixing $x$,
and $f(\cdot,y)$ denotes function \emph{w.r.t.} the first variable
with fixing $y$.
$a_m=O(b_m)$ denotes $a_m \leq c b_m$ for some constant $c>0$. The notation $\tilde{O}(\cdot)$ hides logarithmic terms.

\subsection{Assumptions}
In this subsection, we give some mild assumptions on the problem \eqref{eq:1}.
\begin{assumption} \label{ass:1}
For any $m\in[M]$, $x$ and $\zeta^m$, $g^m(x,y;\zeta^m)$ is $L_g$-smooth and $\mu$-strongly convex function on variable $y$, i.e., $L_g I_p\succeq \nabla^2_{yy}g^m(x,y;\zeta^m) \succeq \mu I_p$.
\end{assumption}

\begin{assumption} \label{ass:2}
For any $m\in[M]$, $x\in \mathbb{R}^d$ and $y\in \mathbb{R}^p$, $\xi^m$ and $\zeta^m$, stochastic functions $f^m(x,y;\xi^m)$ and $g^m(x,y;\zeta^m)$ satisfy the following conditions hold:
$\nabla_x f^m(x,y;\xi^m)$ and $\nabla_y f^m(x,y;\xi^m)$ are $L_{f}$-Lipschitz continuous,
$\nabla_y g^m(x,y;\zeta^m)$ is $L_{g}$-Lipschitz continuous, $\nabla^2_{xy}g^m(x,y;\zeta^m)$ is $L_{gxy}$-Lipschitz continuous,
$\nabla^2_{yy}g^m(x,y;\zeta^m)$ is $L_{gyy}$-Lipschitz continuous. For example, for all $x,x_1,x_2 \in \mathbb{R}^d$ and $y,y_1,y_2\in \mathbb{R}^p$, we have
\begin{align}
  & \|\nabla_x f^m(x_1,y;\xi^m)-\nabla_x f^m(x_2,y;\xi^m)\| \leq L_f\|x_1-x_2\|, \nonumber \\
  & \|\nabla_x f^m(x,y_1;\xi^m)-\nabla_x f^m(x,y_2;\xi^m)\| \leq L_f\|y_1-y_2\|. \nonumber
\end{align}
\end{assumption}

\begin{assumption} \label{ass:3}
The partial derivatives $\nabla_y f^m(x,y)$ and $\nabla^2_{xy} g^m(x,y)$ are bounded for any $m\in [M]$, i.e., $\|\nabla_y f^m(x,y)\|^2 \leq C^2_{fy}$
and $\|\nabla^2_{xy} g^m(x,y)\|^2 \leq C^2_{gxy}$.
\end{assumption}

\begin{assumption} \label{ass:4}
Stochastic functions $f^m(x,y;\xi^m)$ and $g^m(x,y;\zeta^m)$ have unbiased stochastic partial derivatives
with bounded variance, e.g.,
\begin{align}
\mathbb{E}[\nabla_x f^m(x,y;\xi^m)] = \nabla_x f^m(x,y), \ \mathbb{E}\|\nabla_x f^m(x,y;\xi^m) - \nabla_x f^m(x,y) \|^2 \leq \sigma^2. \nonumber
\end{align}
The same assumptions hold for $\nabla_y f^m(x,y;\xi^m)$, $\nabla_y g^m(x,y;\zeta^m)$, $\nabla^2_{xy} g^m(x,y;\zeta^m)$ and $\nabla^2_{yy} g^m(x,y;\zeta^m)$.
\end{assumption}

Assumptions 1-4 are commonly used in stochastic bilevel optimization problems \citep{ghadimi2018approximation,hong2020two,ji2020bilevel,chen2021single,khanduri2021near,guo2021stochastic}.
Based on Assumption \ref{ass:2}, we have $\|\nabla_x f^m(x_1,y)-\nabla_x f^m(x_2,y)\|=\|\mathbb{E}[\nabla_x f^m(x_1,y;\xi^m)-\nabla_x f^m(x_2,y;\xi^m)]\|
\leq \mathbb{E}\|\nabla_x f^m(x_1,y;\xi^m)-\nabla_x f^m(x_2,y;\xi^m)\| \leq L_f\|x_1-x_2\|$, i.e., we can also obtain $\nabla_x f^m(x,y)$ is $L_f$-Lipschitz continuous,
which is similar for $\nabla_y f^m(x,y)$, $\nabla_y g^m(x,y)$, $\nabla^2_{xy}g^m(x,y)$ and $\nabla^2_{yy}g^m(x,y)$.
At the same time, based on Assumptions \ref{ass:3} and \ref{ass:4}, we also have
$\|\nabla_y f^m(x,y;\xi^m)\|^2 = \|\nabla_y f^m(x,y;\xi^m)-\nabla_y f^m(x,y)-\nabla_y f^m(x,y)\|^2 \leq 2\|\nabla_y f^m(x,y;\xi^m)-\nabla_y f^m(x,y)\|^2 + 2\|\nabla_y f^m(x,y)\|^2 \leq 2\sigma^2 + 2C^2_{fy}$ and $\|\nabla^2_{xy} g^m(x,y;\zeta^m)\|^2 \leq 2\sigma^2 + 2C^2_{gxy}$. Clearly, under Assumption \ref{ass:4}, the bounded $\nabla_y f^m(x,y)$ and $\nabla^2_{xy} g^m(x,y)$ are not milder than the bounded $\nabla_y f^m(x,y;\xi^m)$ and $\nabla^2_{xy} g^m(x,y;\zeta^m)$ for all $\xi^m$ and $\zeta^m$.

\subsection{ Distributed Bilevel Optimization }
In this subsection, we review the basic first-order method for solving the distributed bilevel problem,
defined as
\begin{align} \label{eq:5}
 \min_{x \in \mathbb{R}^d} \ F(x):=\frac{1}{M}\sum_{m=1}^Mf^m(x,y^*(x)), \ \mbox{s.t.} \ y^*(x) \in \arg\min_{y\in \mathbb{R}^p} \ g(x,y):=\frac{1}{M}\sum_{m=1}^M g^m(x,y).
\end{align}
Naturally, we use the gradient descent to update the variables $x,y$: at the $t$-th step
\begin{align}
 y_{t+1} = y_t - \lambda \nabla_y g(x_t,y_t), \quad x_{t+1} = x_t - \gamma \nabla F(x_t), \nonumber
\end{align}
where $\lambda>0$ and $\gamma>0$ denote the step sizes.
Clearly, if there does not exist a closed form solution of the inner problem in Problem \eqref{eq:5}, i.e., $y_{t+1}\neq y^*(x_t)$, we can not
easily obtain the gradient $\nabla F(x_t) = \nabla f(x_t,y^*(x_t))$. Thus, one of key points in solving Problem \eqref{eq:5} is to estimate the gradient $\nabla F(x_t)$.
For notational simplicity, let $\nabla^2_{xy}=\nabla_x\nabla_y$ 
and $\nabla^2_{yy}=\nabla_y\nabla_y$. In the following, we give two useful lemmas, 
whose proofs are provided in the Appendix. 
\begin{lemma} \label{lem:01}
Under the above Assumption \ref{ass:1}, we have, for any $x\in \mathbb{R}^d$,
\begin{align}
 &\nabla F(x) = \frac{1}{M}\sum_{m=1}^M\nabla_x f^m(x,y^*(x)) + \nabla y^*(x)^T\Big(\frac{1}{M}\sum_{m=1}^M\nabla_y f^m(x,y^*(x))\Big) \\
 & = \frac{1}{M}\sum_{m=1}^M\nabla_x f^m(x,y^*(x)) - \frac{1}{M}\sum_{m=1}^M\nabla^2_{xy} g^m(x,y^*(x))\Big[\frac{1}{M}\sum_{m=1}^M\nabla^2_{yy}g^m(x,y^*(x))\Big]^{-1}\Big(\frac{1}{M}\sum_{m=1}^M\nabla_y f^m(x,y^*(x))\Big). \nonumber
\end{align}
\end{lemma}
From the above Lemma \ref{lem:01}, when $M=1$, we have $$\nabla F(x) =\nabla_xf(x,y^*(x)) - \nabla^2_{xy}g(x,y^*(x)) \big(\nabla^2_{yy}g(x,y^*(x))\big)^{-1} \nabla_yf(x,y^*(x)),$$ which is the same result in Lemma 2.1 of \cite{ghadimi2018approximation}. Then, it is natural to use the following form to estimate $\nabla F(x)$, defined as
\begin{align}
 \hat{\nabla} f(x,y) = \nabla_xf(x,y) - \nabla^2_{xy}g(x,y) \big(\nabla^2_{yy}g(x,y)\big)^{-1} \nabla_yf(x,y). \nonumber
\end{align}
Since the cost of explicitly compute Hessian inverse matrix $\big(\nabla^2_{yy}g(x,y)\big)^{-1} $ is very expensive, 
we can use the Neumann series $\mu\sum_{k=0}^K\big(I_p-\nabla^2_{yy}g(x,y)\mu\big)^k \ (K\geq 1)$ to approximate $\big(\nabla^2_{yy}g(x,y)\big)^{-1}$, where
$\mu>0$. When $K=+ \infty$, we have $\mu\sum_{i=0}^\infty(I_p-\nabla^2_{yy}g(x,y)\mu)=\big(\nabla^2_{yy}g(x,y)\big)^{-1}$.
For each function $f^m(x,y^*(x))$, we also provide an approximated 
gradient $\hat{\nabla} f^m(x,y)$, defined as 
\begin{align}
 \hat{\nabla} f^m(x,y) =  \nabla_xf^m(x,y) - \nabla^2_{xy}g^m(x,y) \big(\nabla^2_{yy}g^m(x,y)\big)^{-1} \nabla_yf^m(x,y), \quad \forall m\in [M]. 
\end{align}
In the paper, we consider the distributed stochastic bilevel optimization problem  (\ref{eq:1}). For any $m\in[M]$, $f^m(x,y^*(x))=\mathbb{E}_{\xi^m}\big[f^m(x,y^*(x);\xi^m)\big]$ 
and $g^m(x,y)=\mathbb{E}_{\zeta^m}\big[g^m(x,y;\zeta^m)\big]$.  

\begin{lemma} \label{lem:02}
Under the above Assumptions (\ref{ass:1}, \ref{ass:2}, \ref{ass:3}), the global functions (or mappings) $F(x)=\frac{1}{M}\sum_{m=1}^Mf^m(x,y^*(x))$ and $y^*(x)=\arg\min_{y\in \mathbb{R}^p}\frac{1}{M}\sum_{m=1}^Mg^m(x,y)$ satisfy, for all $x_1,x_2\in \mathbb{R}^d$, 
\begin{align}
 & \|y^*(x_1)-y^*(x_2)\| \leq \kappa\|x_1-x_2\|, \quad \|\nabla y^*(x_1) - \nabla y^*(x_2)\| \leq L_y\|x_1-x_2\| \nonumber \\
 & \|\nabla F(x_1) - \nabla F(x_2)\|\leq L\|x_1-x_2\|,  \nonumber
\end{align}
where $\kappa=C_{gxy}/\mu$, $L_y=\big( \frac{C_{gxy}L_{gyy}}{\mu^2} +  \frac{L_{gxy}}{\mu} \big) (1+ \frac{C_{gxy}}{\mu})$ and
$L=\Big(L_f + \frac{C_{gxy}L_f}{\mu} + C_{fy}\big( \frac{C_{gxy}L_{gyy}}{\mu^2} +  \frac{L_{gxy}}{\mu} \big)\Big)(1+\frac{C_{gxy}}{\mu})$.
\end{lemma}
The above Lemma \ref{lem:02} shows that $y^*(x)$ is $\kappa$-Lipschitz continuous, 
 $\nabla y^*(x)$ is $L_y$-Lipschitz continuous, and $\nabla F(x)$ is $L$-Lipschitz continuous.
 
\begin{algorithm}[t]
\caption{Adaptive Federated Bilevel Optimization (\textbf{AdaFBiO}) Algorithm}
\label{alg:1}
\begin{algorithmic}[1]
\STATE {\bfseries Input:} $T, q, K \in \mathbb{N}^+$, tuning parameters $\{\gamma, \lambda, \eta_t, \alpha_t, \beta_t\}$, initial inputs $x_1\in \mathbb{R}^d$, $y_1\in \mathbb{R}^p$; \\
\STATE {\bfseries initialize:} Set $x^m_1=x_1$ and $y^m_1=y_1$ for $m \in [M]$, and draw $q(K+2)$ independent samples $\{\bar{\xi}^{m,i}_1\}_{i=1}^q=\{\xi^{m,i}_1,\zeta^{m,i}_{1,0},\zeta^{m,i}_{1,1}, \cdots, \zeta^{m,i}_{1,K-1}\}_{i=1}^q$ and $\{\zeta^{m,i}_1\}_{i=1}^q$,
and then compute $v^m_1 = \frac{1}{q}\sum_{i=1}^q\nabla_y g^m(x^m_1,y^m_1;\zeta^{m,i}_1)$, and $w^m_1 = \frac{1}{q}\sum_{i=1}^q\hat{\nabla}f^m(x^m_1,y^m_1;\bar{\xi}^{m,i}_1)$ for all $m \in [M]$;
Generate adaptive matrices $A_1 \in \mathbb{R}^{d \times d}$ and $B_1 \in \mathbb{R}^{p \times p}$. \\
\FOR{$t=1$ \textbf{to} $T$}
\IF {$\mod(t,q)=0$}
\STATE $\bar{v}_t = \frac{1}{M} \sum_{m=1}^{M} v^m_t$, $\bar{w}_t = \frac{1}{M} \sum_{m=1}^{M} w^m_t$, $\bar{y}_t = \frac{1}{M} \sum_{m=1}^{M} y^m_t$, $\bar{x}_t = \frac{1}{M} \sum_{m=1}^{M} x^m_t$; \\
\STATE Generate the adaptive matrices $A_t \in \mathbb{R}^{d \times d}$ and $B_t \in \mathbb{R}^{p \times p}$;\\
\textcolor{blue}{One example of $A_t$ and $B_t$ by using update rule ($a_0 = 0$, $b_0 = 0$, $ 0 < \varrho_t < 1$, $\rho>0$.) } \\
\textcolor{blue}{ Compute $ a_t = \varrho_t a_{t-1} + (1 - \varrho_{t})\bar{w}_t^2$, $A_t = \mbox{diag}(\sqrt{a_t} + \rho)$}; \\
\textcolor{blue}{ Compute $ b_t = \varrho_t b_{t-1} + (1 - \varrho_t)||\bar{v}_{t}||$, $B_t = (b_t + \rho)I_p$}; \\
\STATE $\hat{y}^m_{t+1}=\hat{y}_{t+1} = \bar{y}_t - \lambda B_{t}^{-1} \bar{v}_t$, $\hat{x}^m_{t+1}=\hat{x}_{t+1} = \bar{x}_t - \gamma A_{t}^{-1} \bar{w}_t$; \\
\STATE $y^m_{t+1}=\bar{y}_{t+1} = \bar{y}_t + \eta_t(\hat{y}_{t+1}-\bar{y}_t)$,
$x^m_{t+1}=\bar{x}_{t+1} = \bar{x}_t + \eta_t(\hat{x}_{t+1}-\bar{x}_t)$; (Sent them to Clients) \\
\ELSE
\FOR{$m=1,\cdots,M$ \ (\textbf{in parallel})}
\STATE $\hat{y}^m_{t+1} = y^m_t - \lambda B_t^{-1} v^m_t$, $\hat{x}^m_{t+1} = x^m_t - \gamma A_t^{-1} w^m_t$; \\
\STATE $y^m_{t+1} = y^m_t + \eta_t(\hat{y}^m_{t+1}-y^m_t)$, $x^m_{t+1} = x^m_t + \eta_t(\hat{x}^m_{t+1}-x^m_t)$; \\
\STATE $A_{t+1} = A_t$, $B_{t+1} = B_t$; \\
\ENDFOR
\ENDIF
\FOR{$m=1,\cdots,M$ \ (\textbf{in parallel})}
\STATE Draw independent samples $\bar{\xi}^m_{t+1}=\{\xi^m_{t+1},\zeta^m_{t+1,0}, \cdots, \zeta^m_{t+1,K-1}\}$ and $\zeta^m_{t+1}$;
\STATE $v^m_{t+1} = \nabla_y g^m(x^m_{t+1},y^m_{t+1};\zeta^m_{t+1}) + (1-\alpha_{t+1})\big[v^m_t - \nabla_y g^m(x^m_t,y^m_t;\zeta^m_{t+1})\big]$; \\
\STATE $w^m_{t+1} = \hat{\nabla} f^m(x^m_{t+1},y^m_{t+1};\bar{\xi}^m_{t+1}) + (1-\beta_{t+1})\big[w^m_t - \hat{\nabla}
f^m(x^m_t,y^m_t;\bar{\xi}^m_{t+1})\big] $; \\
\ENDFOR
\ENDFOR
\STATE {\bfseries Output:} Chosen uniformly random from $\{\bar{x}_t\}_{t=1}^{T}$.
\end{algorithmic}
\end{algorithm}

\section{ Adaptive Federated Bilevel Optimization Algorithm }
In this section, we propose a fast adaptive federated bilevel optimization (AdaFBiO) algorithm to
solve Problem \eqref{eq:1}, which builds on the momentum-based variance reduced and
local-SDG techniques. Meanwhile, our AdaFBiO algorithm uses the unified adaptive matrices to flexibly incorporate various adaptive learning rates to
update variables in both UL and LL subproblems of Problem \eqref{eq:1}.
Specifically, Algorithm \ref{alg:1} provides a procedure framework of our AdaFBiO algorithm.

In Algorithm \ref{alg:1}, when $\mbox{mod}(t,q)= 0$ (i.e., synchronization step), the server receives the updated variables $\{y^m_t, x^m_t\}_{m=1}^M$ and
estimated stochastic gradients $\{v_t^m,w_t^m\}_{m=1}^M$ from the clients, and then averages them to obtain
the averaged variables $\{\bar{y}_t,\bar{x}_t\}$ and averaged gradients $\{\bar{v}_t,\bar{w}_t\}$. Based on
these averaged gradients, we can generate some adaptive matrices (i.e., adaptive learning rates). Besides one example given
at the line 6 of Algorithm \ref{alg:1}, we can also generate many adaptive matrices. For example, we can generate
adaptive matrices as in AdaBelief~\citep{zhuang2020adabelief} algorithm, defined as
\begin{align}
 & a_t = \varrho_t a_{t-1} + (1 - \varrho_{t})\big(\bar{w}_t - \bar{w}_{t_0}\big)^2, \quad A_t = \mbox{diag}(\sqrt{a_t} + \rho), \\
 & b_t = \varrho_t b_{t-1} + (1 - \varrho_t)||\bar{v}_{t}-\bar{v}_{t_0}||, \quad B_t = (b_t + \rho)I_p,
\end{align}
where $t_0=t-q$. Note that we can directly choose $\alpha_t$ and $\beta_t$ instead of $\varrho_t$ to reduce the number of tuning parameters in
our algorithm. Next, based on these adaptive matrices, we can update the variables $x$ and $y$ in the server,
then sent these updated variables to each client.

When $\mbox{mod}(t,q) \neq 0$ (i.e., asynchronization step), the clients receive the updated variables $\{\bar{x}_{t+1},\bar{y}_{t+1}\}$
and the generated adaptive matrices $\{A_t,B_t\}$ from the server. Then the clients use the momentum-based variance reduced technique of STORM~\citep{cutkosky2019momentum}
to update the stochastic gradients based on local data: for $m\in [M]$
\begin{align}
& v^m_{t+1} = \nabla_y g^m(x^m_{t+1},y^m_{t+1};\zeta^m_{t+1}) + (1-\alpha_{t+1})\big[v^m_t - \nabla_y g^m(x^m_t,y^m_t;\zeta^m_{t+1})\big]  \\
& w^m_{t+1} = \hat{\nabla} f^m(x^m_{t+1},y^m_{t+1};\bar{\xi}^m_{t+1}) + (1-\beta_{t+1})\big[w^m_t - \hat{\nabla} f^m(x^m_t,y^m_t;\bar{\xi}^m_{t+1})\big]
\end{align}
where $\alpha_{t+1}\in (0,1)$ and $\beta_{t+1}\in (0,1)$. Based on the estimated stochastic gradients and adaptive matrices, the clients
update the variables $\{x^m_t,y^m_t\}_{m=1}^M$, defined as
\begin{align}
 & \hat{y}^m_{t+1} = y^m_t - \lambda B_t^{-1} v^m_t, \quad y^m_{t+1} = y^m_t + \eta_t(\hat{y}^m_{t+1}-y^m_t), \\
 & \hat{x}^m_{t+1} = x^m_t - \gamma A_t^{-1} w^m_t, \quad x^m_{t+1} = x^m_t + \eta_t(\hat{x}^m_{t+1}-x^m_t)
\end{align}
which is equivalent to the following formation:
\begin{align}
 y^m_{t+1} = y^m_t - \lambda \eta_t B_t^{-1} v^m_t,
 \quad x^m_{t+1} = x^m_t - \gamma \eta_t A_t^{-1} w^m_t.
\end{align}
In Algorithm \ref{alg:1}, all clients use the same adaptive matrices generated from the server. Moreover, our AdaFBiO algorithm uses the unified adaptive matrices $\{A_t,B_t\}$ to flexibly incorporate various adaptive learning rates.

At the line 17 of Algorithm \ref{alg:1},
we draw $K+1$ independent samples $\bar{\xi}^m = \{\xi^m,\zeta^m_0,\zeta^m_1,\cdots, \zeta^m_{K-1}\}$, then
provide a stochastic gradient estimator of the function $f^m(x,y^*(x))$, defined as
\begin{align} \label{eq:13}
 \hat{\nabla} f^m(x,y,\bar{\xi}^m) &  = \nabla_xf^m(x,y;\xi^m) - \nabla^2_{xy}g^m(x,y;\zeta^m_0) \nonumber \\
 & \qquad \cdot \bigg[ \frac{K}{L_g}\prod_{i=1}^k \big(I_p - \frac{1}{L_g}\nabla^2_{yy}g^m(x,y;\zeta^m_i)\big) \bigg] \nabla_yf^m(x,y;\xi^m),
\end{align}
where $K\geq 1$ and $k\sim \mathcal{U}\{0,1, \cdots, K-1\}$ is a uniform random variable independent on $\bar{\xi}$. Here we use the term $\frac{K}{L_g}\prod_{i=1}^k \big( I_p - \frac{1}{L_g}\nabla^2_{yy}g^m(x,y;\zeta^m_i)\big)$ to
approximate the Hessian inverse, i.e., $\big(\nabla^2_{yy} g^m(x,y;\zeta^m)\big)^{-1}$. In practice, we can use a tuning parameter $\vartheta\in (0,\frac{1}{L_g}]$ instead of
the parameter $\frac{1}{L_g}$ in the above gradient estimator \eqref{eq:13}, as in~\cite{khanduri2021near}.
Clearly, the above $\hat{\nabla} f^m(x,y,\bar{\xi}^m)$ is a biased estimator in estimating $\hat{\nabla} f^m(x,y)$, i.e.
$\mathbb{E}_{\bar{\xi}^m}\big[\hat{\nabla}f^m(x,y;\bar{\xi}^m)\big] \neq \hat{\nabla}f^m(x,y)$.

\textbf{Note that} our AdaFBiO algorithm does not estimate the global gradient $$\nabla F(x)=\frac{1}{M}\sum_{m=1}^M
\Big(\nabla_xf^m(x,y^*(x))- \nabla^2_{xy} g^m(x,y^*(x))\Big[\frac{1}{M}\sum_{m=1}^M\nabla^2_{yy}g^m(x,y^*(x))\Big]^{-1}\big(\frac{1}{M}\sum_{m=1}^M\nabla_y f^m(x,y^*(x))\big)\Big)$$ or the global information of $g(x,y^*(x))=\frac{1}{M}\sum_{m=1}^Mg^m(x,y^*(x))$,
i.e.,$$\Big(\frac{1}{M}\sum_{m=1}^M\nabla^2_{xy}g^m(x,y^*(x))\Big)\Big(\frac{1}{M}\sum_{m=1}^M\nabla^2_{yy}g^m(x,y^*(x))\Big)^{-1},$$ as in~\citep{tarzanagh2022fednest,chen2022decentralized}
to control the variance generated from data heterogeneity. However, we only estimates the local gradient $\nabla_xf^m(x,y^*(x))$ at each client based on the local data, which can also guarantee the convergence 
of our algorithms.
Moreover, our AdaFBiO algorithm still can obtain the best known sample and communication complexities simultaneously under the non-i.i.d. setting (Please see the following convergence analysis).
At the same time, our AdaFBiO algorithm does not rely on some unreasonable assumptions such as the Assumption 4 of~\citep{gao2022convergence}.

\section{Theoretical Analysis}
In this section, we study the convergence properties of our AdaFBiO algorithm under some mild assumptions.
All related proofs are provided in the Appendix. We first review some useful lemmas and assumptions.

\begin{lemma} \label{lem:2}
(Lemma 2.1 of \cite{khanduri2021near})
 Under the about Assumptions \ref{ass:1} and \ref{ass:3}, for any $K\geq 1$, $m\in [M]$, the gradient estimator in \eqref{eq:13} satisfies
 \begin{align}
  \|R^m(x,y)\| \leq \frac{C_{gxy}C_{fy}}{\mu}\big(1 - \frac{\mu}{L_g}\big)^K,
 \end{align}
where $R^m(x,y)=\hat{\nabla}f^m(x,y) - \mathbb{E}_{\bar{\xi}^m}\big[\hat{\nabla} f^m(x,y;\bar{\xi}^m)\big]$.
\end{lemma}

\begin{lemma} \label{lem:3}
(Lemma 3.1 of \cite{khanduri2021near})
Under the above Assumptions \ref{ass:1}-\ref{ass:3}, stochastic gradient estimator $\hat{\nabla}f^m(x,y;\bar{\xi}^m)$ for any $m\in [M]$ is $L_K$-Lipschitz continuous,
 such that for $x,x_1,x_2\in \mathbb{R}^d$ and $y,y_1,y_2\in \mathbb{R}^p$,
 \begin{align}
   & \mathbb{E}_{\bar{\xi}^m}\|\hat{\nabla} f^m(x_1,y;\bar{\xi}^m) - \hat{\nabla} f^m(x_2,y;\bar{\xi}^m)\|^2 \leq L^2_K\|x_1-x_2\|^2, \nonumber \\
   & \mathbb{E}_{\bar{\xi}^m}\|\hat{\nabla} f^m(x,y_1;\bar{\xi}^m) - \hat{\nabla} f^m(x,y_2;\bar{\xi}^m)\|^2 \leq L^2_K\|y_1-y_2\|^2, \nonumber
 \end{align}
 where $L_K^2 = 2L^2_f + 6C^2_{gxy}L^2_f\frac{K}{2\mu L_g - \mu^2} + 6C^2_{fy}L^2_{gxy}\frac{K}{2\mu L_g - \mu^2} + 6C^2_{gxy}L^2_f\frac{K^3L^2_{gyy}}{(L_g-\mu)^2(2\mu L_g - \mu^2)}$.
\end{lemma}

According to Lemma \ref{lem:2}, choose $K=\frac{L_g}{\mu}\log(C_{gxy}C_{fy}T/\mu)$
in Algorithm \ref{alg:1}, we have
 $\|R^m(x,y)\|\leq \frac{1}{T}$ for all $t\geq 1$. Lemma \ref{lem:2}
  shows that the bias $R^m(x,y)$ decays exponentially fast with number $K$.
For notational simplicity, let $R^m_t=R^m(x_t,y_t)$ for all $t\geq 1$.
Lemma \ref{lem:3} shows the smoothness of gradient estimator $\hat{\nabla} f^m(x,y;\bar{\xi}^m)$, for any $\xi^m \sim \mathcal{D}^m$, $m\in [M]$.

\begin{assumption} \label{ass:5}
 The estimated stochastic partial derivative $\hat{\nabla} f^m(x,y;\bar{\xi}^m)$  for any $m\in [M]$ satisfies $\mathbb{E}_{\bar{\xi}^m}\big[\hat{\nabla} f^m(x,y;\bar{\xi}^m)\big]
 = \hat{\nabla} f^m(x,y) + R^m(x,y)$,
 $\mathbb{E}_{\bar{\xi}^m}\|\hat{\nabla}f^m(x,y;\bar{\xi}^m) - \hat{\nabla} f^m(x,y) - R^m(x,y)\|^2 \leq \sigma^2$.
\end{assumption}

\begin{assumption} \label{ass:6}
In our algorithms, the adaptive matrices $A_t\succeq \rho I_d \succ 0$ for all $t\geq 1$ for updating the variables $x$, where $\rho>0$ is an appropriate positive number.
The adaptive matrices $B_t=b_tI_p \ (\hat{b} \geq b_t \geq \rho>0)$ for all $t\geq 1$ for updating the variables $y$.
\end{assumption}

\begin{assumption} \label{ass:7}
 For any $m, j \in [M]$, $x\in \mathbb{R}^d$ and $y\in \mathbb{R}^p$, we have $ \| \nabla_x f^{m}(x, y) -  \nabla_x f^{j}(x, y) \| \leq \delta_f$, $\|\nabla_y f^{m} (x, y) -  \nabla_y f^{j}(x, y)\| \leq \delta_f$,
 $ \| \nabla_{y} g^{m}(x, y) - \nabla_{y} g^{j}(x, y)\| \leq \delta_{g}$, $ \| \nabla_{xy} g^{m}(x, y) - \nabla_{xy} g^{j}(x, y)\| \leq \delta_{g}$, $\|\nabla_{yy}g^{m}(x, y)- \nabla_{yy}g^{j}(x, y)\| \leq \delta_{g}$, where $\delta_f>0$ and $\delta_{g}>0$ are constants.
\end{assumption}

\begin{assumption} \label{ass:8}
The function $F(x)$ is lower bounded, i.e.,  $F^* = \inf_{x\in \mathbb{R}^d} F(x)$.
\end{assumption}

Assumption \ref{ass:5} is commonly used in stochastic bilevel optimization~\citep{khanduri2021near,tarzanagh2022fednest}.
Assumption \ref{ass:6} is commonly used in the adaptive gradient methods~\cite{huang2021super,huang2021biadam}.
Assumption \ref{ass:7} guarantees to constrain the data heterogeneity in non-i.i.d setting,
which is commonly used in FL algorithm~\cite{li2022local}.
Assumption~\ref{ass:8} ensures the feasibility of Problem~\eqref{eq:1}.

\begin{lemma} \label{lem:5}
Assume the sequences $\big\{\bar{w}_t,\bar{x}_t,\bar{y}_t\big\}_{t=1}^T$ generated from Algorithm \ref{alg:1},
 we have
 \begin{align}
 \|\bar{w}_t-\nabla F(\bar{x}_t)\|^2 \leq 8\bar{L}^2\|y^*(\bar{x}_t)-\bar{y}_t\|^2 + 2\|\bar{w}_t-\hat{\nabla} f(\bar{x}_t,\bar{y}_t)\|^2,
\end{align}
where $\bar{L}^2= L^2_f+ \frac{L^2_{gxy}C^2_{fy}}{\mu^2} + \frac{L^2_{gyy} C^2_{gxy}C^2_{fy}}{\mu^4} +
 \frac{L^2_fC^2_{gxy}}{\mu^2}$.
\end{lemma}

Based on the about Assumption \ref{ass:7}, as in \cite{li2022local}, we show
the bound of heterogeneity in estimated gradients $\hat{\nabla}f^m(x,y)$ and $\hat{\nabla}f^j(x,y)$ for any $m,j\in[M]$ in the following lemma.
\begin{lemma} \label{lem:6}
Based on the above Assumptions \ref{ass:1}, \ref{ass:3} and \ref{ass:7}, for any $m,j\in[M]$, we have
\begin{align}
 \big\|\hat{\nabla}f^m(x,y)-\hat{\nabla}f^j(x,y)\big\|^2 \leq \hat{\delta}^2,
\end{align}
where $\hat{\delta}^2=4\delta^2_f + \frac{4C^2_{fy}\delta^2_{g}}{\mu^2} + \frac{4C^2_{gxy}C^2_{fy}\delta^2_{g}}{\mu^4} + \frac{4C^2_{gxy}\delta^2_{f}}{\mu^2}$.
\end{lemma}

From Lemma \ref{lem:6}, since $\max\big(C_{fy},C_{gxy}\big)>\mu$, the variance of heterogeneity in $\{\hat{\nabla}f^m(x,y)\}_{m=1}^M$ is much larger than
the variance of heterogeneity in $\{\nabla_y g^m(x,y)\}_{m=1}^M$. Clearly, the Assumption 4 of~\citep{gao2022convergence} is not reasonable.
Meanwhile, the following lemma
shows that each gradient $\hat{\nabla}f^m(x,y)$ for $m\in [M]$ is $\hat{L}$-Lipschitz continuous.

\begin{lemma} \label{lem:7}
Based on the above Assumptions \ref{ass:1}-\ref{ass:3}, we have, for any $m\in[M]$, $x_1,x_2\in \mathbb{R}^d$ and $y_1,y_2\in \mathbb{R}^p$,
\begin{align}
\|\hat{\nabla}f^m(x_1,y_1)-\hat{\nabla}f^m(x_2,y_2)\|^2 \leq \hat{L}^2\big(\|x_1-x_2\|^2+\|y_1-y_2\|^2\big),
\end{align}
where $\hat{L}=\sqrt{8L^2_f + \frac{8L^2_{gxy}C^2_{fy}}{\mu^2} + \frac{8L^2_{gyy}C^2_{gxy}C^2_{fy}}{\mu^4} + \frac{8L^2_fC^2_{gxy}}{\mu^2}}$.
\end{lemma}

Next, based on the above assumptions, we give the convergence properties of our AdaFBiO algorithm.

\subsection{ Convergence Analysis of AdaFBiO }

\begin{theorem} \label{th:1}
(\textbf{Adaptive Algorithm} \ref{alg:1})
Suppose the sequence $\{\bar{x}_t,\bar{y}_t\}_{t=1}^T$ be generated from Algorithm \ref{alg:1}.
 Under the above Assumptions, and let $\eta_t=\frac{kM^{1/3}}{(n+t)^{1/3}}$ for all $t\geq 0$, $\alpha_{t+1}=c_1\eta_t^2$, $\beta_{t+1}=c_2\eta_t^2$, $n \geq \max\big(2, Mk^3, M(c_1k)^3, M(c_2k)^3, \frac{(12k\lambda q)^3M^{5/2}(L^2_K+L^2_g)^{3/2}}{(\theta\rho)^3}\big)$, $k>0$, $c_2=\vartheta c_1 \ (\vartheta>0)$,
 $\frac{2}{3k^3} + \frac{1000\bar{L}^2}{3\mu^2} \leq c_1 \leq \frac{72\lambda^2q(L^2_K+L^2_g)}{\rho^2\sqrt{\vartheta^2\hat{L}^2+L^2_g}}$, $c_2 \geq \frac{2}{3k^3} + 34$, $\lambda=\tau\gamma$,
 $0<\tau \leq \min\Big(\frac{1}{8}\sqrt{\frac{15M\rho\gamma}{\Gamma}},1\Big)$, $0<\theta\leq \min\Big(9\bar{L}\sqrt{\frac{75\lambda(L^2_K+L^2_g)M\mu}{\rho\big(30\hat{L}^2\mu^2
 +1000\bar{L}^2L^2_g + 52\hat{L}^2\mu^2\big)}},1\Big)$, \\
$0< \gamma \leq \min\Big(\frac{\rho}{8}\sqrt{\frac{1}{(125\bar{L}^2\kappa^2\hat{b}^2)/(6\mu^2\lambda^2)+(L^2_g+L^2_K)/M}}, \frac{n^{1/3}\rho}{4LkM^{1/3}} \Big)$, $0< \lambda \leq \frac{225M\rho\bar{L}^2}{184\mu(L^2_K+L^2_g)}$ and $K=\frac{L_g}{\mu}\log(C_{gxy}C_{fy}T/\mu)$, we have
\begin{align}
 \frac{1}{T}\sum_{t=1}^T\mathbb{E}\|\nabla F(\bar{x}_t)\| \leq \Big( \frac{\sqrt{3G}n^{1/6}}{M^{1/6}T^{1/2}} + \frac{\sqrt{3G}}{M^{1/6}T^{1/3}}\Big)\sqrt{\frac{1}{T}\sum_{t=1}^T\mathbb{E}\|A_t\|^2},
\end{align}
where $G = \frac{4(F(\bar{x}_1) - F^*)}{k\rho\gamma} + \frac{160b_1\bar{L}^2\Delta_0}{k\lambda\mu\rho^2} + \frac{8n^{1/3}\sigma^2}{qM^{4/3}k^2\rho^2} + 8k^2\Big(\frac{(c^2_1+c^2_2)\sigma^2}{\rho^2}
 + \frac{2\hat{c}^2\Gamma }{15\rho\gamma\lambda^2(L^2_K+L^2_g)}\Big)\ln(n+T) + \frac{54kM^{1/3}(n+T)^{2/3}}{T^2\rho^2}$, $\Delta_0 = \|\bar{y}_1-y^*(\bar{x}_1)\|^2$,
 $\hat{c}^2 = \frac{2c^2_2}{T^2} + 2c^2_2\sigma^2 + c^2_1\sigma^2 + 3c^2_2\hat{\delta}^2 + 3c^2_1\delta_{g}^2$, $\hat{\delta}^2=4\delta^2_f + \frac{4C^2_{fy}\delta^2_{g}}{\mu^2} + \frac{4C^2_{gxy}C^2_{fy}\delta^2_{g}}{\mu^4} + \frac{4C^2_{gxy}\delta^2_{f}}{\mu^2}$ and $\Gamma = \frac{5\theta^2\hat{L}^2\gamma\rho}{36(L^2_K+L^2_g)} + \frac{125\theta^2\bar{L}^2 L^2_g\gamma \rho}{27\mu^2(L^2_K+L^2_g)} + \frac{8(L^2_K+L^2_g)\lambda^2\gamma}{\rho} + \frac{17\theta^2\hat{L}^2\rho\gamma }{72(L^2_K+L^2_g)}$.
\end{theorem}

\begin{remark}
Under Assumptions \ref{ass:1} and \ref{ass:3}, and assume the bounded gradient $\|\nabla_x f^m(x^m_t,y^m_t;\xi^m_t)\| \leq C_{fx}$ for all $m\in [M]$, then we have $\big\|\frac{1}{M}\sum_{m=1}^M\hat{\nabla} f^m(x^m_t,y^m_t;\bar{\xi}^m_t)\big\| \leq C_{fx} + K(C_{gxy}+\sigma)(C_{fy}+\sigma)/L_g$. When the matrix $A_t$ is generated from the line 6 of Algorithm \ref{alg:1}, we have $\sqrt{\frac{1}{T}\sum_{t=1}^T\mathbb{E}\|A_t\|^2} \leq \sqrt{2C_{fx} + 2K(C_{gxy}+\sigma)(C_{fy}+\sigma)/L_g + 2\rho}$.
\end{remark}

\begin{remark}
Without loss of generality, let $k=O(1)$, $\rho=O(1)$, $\hat{b}=O(1)$, $c_1=O(1)$, $c_2=O(1)$ and $n=O(q^3)$,
we have $G=\tilde{O}(1)$ and $\sqrt{\frac{1}{T}\sum_{t=1}^T\mathbb{E}\|A_t\|^2}=O(1)$. Based on the above Theorem \ref{th:1}, let
$q=T^{1/3}$ and
\begin{align}
\frac{1}{T}\sum_{t=1}^T\mathbb{E}\|\nabla F(\bar{x}_t)\| \leq \tilde{O}\Big(\frac{\sqrt{q}}{\sqrt{T}}+\frac{1}{T^{1/3}}\Big)= \tilde{O}\Big(\frac{1}{T^{1/3}}\Big) \leq \epsilon,
\end{align}
then we can obtain $T=\tilde{O}(\epsilon^{-3})$ and $K=\tilde{O}(1)$. Our AdaFBiO algorithm requires $K+2$ samples at each iteration expect for the first iteration requires $q(K+2)$ samples,
so it has a sample complexity of $q(K+2)+(K+2)T = \tilde{O}(\epsilon^{-3})$ 
for finding an $\epsilon$-stationary point of Problem~(\ref{eq:1}). 
Meanwhile, our AdaFBiO algorithm requires a communication complexity of $\frac{T}{q} = T^{2/3}=\tilde{O}(\epsilon^{-2})$.

\end{remark}

\begin{remark}
From the above Theorem \ref{th:1}, our AdaFBiO algorithm simultaneously obtain the best known sample and communication complexities in finding an $\epsilon$-stationary point of the above bilevel problem \eqref{eq:1}. Besides using effective adaptive learning rates in our algorithm, our convergence analysis is more reasonable than that of~\cite{gao2022convergence},
and our studied distributed bilevel problem \eqref{eq:1} is more general than that of~\cite{li2022local}.
Meanwhile, our AdaFBiO algorithm simultaneously have lower sample and communication complexities than these of FEDNEST algorithm~\citep{tarzanagh2022fednest}.
Moreover, our AdaMFCGD algorithm simultaneously have lower sample and communication complexities than the existing adaptive single-level FL methods such as the local-AMSGrad~\citep{chen2020toward} that requires the sample complexity of $O(\epsilon^{-4})$ and communication complexity of $O(\epsilon^{-3})$ to find an $\epsilon$-stationary point of the single-level optimization problems.
\end{remark}

\subsection{ Convergence Analysis of Non-Adaptive Version of AdaFBiO }

\begin{theorem} \label{th:2}
(\textbf{Non-Adaptive Algorithm \ref{alg:1}}, i.e., $A_t=I_d$ and $B_t=I_p$ for all $t\geq 1$)
Suppose the sequence $\{\bar{x}_t,\bar{y}_t\}_{t=1}^T$ be generated from Algorithm \ref{alg:1}.
 Under the above Assumptions, and let $\eta_t=\frac{kM^{1/3}}{(n+t)^{1/3}}$ for all $t\geq 0$, $\alpha_{t+1}=c_1\eta_t^2$, $\beta_{t+1}=c_2\eta_t^2$, \\ $n \geq \max\big(2, Mk^3, M(c_1k)^3, M(c_2k)^3, \frac{(12k\lambda q)^3M^{5/2}(L^2_K+L^2_g)^{3/2}}{\theta^3}\big)$, $k>0$, $c_2=\vartheta c_1 \ (\vartheta>0)$,
 $\frac{2}{3k^3} + \frac{1000\bar{L}^2}{3\mu^2} \leq c_1 \leq \frac{72\lambda^2q(L^2_K+L^2_g)}{\sqrt{\vartheta^2\hat{L}^2+L^2_g}}$, $c_2 \geq \frac{2}{3k^3} + 34$, $\lambda=\tau\gamma$,
 $0<\tau \leq \min\Big(\frac{1}{8}\sqrt{\frac{15M\gamma}{\Gamma}},1\Big)$, $0<\theta\leq \min\Big(9\bar{L}\sqrt{\frac{75\lambda(L^2_K+L^2_g)M\mu}{
 1000\bar{L}^2L^2_g + 82\hat{L}^2\mu^2}},1\Big)$,
$0< \gamma \leq \min\Big(\frac{1}{8}\sqrt{\frac{1}{(125\bar{L}^2\kappa^2)/(6\mu^2\lambda^2)+(L^2_g+L^2_K)/M}}, \frac{n^{1/3}}{4LkM^{1/3}} \Big)$, $0< \lambda \leq \frac{225M\bar{L}^2}{184\mu(L^2_K+L^2_g)}$ and $K=\frac{L_g}{\mu}\log(C_{gxy}C_{fy}T/\mu)$, we have
\begin{align}
 \frac{1}{T}\sum_{t=1}^T\mathbb{E}\|\nabla F(\bar{x}_t)\| \leq \frac{\sqrt{3G}n^{1/6}}{M^{1/6}T^{1/2}} + \frac{\sqrt{3G}}{M^{1/6}T^{1/3}},
\end{align}
where $G = \frac{4(F(\bar{x}_1) - F^*)}{k\gamma} + \frac{160\bar{L}^2\Delta_0}{k\lambda\mu} + \frac{8n^{1/3}\sigma^2}{qM^{4/3}k^2} + 8k^2\Big((c^2_1+c^2_2)\sigma^2
 + \frac{2\hat{c}^2\Gamma}{15\gamma\lambda^2(L^2_K+L^2_g)}\Big)\ln(n+T) + \frac{54kM^{1/3}(n+T)^{2/3}}{T^2}$, $\Delta_0 = \|\bar{y}_1-y^*(\bar{x}_1)\|^2$,
 $\hat{c}^2 = \frac{2c^2_2}{T^2} + 2c^2_2\sigma^2 + c^2_1\sigma^2 + 3c^2_2\hat{\delta}^2 + 3c^2_1\delta_{g}^2$, $\hat{\delta}^2=4\delta^2_f + \frac{4C^2_{fy}\delta^2_{g}}{\mu^2} + \frac{4C^2_{gxy}C^2_{fy}\delta^2_{g}}{\mu^4} + \frac{4C^2_{gxy}\delta^2_{f}}{\mu^2}$ and $\Gamma = \frac{5\theta^2\hat{L}^2\gamma}{36(L^2_K+L^2_g)} + \frac{125\theta^2\bar{L}^2 L^2_g\gamma }{27\mu^2(L^2_K+L^2_g)} + 8(L^2_K+L^2_g)\lambda^2\gamma + \frac{17\theta^2\hat{L}^2\gamma }{72(L^2_K+L^2_g)}$.
\end{theorem}

\begin{remark}
The proofs of Theorem \ref{th:2} can totally follow the proofs of the above Theorem \ref{th:1} with $A_t=I_d$ and $B_t=I_p$ for all $t\geq 1$, and $\rho=\hat{b}=1$.
Since the conditions of Theorem \ref{th:2} are similar to these of of Theorem \ref{th:1}, clearly, the \textbf{non-adaptive version} of AdaFBiO algorithm still can obtain the best known sample (or gradient) complexity of $\tilde{O}(\epsilon^{-3})$ and communication complexity of $\tilde{O}(\epsilon^{-2})$ in finding an $\epsilon$-stationary point of the above bilevel problem \eqref{eq:1}.
\end{remark}

\section{Numerical Experiments}
In this section, we apply some numerical experiments to demonstrate the efficiency of our AdaFBiO algorithm on federated hyper-representation learning
and federated data hyper-cleaning tasks.
We compare our AdaFBiO algorithm with the existing state-of-the-art algorithms in Table \ref{tab:1} for solving bilevel optimization problems.

\subsection{ Federated Hyper-Representation Learning }

\subsection{ Federated Data Hyper-Cleaning }

\section{Conclusion}
In the paper, we studied a class of novel distributed bilevel optimization problems that are widely applied many machine learning problems such as
distributed meta learning and federated data hyper-cleaning. Meanwhile, we proposed an effective and efficient adaptive federated bilevel optimization algorithm (i.e., AdaFBiO) to solve
these distributed bilevel optimization problems. Moreover, we provide a convergence analysis framework for our AdaFBiO algorithm,
and prove that it reaches the best known sample and communication complexities simultaneously.
In particular, we uses the unified adaptive matrices to our AdaFBiO algorithm, which can flexibly incorporate various adaptive learning rates to
update variables in both UL and LL problems.


\small

\bibliographystyle{plainnat}

\bibliography{AdaBiFL}

\newpage

\appendix

\section{Appendix}
In this section, we provide the detailed convergence analysis of our algorithms.

We first introduce some useful notations: $\bar{v}_t = \frac{1}{M} \sum_{m=1}^{M} v^m_t$, $\bar{w}_t = \frac{1}{M} \sum_{m=1}^{M} w^m_t$,
$\bar{x}_t = \frac{1}{M} \sum_{m=1}^{M} x^m_t$, $\hat{x}_t = \frac{1}{M}\sum_{m=1}^M\hat{x}^m_t$,
$\bar{y}_t = \frac{1}{M} \sum_{m=1}^{M} y^m_t$,
\begin{align}
& F(x)=\frac{1}{M}\sum_{m=1}^Mf^m(x,y^*(x)), \quad \hat{\nabla} f(x,y) = \frac{1}{M}\sum_{m=1}^{M}\hat{\nabla} f^m(x,y), \quad \nabla_y g(x,y) = \frac{1}{M}\sum_{m=1}^{M}\nabla_y g^m(x,y), \nonumber \\
& \bar{\nabla}f(x_t,y_t)=\frac{1}{M}\sum_{m=1}^M\hat{\nabla }f^m(x^m_t,y^m_t), \quad \bar{\nabla}_y g(x_t,y_t) = \frac{1}{M}\sum_{m=1}^M \nabla_y g^m(x^m_t,y^m_t), \nonumber \\
& R^m_t = \mathbb{E}_{\bar{\xi}^m}\big[\hat{\nabla} f^m(x^m_t,y^m_t;\bar{\xi}^m)\big] -\hat{\nabla} f^m(x^m_t,y^m_t), \quad R_t = \frac{1}{M}\sum_{m=1}^MR^m_t = \frac{1}{M}\sum_{m=1}^MR^m_t(x_t,y_t), \quad \forall t\geq 1. \nonumber
\end{align}

Next, we review and provide some useful lemmas.

\begin{lemma} \label{lem:A1}
Given $M$ vectors $\{u^m\}_{m=1}^M$, the following inequalities satisfy: $||u^m + u^j||^2 \leq (1+c)||u^m||^2 + (1+\frac{1}{c})||u^j||^2$ for any $c > 0$, and
$||\sum_{m=1}^M u^m||^2 \le M\sum_{m=1}^{M} ||u^m||^2$.
\end{lemma}

\begin{lemma} \label{lem:A2}
Given a finite sequence $\{u^{m}\}_{m=1}^M$, and $\bar{u} = \frac{1}{M}\sum_{m=1}^M u^{m}$,
the following inequality satisfies $\sum_{m=1}^M \|u^{m} - \bar{u}\|^2 \leq \sum_{m=1}^M \|u^{m}\|^2$.
\end{lemma}

Given a $\rho$-strongly convex function $\phi(x)$, we define a prox-function (Bregman distance) \cite{censor1981iterative,censor1992proximal}
associated with $\phi(x)$ as follows:
\begin{align}
 D(z,x) = \phi(z) - \big[\phi(x) + \langle\nabla \phi(x),z-x\rangle\big].
\end{align}
Then we define a generalized projection problem as in \cite{ghadimi2016mini}:
\begin{align} \label{eq:A1}
 x^* = \arg\min_{z\in \mathcal{X}} \big\{\langle z,w\rangle + \frac{1}{\gamma}D(z,x) + h(z)\big\},
\end{align}
where $ \mathcal{X} \subseteq \mathbb{R}^d$, $w \in \mathbb{R}^d$ and $\gamma>0$.
Here $h(x)$ is convex and possibly nonsmooth function.

\begin{lemma} \label{lem:A3}
(Lemma 1 of \cite{ghadimi2016mini})
Let $x^*$ be given in \eqref{eq:A1}. Then, for any $x\in \mathcal{X}$, $w\in \mathbb{R}^d$ and $\gamma >0$,
we have
\begin{align}
 \langle w,  \mathcal{G}_{\mathcal{X}}(x,w,\gamma)\rangle \geq \rho \|\mathcal{G}_{\mathcal{X}}(x,w,\gamma)\|^2
 + \frac{1}{\gamma}\big[h(x^*)-h(x)\big],
\end{align}
where $\mathcal{G}_{\mathcal{X}}(x,w,\gamma) = \frac{1}{\gamma}(x-x^*)$, and $\rho>0$ depends on $\rho$-strongly convex function $\phi(x)$.
\end{lemma}
When $h(x)=0$, in the above lemma \ref{lem:A3}, we have
\begin{align} 
 \langle w, \mathcal{G}_{\mathcal{X}}(x,w,\gamma)\rangle \geq \rho \|\mathcal{G}_{\mathcal{X}}(x,w,\gamma)\|^2.
\end{align}

\begin{lemma} \label{lem:A01}
(Restatement of Lemma \ref{lem:01})
Under the above Assumption \ref{ass:1}, we have, for any $x\in \mathbb{R}^d$,
\begin{align}
 &\nabla F(x) = \frac{1}{M}\sum_{m=1}^M\nabla_x f^m(x,y^*(x)) + \nabla y^*(x)^T\Big(\frac{1}{M}\sum_{m=1}^M\nabla_y f^m(x,y^*(x))\Big) \\
 & = \frac{1}{M}\sum_{m=1}^M\nabla_x f^m(x,y^*(x)) - \frac{1}{M}\sum_{m=1}^M\nabla^2_{xy} g^m(x,y^*(x))\Big[\frac{1}{M}\sum_{m=1}^M\nabla^2_{yy}g^m(x,y^*(x))\Big]^{-1}\Big(\frac{1}{M}\sum_{m=1}^M\nabla_y f^m(x,y^*(x))\Big). \nonumber
\end{align}
\end{lemma}

\begin{proof}
Since $F(x)=\frac{1}{M}\sum_{m=1}^Mf^m(x,y^*(x))$, we have 
\begin{align}
\nabla F(x)=\frac{1}{M}\sum_{m=1}^M\nabla_x f^m(x,y^*(x)) + \nabla y^*(x)\Big(\frac{1}{M}\sum_{m=1}^M\nabla_y f^m(x,y^*(x))\Big). 
\end{align}
Meanwhile, according to the optimal condition of the LL subproblem in Problem \eqref{eq:1}, we have 
\begin{align} \label{eq:A2}
 \big(\frac{1}{M}\sum_{m=1}^M\nabla_y g^m(x,y^*(x))\big)=0,
\end{align}
then further differentiating the above equality \eqref{eq:A2} on the variable $x$, we can obtain 
\begin{align}
 \frac{1}{M}\sum_{m=1}^M\nabla_x\nabla_y g^m(x,y^*(x)) + \nabla y^*(x)\big(\frac{1}{M}\sum_{m=1}^M\nabla_y\nabla_y g^m(x,y^*(x))\big)=0. 
\end{align}
According to Assumption \ref{ass:1}, the matrix  $\frac{1}{M}\sum_{m=1}^M\nabla_y\nabla_y g^m(x,y^*(x))$ is reversible. Thus, we have 
\begin{align}
\nabla y^*(x) = - \frac{1}{M}\sum_{m=1}^M\nabla_x\nabla_y g^m(x,y^*(x))\Big(\frac{1}{M}\sum_{m=1}^M\nabla_y\nabla_y g^m(x,y^*(x))\Big)^{-1}.
\end{align}
Then we have 
\begin{align}
\nabla F(x) & =\frac{1}{M}\sum_{m=1}^M\nabla_x f^m(x,y^*(x)) - \frac{1}{M}\sum_{m=1}^M\nabla_x\nabla_y g^m(x,y^*(x))\nonumber \\
& \quad \cdot \Big(\frac{1}{M}\sum_{m=1}^M\nabla_y\nabla_y g^m(x,y^*(x))\Big)^{-1}\Big(\frac{1}{M}\sum_{m=1}^M\nabla_y f^m(x,y^*(x))\Big).
\end{align}

\end{proof}

\begin{lemma} \label{lem:A02}
(Restatement of Lemma \ref{lem:02})
Under the above Assumptions (\ref{ass:1}, \ref{ass:2}, \ref{ass:3}), the global functions (or mappings) $F(x)=\frac{1}{M}\sum_{m=1}^Mf^m(x,y^*(x))$ and $y^*(x)=\arg\min_{y\in \mathbb{R}^p}\frac{1}{M}\sum_{m=1}^Mg^m(x,y)$ satisfy, for all $x_1,x_2\in \mathbb{R}^d$,
\begin{align}
 & \|y^*(x_1)-y^*(x_2)\| \leq \kappa\|x_1-x_2\|, \quad \|\nabla y^*(x_1) - \nabla y^*(x_2)\| \leq L_y\|x_1-x_2\| \nonumber \\
 & \|\nabla F(x_1) - \nabla F(x_2)\|\leq L\|x_1-x_2\|,  \nonumber
\end{align}
where $\kappa=C_{gxy}/\mu$, $L_y=\big( \frac{C_{gxy}L_{gyy}}{\mu^2} +  \frac{L_{gxy}}{\mu} \big) (1+ \frac{C_{gxy}}{\mu})$ and
$L=\Big(L_f + \frac{C_{gxy}L_f}{\mu} + C_{fy}\big( \frac{C_{gxy}L_{gyy}}{\mu^2} +  \frac{L_{gxy}}{\mu} \big)\Big)(1+\frac{C_{gxy}}{\mu})$.
\end{lemma}

\begin{proof}
From the above lemma \ref{lem:A01}, we have 
\begin{align}
\nabla y^*(x) = - \frac{1}{M}\sum_{m=1}^M\nabla^2_{xy}g^m(x,y^*(x))\Big(\frac{1}{M}\sum_{m=1}^M\nabla^2_{yy} g^m(x,y^*(x))\Big)^{-1}.
\end{align}
According to Assumptions \ref{ass:1} and \ref{ass:3}, we have for any $x\in \mathbb{R}^d$, 
\begin{align} 
\|\nabla y^*(x)\| & = \|- \frac{1}{M}\sum_{m=1}^M\nabla^2_{xy} g^m(x,y^*(x))\Big(\frac{1}{M}\sum_{m=1}^M\nabla^2_{yy} g^m(x,y^*(x))\Big)^{-1}\| \nonumber \\
& \leq \| \frac{1}{M}\sum_{m=1}^M\nabla^2_{xy} g^m(x,y^*(x))\|\|\frac{1}{M}\sum_{m=1}^M\nabla^2_{yy} g^m(x,y^*(x))\Big)^{-1}\| \leq \frac{C_{gxy}}{\mu}.
\end{align}
By using the Mean Value Theorem, we have for any $x_1,x_2\in \mathbb{R}^d$
\begin{align} \label{eq:A2}
\|y^*(x_1)-y^*(x_1)\| = \|\nabla y^*(z)(x_1-x_2)\| \leq \frac{C_{gxy}}{\mu}\|x_1-x_2\|, 
\end{align}
where $z\in (x_1,x_2)$. 

For any $x_1,x_2\in \mathbb{R}^d$, we have 
\begin{align} \label{eq:A3}
 & \|\nabla y^*(x_1) - \nabla y^*(x_2)\| \nonumber \\ 
 & = \|\frac{1}{M}\sum_{m=1}^M\nabla^2_{xy} g^m(x_1,y^*(x_1))\Big(\frac{1}{M}\sum_{m=1}^M\nabla^2_{yy} g^m(x_1,y^*(x_1))\Big)^{-1} \nonumber \\
 & \quad -  \frac{1}{M}\sum_{m=1}^M\nabla^2_{xy} g^m(x_2,y^*(x_2))\Big(\frac{1}{M}\sum_{m=1}^M\nabla^2_{yy} g^m(x_2,y^*(x_2))\Big)^{-1}\| \nonumber \\
 & \leq \frac{1}{M}\sum_{m=1}^M\|\nabla^2_{xy} g^m(x_1,y^*(x_1))\Big(\frac{1}{M}\sum_{m=1}^M\nabla^2_{yy} g^m(x_1,y^*(x_1))\Big)^{-1} - \nabla^2_{xy} g^m(x_1,y^*(x_1))\Big(\frac{1}{M}\sum_{m=1}^M\nabla^2_{yy} g^m(x_2,y^*(x_2))\Big)^{-1}\| \nonumber \\
 & \quad + \frac{1}{M}\sum_{m=1}^M\|\nabla^2_{xy} g^m(x_1,y^*(x_1))\Big(\frac{1}{M}\sum_{m=1}^M\nabla^2_{yy} g^m(x_2,y^*(x_2))\Big)^{-1} -  \nabla^2_{xy} g^m(x_2,y^*(x_2))\Big(\frac{1}{M}\sum_{m=1}^M\nabla^2_{yy} g^m(x_2,y^*(x_2))\Big)^{-1}\| \nonumber \\
 & \leq \frac{C_{gxy}}{M}\sum_{m=1}^M\|\Big(\frac{1}{M}\sum_{m=1}^M\nabla^2_{yy} g^m(x_1,y^*(x_1))\Big)^{-1} - \Big(\frac{1}{M}\sum_{m=1}^M\nabla^2_{yy} g^m(x_2,y^*(x_2))\Big)^{-1}\| \nonumber \\
 & \quad + \frac{1}{M\mu}\sum_{m=1}^M\|\nabla^2_{xy} g^m(x_1,y^*(x_1))-  \nabla^2_{xy} g^m(x_2,y^*(x_2))\| \nonumber \\
 & \leq \frac{C_{gxy}}{M}\sum_{m=1}^M\|\Big(\frac{1}{M}\sum_{m=1}^M\nabla^2_{yy} g^m(x_2,y^*(x_2))\Big)^{-1}\Big(\frac{1}{M}\sum_{m=1}^M\nabla^2_{yy} g^m(x_2,y^*(x_2)) - \frac{1}{M}\sum_{m=1}^M\nabla^2_{yy} g^m(x_1,y^*(x_1))\Big) \nonumber \\
 & \quad \cdot \Big(\frac{1}{M}\sum_{m=1}^M\nabla^2_{yy} g^m(x_1,y^*(x_1))\Big)^{-1}\| 
 + \frac{1}{M\mu}\sum_{m=1}^M\Big( L_{gxy}\|x_1-x_2\| + L_{gxy}\|y^*(x_1)-y^*(x_2)\|\Big) \nonumber \\
 & \leq \frac{C_{gxy}}{M\mu^2}\sum_{m=1}^M\|\frac{1}{M}\sum_{m=1}^M\big(\nabla^2_{yy} g^m(x_2,y^*(x_2)) - 
 \nabla^2_{yy} g^m(x_1,y^*(x_1))\big)\|
 + \frac{1}{M\mu}\sum_{m=1}^M\Big( L_{gxy}\|x_1-x_2\| + L_{gxy}\|y^*(x_1)-y^*(x_2)\|\Big) \nonumber \\
 & \leq \frac{C_{gxy}}{M\mu^2}\sum_{m=1}^M\frac{1}{M}\sum_{m=1}^M\big(L_{gyy}\|x_1-x_2\| + L_{gyy}\|y^*(x_1)-y^*(x_2)\|\big) + \frac{1}{M\mu}\sum_{m=1}^M\Big( L_{gxy}\|x_1-x_2\| + L_{gxy}\|y^*(x_1)-y^*(x_2)\|\Big) \nonumber \\
 & \leq \big( \frac{C_{gxy}L_{gyy}}{\mu^2} +  \frac{L_{gxy}}{\mu} \big) (1+ \frac{C_{gxy}}{\mu})\|x_1-x_2\|, 
\end{align}
where the last inequality holds by the above inequality (\ref{eq:A2}). 

From the above lemma \ref{lem:A01}, we have 
\begin{align}
 \nabla F(x) = \frac{1}{M}\sum_{m=1}^M\nabla_x f^m(x,y^*(x)) + \nabla y^*(x)^T\Big(\frac{1}{M}\sum_{m=1}^M\nabla_y f^m(x,y^*(x))\Big),
\end{align}
then for any $x_1,x_2\in \mathbb{R}^d$, we have 
\begin{align}
 \|\nabla F(x_1) -\nabla F(x_2)\| & = \|\frac{1}{M}\sum_{m=1}^M\nabla_x f^m(x_1,y^*(x_1)) + \nabla y^*(x_1)^T\Big(\frac{1}{M}\sum_{m=1}^M\nabla_y f^m(x_1,y^*(x_1))\Big) \nonumber \\
 & \quad - \frac{1}{M}\sum_{m=1}^M\nabla_x f^m(x_1,y^*(x_2)) - \nabla y^*(x_2)^T\Big(\frac{1}{M}\sum_{m=1}^M\nabla_y f^m(x_2,y^*(x_2))\Big) \| \nonumber \\
 & \leq L_f\big(\|x_1-x_2\|+\|y^*(x_1)-y^*(x_2)\|\big) + \|\nabla y^*(x_1)^T\Big(\frac{1}{M}\sum_{m=1}^M\nabla_y f^m(x_1,y^*(x_1))\Big) \nonumber \\
 & \quad - \nabla y^*(x_2)^T\Big(\frac{1}{M}\sum_{m=1}^M\nabla_y f^m(x_1,y^*(x_1))\Big)  + \nabla y^*(x_2)^T\Big(\frac{1}{M}\sum_{m=1}^M\nabla_y f^m(x_1,y^*(x_1))\Big)  \nonumber \\
 & \quad - \nabla y^*(x_2)^T\Big(\frac{1}{M}\sum_{m=1}^M\nabla_y f^m(x_2,y^*(x_2))\Big)\| \nonumber \\
 & \leq L_f\big(\|x_1-x_2\|+\|y^*(x_1)-y^*(x_2)\|\big) + C_{fy}\|\nabla y^*(x_1)-\nabla y^*(x_2)\| \nonumber \\
 & \quad + \frac{C_{gxy}}{\mu}\frac{1}{M}\sum_{m=1}^M\|\nabla_y f^m(x_1,y^*(x_1)) - \nabla_y f^m(x_2,y^*(x_2))\| \nonumber \\
 & \leq L_f(1+\frac{C_{gxy}}{\mu})\|x_1-x_2\| + C_{fy}\|\nabla y^*(x_1)-\nabla y^*(x_2)\| + \frac{C_{gxy}L_f}{\mu}(1+\frac{C_{gxy}}{\mu})\|x_1-x_2\| \nonumber \\
 & \leq \Big(L_f + \frac{C_{gxy}L_f}{\mu} + C_{fy}\big( \frac{C_{gxy}L_{gyy}}{\mu^2} +  \frac{L_{gxy}}{\mu} \big)\Big)(1+\frac{C_{gxy}}{\mu})\|x_1-x_2\|,
\end{align}
where the last inequality holds by (\ref{eq:A3}). 

\end{proof}

\begin{lemma} \label{lem:A4}
(Restatement of Lemma \ref{lem:5})
Assume the sequences $\big\{\bar{w}_t,\bar{x}_t,\bar{y}_t\big\}_{t=1}^T$ generated from Algorithm \ref{alg:1},
 we have
 \begin{align}
 \|\bar{w}_t-\nabla F(\bar{x}_t)\|^2 \leq 8\bar{L}^2\|y^*(\bar{x}_t)-\bar{y}_t\|^2 + 2\|\bar{w}_t-\hat{\nabla} f(\bar{x}_t,\bar{y}_t)\|^2,
\end{align}
where $\bar{L}^2= L^2_f+ \frac{L^2_{gxy}C^2_{fy}}{\mu^2} + \frac{L^2_{gyy} C^2_{gxy}C^2_{fy}}{\mu^4} +
 \frac{L^2_fC^2_{gxy}}{\mu^2}$.
\end{lemma}

\begin{proof}
 We first consider the term $\|\nabla F(\bar{x}_t)-\hat{\nabla} f(\bar{x}_t,\bar{y}_t)\|^2$. Since $\nabla F(\bar{x}_t)=\frac{1}{M} \sum_{m=1}^{M}\nabla f^m(\bar{x}_t,y^*(\bar{x}_t))$
 and $\hat{\nabla} f(\bar{x}_t,\bar{y}_t) = \frac{1}{M}\sum_{m=1}^{M}\hat{\nabla} f^m(\bar{x}_t,\bar{y}_t)$, we have
\begin{align}
& \|\nabla F(\bar{x}_t) - \hat{\nabla} f(\bar{x}_t,\bar{y}_t)\|^2 \nonumber \\
 & = \|\frac{1}{M}\sum_{m=1}^{M}\nabla f^m(\bar{x}_t,y^*(\bar{x}_t))-\frac{1}{M}\sum_{m=1}^{M}\hat{\nabla} f^m(\bar{x}_t,\bar{y}_t)\|^2 \nonumber \\
 & \leq \frac{1}{M}\sum_{m=1}^{M}\|\nabla f^m(\bar{x}_t,y^*(\bar{x}_t)) - \hat{\nabla} f^m(\bar{x}_t,\bar{y}_t)\|^2 \nonumber \\
 & = \frac{1}{M}\sum_{m=1}^{M}\|\nabla_xf^m(\bar{x}_t,y^*(\bar{x}_t)) - \nabla^2_{xy}g^m(\bar{x}_t,y^*(\bar{x}_t)) \big(\nabla^2_{yy}g^m(\bar{x}_t,y^*(\bar{x}_t))\big)^{-1}\nabla_yf^m(\bar{x}_t,y^*(x)) \nonumber \\
 & \quad  - \nabla_xf^m(\bar{x}_t,\bar{y}_t) + \nabla^2_{xy}g^m(\bar{x}_t,\bar{y}_t) \big(\nabla^2_{yy}g^m(\bar{x}_t,\bar{y}_t)\big)^{-1}\nabla_yf^m(\bar{x}_t,\bar{y}_t)\|^2 \nonumber \\
 & = \frac{1}{M}\sum_{m=1}^{M}\|\nabla_xf^m(\bar{x}_t,y^*(\bar{x}_t)) - \nabla_xf^m(\bar{x}_t,\bar{y}_t) - \nabla^2_{xy}g^m(\bar{x}_t,y^*(\bar{x}_t)) \big(\nabla^2_{yy}g^m(\bar{x}_t,y^*(\bar{x}_t))\big)^{-1}\nabla_yf^m(\bar{x}_t,y^*(\bar{x}_t)) \nonumber \\
 & \quad  + \nabla^2_{xy}g^m(\bar{x}_t,\bar{y}_t) \big(\nabla^2_{yy}g^m(\bar{x}_t,y^*(\bar{x}_t))\big)^{-1}\nabla_yf^m(\bar{x}_t,y^*(\bar{x}_t)) - \nabla^2_{xy}g^m(\bar{x}_t,\bar{y}_t) \big(\nabla^2_{yy}g^m(\bar{x}_t,y^*(\bar{x}_t))\big)^{-1}\nabla_yf^m(\bar{x}_t,y^*(\bar{x}_t)) \nonumber \\
 & \quad + \nabla^2_{xy}g^m(\bar{x}_t,\bar{y}_t) \big(\nabla^2_{yy}g^m(\bar{x}_t,\bar{y}_t)\big)^{-1}\nabla_yf^m(\bar{x}_t,y^*(\bar{x}_t)) -  \nabla^2_{xy}g^m(\bar{x}_t,\bar{y}_t) \big(\nabla^2_{yy}g^m(\bar{x}_t,\bar{y}_t)\big)^{-1}\nabla_yf^m(\bar{x}_t,y^*(\bar{x}_t)) \nonumber \\
 & \quad + \nabla^2_{xy}g^m(\bar{x}_t,\bar{y}_t) \big(\nabla^2_{yy}g^m(\bar{x}_t,\bar{y}_t)\big)^{-1}\nabla_yf^m(\bar{x}_t,\bar{y}_t)\|^2 \nonumber \\
 & \leq \frac{1}{M}\sum_{m=1}^{M} \bigg(4\|\nabla_xf^m(\bar{x}_t,y^*(\bar{x}_t)) - \nabla_xf^m(\bar{x}_t,\bar{y}_t)\|^2 + \frac{4C^2_{fy}}{\mu^2}\|\nabla^2_{xy}g^m(\bar{x}_t,y^*(\bar{x}_t))-\nabla^2_{xy}g^m(\bar{x}_t,\bar{y}_t)\|^2
 \nonumber \\
 & \quad + \frac{4C^2_{gxy}C^2_{fy}}{\mu^4}\|\nabla^2_{yy}g^m(\bar{x}_t,y^*(\bar{x}_t))-\nabla^2_{yy}g^m(\bar{x}_t,\bar{y}_t)\|^2 +
 \frac{4C^2_{gxy}}{\mu^2}\|\nabla_yf^m(\bar{x}_t,y^*(\bar{x}_t))-\nabla_yf^m(\bar{x}_t,\bar{y}_t)\|^2 \bigg) \nonumber \\
 & \leq 4\big(L^2_f+ \frac{L^2_{gxy}C^2_{fy}}{\mu^2} + \frac{L^2_{gyy} C^2_{gxy}C^2_{fy}}{\mu^4} +
 \frac{L^2_fC^2_{gxy}}{\mu^2}\big)\|y^*(\bar{x}_t)-\bar{y}_t\|^2 \nonumber \\
 & = 4\bar{L}^2\|y^*(\bar{x}_t)-\bar{y}_t\|^2,
\end{align}
where the second last inequality is due to Assumptions \ref{ass:1}, \ref{ass:2} and \ref{ass:4};
the last equality holds by $ \bar{L}^2=L^2_f+ \frac{L^2_{gxy}C^2_{fy}}{\mu^2} + \frac{L^2_{gyy} C^2_{gxy}C^2_{fy}}{\mu^4} +
 \frac{L^2_fC^2_{gxy}}{\mu^2}$.

Then we have
\begin{align}
 \|\bar{w}_t-\nabla F(\bar{x}_t)\|^2 & = \|\bar{w}_t-\hat{\nabla} f(\bar{x}_t,\bar{y}_t) + \hat{\nabla} f(\bar{x}_t,\bar{y}_t)-\nabla F(\bar{x}_t)\|^2 \nonumber \\
 & \leq 2\|\bar{w}_t-\hat{\nabla} f(\bar{x}_t,\bar{y}_t)\|^2 + 2\|\hat{\nabla} f(\bar{x}_t,\bar{y}_t)-\nabla F(\bar{x}_t)\|^2 \nonumber \\
 & \leq 2\|\bar{w}_t-\hat{\nabla} f(\bar{x}_t,\bar{y}_t)\|^2 + 8\bar{L}^2\|y^*(\bar{x}_t)-\bar{y}_t\|^2.
\end{align}

\end{proof}

\begin{lemma} \label{lem:A5}
 Suppose that the sequence $\big\{\bar{x}_t,\hat{x}_t\big\}_{t=1}^T$ be generated from Algorithm \ref{alg:1}, where $\bar{x}_t = \frac{1}{M}\sum_{m=1}^Mx^m_t$,
 $\hat{x}_t = \frac{1}{M}\sum_{m=1}^M\hat{x}^m_t$.
 Let $0<\eta_t \leq 1$ and $0< \gamma \leq \frac{\rho}{2L\eta_t}$,
 then we have
 \begin{align}
  F(\bar{x}_{t+1}) \leq F(\bar{x}_t) + \frac{8\bar{L}^2\gamma\eta_t}{\rho}\|y^*(\bar{x}_t)-\bar{y}_t\|^2 +\frac{2\gamma\eta_t}{\rho}\|\bar{w}_t-\hat{\nabla} f(\bar{x}_t,\bar{y}_t)\|^2 -\frac{\rho\eta_t}{2\gamma}\|\hat{x}_{t+1}-\bar{x}_t\|^2.
 \end{align}
\end{lemma}

\begin{proof}
For notational simplicity, let $s_t=q\lfloor t/q\rfloor+1$. When $t=s_t$, according to the line 8 of Algorithm \ref{alg:1},
we have $\bar{x}_{t+1} = \bar{x}_t + \eta_t(\hat{x}_{t+1}-\bar{x}_t)$.
When $t\in (s_t,s_t+q)$, according to the line 12 of Algorithm \ref{alg:1}, we have $\bar{x}_{t+1}=\frac{1}{M}\sum_{m=1}^Mx^m_{t+1} = \frac{1}{M}\sum_{m=1}^M\big(x^m_t + \eta_t(\hat{x}^m_{t+1}-x^m_t)\big)
= \bar{x}_t + \eta_t(\hat{x}_{t+1}-\bar{x}_t)$.

According to the above Lemma \ref{lem:02}, the function $F(x)$ is $L$-smooth. Thus we have
 \begin{align} \label{eq:B1}
  F(\bar{x}_{t+1}) & \leq F(\bar{x}_t) + \langle\nabla F(\bar{x}_t), \bar{x}_{t+1}-\bar{x}_t\rangle + \frac{L}{2}\|\bar{x}_{t+1}-\bar{x}_t\|^2 \\
  & = F(\bar{x}_t)+ \langle \nabla F(\bar{x}_t),\eta_t(\hat{x}_{t+1}-\bar{x}_t)\rangle + \frac{L}{2}\|\eta_t(\hat{x}_{t+1}-\bar{x}_t)\|^2 \nonumber \\
  & = F(\bar{x}_t) + \eta_t\underbrace{\langle \bar{w}_t,\hat{x}_{t+1}-\bar{x}_t\rangle}_{=T_1} + \eta_t\underbrace{\langle \nabla F(\bar{x}_t)-\bar{w}_t,\hat{x}_{t+1}-\bar{x}_t\rangle}_{=T_2} + \frac{L\eta_t^2}{2}\|\hat{x}_{t+1}-\bar{x}_t\|^2, \nonumber
 \end{align}
where the second equality is due to $\bar{x}_{t+1}=\bar{x}_t + \eta_t(\hat{x}_{t+1}-\bar{x}_t)$.

According to Assumption \ref{ass:6}, i.e., $A_t\succ \rho I_d$ for any $t\geq 1$,
the mirror function $\phi_t(x)=\frac{1}{2}x^TA_t x$ is $\rho$-strongly convex, then we can define a Bregman distance as in \cite{ghadimi2016mini},
\begin{align}
 D_t(x,\bar{x}_t) = \phi_t(x) - \big[ \phi_t(\bar{x}_t) + \langle\nabla \phi_t(\bar{x}_t), x-\bar{x}_t\rangle\big] = \frac{1}{2}(x-\bar{x}_t)^TA_t(x-\bar{x}_t).
\end{align}
When $t=s_t$, according to the line 7 of Algorithm \ref{alg:1}, we have
$\hat{x}_{t+1}=\bar{x}_t-\gamma A^{-1}_t\bar{w}_t=\arg\min_{x\in \mathbb{R}^d}\big\{\langle \bar{w}_t, x\rangle + \frac{1}{2\gamma}(x-\bar{x}_t)^T A_t (x-\bar{x}_t)\big\}$.
When $t\in (s_t,s_t+q)$, according to the line 11 of Algorithm \ref{alg:1}, we have $\hat{x}_{t+1}=\frac{1}{M}\sum_{m=1}^M\hat{x}^m_{t+1} = \frac{1}{M}\sum_{m=1}^M\big(x^m_t - \gamma A_t^{-1} w^m_t\big)
=\bar{x}_t -\gamma A_t^{-1} \bar{w}_t =\arg\min_{x\in \mathbb{R}^d}\big\{\langle \bar{w}_t, x\rangle + \frac{1}{2\gamma}(x-\bar{x}_t)^T A_t (x-\bar{x}_t)\big\}$.

By using Lemma 1 in \cite{ghadimi2016mini} to the problem $\hat{x}_{t+1} = \arg\min_{x \in \mathbb{R}^d} \big\{\langle \bar{w}_t, x\rangle + \frac{1}{2\gamma}(x-\bar{x}_t)^T A_t (x-\bar{x}_t)\big\}$, we can obtain
\begin{align}
  \langle \bar{w}_t, \frac{1}{\gamma}(\bar{x}_t - \hat{x}_{t+1})\rangle \geq \rho\|\frac{1}{\gamma}(\bar{x}_t - \hat{x}_{t+1})\|^2.
\end{align}
Then we have
\begin{align} \label{eq:B3}
  T_1=\langle \bar{w}_t, \hat{x}_{t+1}-\bar{x}_t\rangle \leq -\frac{\rho }{\gamma }\|\hat{x}_{t+1}-\bar{x}_t\|^2.
\end{align}

Next, consider the bound of the term $T_2$, we have
\begin{align} \label{eq:B4}
  T_2 & = \langle \nabla F(\bar{x}_t)-\bar{w}_t,\hat{x}_{t+1}-\bar{x}_t\rangle \nonumber \\
  & \leq \|\nabla F(\bar{x}_t)-\bar{w}_t\|\cdot\|\hat{x}_{t+1}-\bar{x}_t\| \nonumber \\
  & \leq \frac{\gamma}{\rho}\|\nabla F(\bar{x}_t)-\bar{w}_t\|^2+\frac{\rho}{4\gamma}\|\hat{x}_{t+1}-\bar{x}_t\|^2,
\end{align}
where the first inequality is due to the Cauchy-Schwarz inequality and the last is due to Young's inequality.
By combining the above inequalities \eqref{eq:B1}, \eqref{eq:B3} with \eqref{eq:B4},
we obtain
\begin{align} \label{eq:B5}
  F(\bar{x}_{t+1}) &\leq F(\bar{x}_t) + \eta_t\langle \nabla F(\bar{x}_t)-\bar{w}_t,\hat{x}_{t+1}-\bar{x}_t\rangle + \eta_t\langle \bar{w}_t,\hat{x}_{t+1}-\bar{x}_t\rangle + \frac{L\eta_t^2}{2}\|\hat{x}_{t+1}-\bar{x}_t\|^2  \nonumber \\
  & \leq F(\bar{x}_t) + \frac{\eta_t\gamma}{\rho}\|\nabla F(\bar{x}_t)-\bar{w}_t\|^2 + \frac{\rho\eta_t}{4\gamma}\|\hat{x}_{t+1}-\bar{x}_t\|^2  -\frac{\rho\eta_t}{\gamma}\|\hat{x}_{t+1}-\bar{x}_t\|^2 + \frac{L\eta_t^2}{2}\|\hat{x}_{t+1}-\bar{x}_t\|^2 \nonumber \\
  & = F(\bar{x}_t) + \frac{\eta_t\gamma}{\rho}\|\nabla F(\bar{x}_t)-\bar{w}_t\|^2 -\frac{\rho\eta_t}{2\gamma}\|\hat{x}_{t+1}-\bar{x}_t\|^2  -\big(\frac{\rho\eta_t}{4\gamma}-\frac{L\eta_t^2}{2}\big)\|\hat{x}_{t+1}-\bar{x}_t\|^2 \nonumber \\
  & \leq F(\bar{x}_t) + \frac{\eta_t\gamma}{\rho}\|\nabla F(\bar{x}_t)-\bar{w}_t\|^2 -\frac{\rho\eta_t}{2\gamma}\|\hat{x}_{t+1}-\bar{x}_t\|^2 \nonumber \\
  & \leq F(\bar{x}_t) + \frac{8\bar{L}^2\gamma\eta_t}{\rho}\|y^*(\bar{x}_t)-\bar{y}_t\|^2 +\frac{2\gamma\eta_t}{\rho}\|\bar{w}_t-\hat{\nabla} f(\bar{x}_t,\bar{y}_t)\|^2 -\frac{\rho\eta_t}{2\gamma}\|\hat{x}_{t+1}-\bar{x}_t\|^2,
\end{align}
where the second last inequality is due to $0< \gamma \leq \frac{\rho}{2L\eta_t}$, and the last inequality holds by the above Lemma \ref{lem:A4}.

\end{proof}

\begin{lemma} \label{lem:A6}
Suppose the sequence $\big\{\bar{x}_t,\bar{y}_t\big\}_{t=1}^T$ be generated from Algorithm \ref{alg:1}.
Under the above assumptions, given $0< \eta_t\leq 1$, $B_t=b_tI_p \ (\hat{b} \geq b_t \geq \rho>0)$ for all $t\geq 1$,
and $0<\lambda\leq \frac{\rho}{6L_g}$, we have
\begin{align}
  \|\bar{y}_{t+1} - y^*(\bar{x}_{t+1})\|^2 &\leq (1-\frac{\eta_t\mu\lambda}{4b_t})\|\bar{y}_t -y^*(\bar{x}_t)\|^2 -\frac{3\eta_t}{4} \|\hat{y}_{t+1}-\bar{y}_t\|^2 \nonumber \\
     & \quad + \frac{25\eta_t\lambda}{6\mu b_t} \|\nabla_y g(\bar{x}_t,\bar{y}_t)-\bar{v}_t\|^2 + \frac{25\kappa^2\eta_tb_t}{6\mu\lambda}\|\hat{x}_{t+1} - \bar{x}_t\|^2,
\end{align}
where $\kappa = L_g/\mu$.
\end{lemma}

\begin{proof}
According to Assumption \ref{ass:1}, for any $m\in [M]$ and $\zeta^m$, the function is $L_g$-smooth and $\mu$-strongly convex function on variable $y$,
i.e., $L_g I_p\succeq \nabla^2_{yy}g^m(x,y;\zeta^m) \succeq \mu I_p$. Then we have $L_g I_p\succeq \nabla^2_{yy}g^m(x,y) = \mathbb{E}\big[\nabla^2_{yy}g^m(x,y;\zeta^m)\big] \succeq \mu I_p$ for any
$m\in [M]$, and $L_g I_p\succeq \frac{1}{M}\sum_{m=1}^M\nabla^2_{yy}g^m(x,y) \succeq \mu I_p$.
In other words, for any $m\in [M]$, $g^m(x,y)$ also is $L_g$-smooth and $\mu$-strongly convex w.r.t $y$.
Meanwhile, $g(x,y)=\frac{1}{M}\sum_{m=1}^Mg^m(x,y)$ also is $L_g$-smooth and $\mu$-strongly convex w.r.t $y$.
Thus, we have
\begin{align} \label{eq:F1}
 g(\bar{x}_t,y) & \geq g(\bar{x}_t,\bar{y}_t) + \langle\nabla_y g(\bar{x}_t,\bar{y}_t), y-\bar{y}_t\rangle + \frac{\mu}{2}\|y-\bar{y}_t\|^2 \nonumber \\
 & = g(\bar{x}_t,\bar{y}_t) + \langle \bar{v}_t, y-\hat{y}_{t+1}\rangle + \langle\nabla_y g(\bar{x}_t,\bar{y}_t)-\bar{v}_t, y-\hat{y}_{t+1}\rangle \nonumber \\
 & \quad +\langle\nabla_y g(\bar{x}_t,\bar{y}_t), \hat{y}_{t+1}-\bar{y}_t\rangle + \frac{\mu}{2}\|y-\bar{y}_t\|^2.
\end{align}
According to the Assumption \ref{ass:2}, i.e., the function $g(x,y)$ is $L_g$-smooth, we have
\begin{align} \label{eq:F2}
 g(\bar{x}_t,\hat{y}_{t+1}) \leq g(\bar{x}_t,y_{t}) + \langle\nabla_y g(\bar{x}_t,\bar{y}_t), \hat{y}_{t+1}-\bar{y}_t\rangle + \frac{L_g}{2}\|\hat{y}_{t+1}-\bar{y}_t\|^2 .
\end{align}
Combining the about inequalities \eqref{eq:F1} with \eqref{eq:F2}, we have
\begin{align} \label{eq:F3}
 g(\bar{x}_t,y) & \geq g(\bar{x}_t,\hat{y}_{t+1}) + \langle \bar{v}_t, y-\hat{y}_{t+1}\rangle + \langle\nabla_y g(\bar{x}_t,\bar{y}_t)-\bar{v}_t, y-\hat{y}_{t+1}\rangle \nonumber \\
 & \quad + \frac{\mu}{2}\|y-\bar{y}_t\|^2 - \frac{L_g}{2}\|\hat{y}_{t+1}-\bar{y}_t\|^2.
\end{align}

When $t=s_t=q\lfloor t/q\rfloor+1$, according to the line 7 of Algorithm \ref{alg:1}, we have
$\hat{y}_{t+1}=\bar{y}_t - \lambda B^{-1}_t\bar{v}_t=\arg\min_{y\in \mathbb{R}^p} \Big\{\langle \bar{v}_t, y\rangle + \frac{1}{2\lambda}(y-\bar{y}_t)^T B_t (y-\bar{y}_t)\Big\}$.
When $t\in (s_t,s_t+q)$, according to the line 11 of Algorithm \ref{alg:1}, we have $\hat{y}_{t+1}=\frac{1}{M}\sum_{m=1}^M\hat{y}^m_{t+1} = \frac{1}{M}\sum_{m=1}^M\big(y^m_t - \lambda B_t^{-1} v^m_t\big)
=\bar{y}_t -\lambda B_t^{-1} \bar{v}_t =\arg\min_{y\in \mathbb{R}^p}\Big\{\langle \bar{v}_t, y\rangle + \frac{1}{2\lambda}(y-\bar{y}_t)^T B_t (y-\bar{y}_t)\Big\}$.

By the optimality of the problem $\hat{y}_{t+1}=\arg\min_{y\in \mathbb{R}^p}\Big\{\langle \bar{v}_t, y\rangle + \frac{1}{2\lambda}(y-\bar{y}_t)^T B_t (y-\bar{y}_t)\Big\}$,
given $B_t=b_tI_p \ (b_t\geq \rho >0)$, we have
\begin{align}
  \langle \bar{v}_t + \frac{b_t}{\lambda}(\hat{y}_{t+1} - \bar{y}_t), y-\hat{y}_{t+1} \rangle \geq 0, \quad \forall y \in \mathbb{R}^p.
\end{align}
Then we obtain
\begin{align} \label{eq:F4}
  \langle \bar{v}_t, y-\hat{y}_{t+1}\rangle & \geq \frac{b_t}{\lambda}\langle \hat{y}_{t+1}- \bar{y}_t, \hat{y}_{t+1}-y\rangle  \nonumber \\
  & = \frac{b_t}{\lambda}\|\hat{y}_{t+1}- \bar{y}_t\|^2 + \frac{b_t}{\lambda}\langle \hat{y}_{t+1}- \bar{y}_t, \bar{y}_t- y\rangle.
\end{align}

By pugging the inequalities \eqref{eq:F4} into \eqref{eq:F3}, we have
\begin{align}
 g(\bar{x}_t,y) & \geq g(\bar{x}_t,\hat{y}_{t+1}) + \frac{b_t}{\lambda}\langle \hat{y}_{t+1}- \bar{y}_t, \bar{y}_t- y\rangle + \frac{b_t}{\lambda}\|\hat{y}_{t+1}- \bar{y}_t\|^2 \nonumber \\
 & \quad + \langle\nabla_y g(\bar{x}_t,\bar{y}_t)-\bar{v}_t, y-\hat{y}_{t+1}\rangle +\frac{\mu}{2}\|y-\bar{y}_t\|^2 -\frac{L_g}{2}\|\hat{y}_{t+1}-\bar{y}_t\|^2.
\end{align}
Let $y=y^*(\bar{x}_t)$, then we have
\begin{align}
 g(\bar{x}_t,y^*(\bar{x}_t)) & \geq g(\bar{x}_t,\hat{y}_{t+1}) + \frac{b_t}{\lambda}\langle \hat{y}_{t+1}- \bar{y}_t, \bar{y}_t - y^*(\bar{x}_t)\rangle
  + (\frac{b_t}{\lambda}-\frac{L_g}{2})\|\hat{y}_{t+1}- \bar{y}_t\|^2 \nonumber \\
 & \quad + \langle\nabla_y g(\bar{x}_t,\bar{y}_t)-\bar{v}_t, y^*(\bar{x}_t)-\hat{y}_{t+1}\rangle + \frac{\mu}{2}\|y^*(\bar{x}_t)-\bar{y}_t\|^2.
\end{align}
Due to the strongly-convexity of $g(\cdot,y)$ and $y^*(\bar{x}_t) =\arg\min_{y\in \mathbb{R}^p} g(\bar{x}_t,y)$, we have $g(\bar{x}_t,y^*(\bar{x}_t)) \leq g(\bar{x}_t,\hat{y}_{t+1})$.
Thus, we obtain
\begin{align} \label{eq:F5}
 0 & \geq  \frac{b_t}{\lambda}\langle \hat{y}_{t+1}- \bar{y}_t, \bar{y}_t - y^*(\bar{x}_t)\rangle
 + \langle\nabla_y g(\bar{x}_t,\bar{y}_t)-\bar{v}_t, y^*(\bar{x}_t)-\hat{y}_{t+1}\rangle \nonumber \\
 & \quad + (\frac{b_t}{\lambda}-\frac{L_g}{2})\|\hat{y}_{t+1}- \bar{y}_t\|^2 + \frac{\mu}{2}\|y^*(\bar{x}_t)-\bar{y}_t\|^2.
\end{align}

According to $\bar{y}_{t+1} = \bar{y}_t + \eta_t(\hat{y}_{t+1}-\bar{y}_t) $, we have
\begin{align}
\|\bar{y}_{t+1}-y^*(\bar{x}_t)\|^2
& = \|\bar{y}_t + \eta_t(\hat{y}_{t+1}-\bar{y}_t) -y^*(\bar{x}_t)\|^2 \nonumber \\
& =  \|\bar{y}_t -y^*(\bar{x}_t)\|^2 + 2\eta_t\langle \hat{y}_{t+1}-\bar{y}_t, \bar{y}_t -y^*(\bar{x}_t)\rangle + \eta_t^2 \|\hat{y}_{t+1}-\bar{y}_t\|^2.
\end{align}
Then we obtain
\begin{align} \label{eq:F6}
 \langle \hat{y}_{t+1}-\bar{y}_t, \bar{y}_t - y^*(\bar{x}_t)\rangle = \frac{1}{2\eta_t}\|\bar{y}_{t+1}-y^*(\bar{x}_t)\|^2 - \frac{1}{2\eta_t}\|\bar{y}_t -y^*(\bar{x}_t)\|^2 - \frac{\eta_t}{2}\|\hat{y}_{t+1}-\bar{y}_t\|^2.
\end{align}
Consider the upper bound of the term $\langle\nabla_y g(\bar{x}_t,\bar{y}_t)-\bar{v}_t, y^*(\bar{x}_t)-\hat{y}_{t+1}\rangle$, we have
\begin{align} \label{eq:F7}
 &\langle\nabla_y g(\bar{x}_t,\bar{y}_t)-\bar{v}_t, y^*(\bar{x}_t)-\hat{y}_{t+1}\rangle \nonumber \\
 & = \langle\nabla_y g(\bar{x}_t,\bar{y}_t)-\bar{v}_t, y^*(\bar{x}_t)-\bar{y}_t\rangle + \langle\nabla_y g(\bar{x}_t,\bar{y}_t)-\bar{v}_t, \bar{y}_t-\hat{y}_{t+1}\rangle \nonumber \\
 & \geq -\frac{1}{\mu} \|\nabla_y g(\bar{x}_t,\bar{y}_t)-\bar{v}_t\|^2 - \frac{\mu}{4}\|y^*(\bar{x}_t)-\bar{y}_t\|^2 - \frac{1}{\mu} \|\nabla_y g(\bar{x}_t,\bar{y}_t)-\bar{v}_t\|^2 - \frac{\mu}{4}\|\bar{y}_t-\hat{y}_{t+1}\|^2 \nonumber \\
 & = -\frac{2}{\mu} \|\nabla_y g(\bar{x}_t,\bar{y}_t)-\bar{v}_t\|^2 - \frac{\mu}{4}\|y^*(\bar{x}_t)-\bar{y}_t\|^2 - \frac{\mu}{4}\|\bar{y}_t-\hat{y}_{t+1}\|^2.
\end{align}

By plugging the inequalities \eqref{eq:F6} and \eqref{eq:F7} into \eqref{eq:F5},
we obtain
\begin{align}
 & \frac{b_t}{2\eta_t\lambda}\|\bar{y}_{t+1}-y^*(\bar{x}_t)\|^2 \nonumber \\
 & \leq (\frac{b_t}{2\eta_t\lambda}-\frac{\mu}{4})\|\bar{y}_t -y^*(\bar{x}_t)\|^2 + ( \frac{b_t\eta_t}{2\lambda} + \frac{\mu}{4} + \frac{L_g}{2}-\frac{b_t}{\lambda}) \|\hat{y}_{t+1}-\bar{y}_t\|^2 + \frac{2}{\mu} \|\nabla_y g(\bar{x}_t,\bar{y}_t)-\bar{v}_t\|^2 \nonumber \\
 & \leq ( \frac{b_t}{2\eta_t\lambda}-\frac{\mu}{4})\|\bar{y}_t -y^*(\bar{x}_t)\|^2 + (\frac{3L_g}{4} -\frac{b_t}{2\lambda}) \|\hat{y}_{t+1}-\bar{y}_t\|^2 + \frac{2}{\mu} \|\nabla_y g(\bar{x}_t,\bar{y}_t)-\bar{v}_t\|^2 \nonumber \\
 & = ( \frac{b_t}{2\eta_t\lambda}-\frac{\mu}{4})\|\bar{y}_t -y^*(\bar{x}_t)\|^2 - \big( \frac{3b_t}{8\lambda} + \frac{b_t}{8\lambda} -\frac{3L_g}{4}\big) \|\hat{y}_{t+1}-\bar{y}_t\|^2 + \frac{2}{\mu} \|\nabla_y g(\bar{x}_t,\bar{y}_t)-\bar{v}_t\|^2 \nonumber \\
 & \leq  (\frac{b_t}{2\eta_t\lambda}-\frac{\mu}{4})\|\bar{y}_t -y^*(\bar{x}_t)\|^2 - \frac{3b_t}{8\lambda} \|\hat{y}_{t+1}-\bar{y}_t\|^2 + \frac{2}{\mu}\|\nabla_y g(\bar{x}_t,\bar{y}_t)-\bar{v}_t\|^2,
\end{align}
where the second inequality holds by $L_g \geq \mu$ and $0< \eta_t \leq 1$, and the last inequality is due to
$0< \lambda \leq \frac{\rho}{6L_g} \leq \frac{b_t}{6L_g}$.
It implies that
\begin{align} \label{eq:F8}
\|\bar{y}_{t+1}-y^*(\bar{x}_t)\|^2 \leq ( 1 -\frac{\eta_t\mu\lambda}{2b_t})\|\bar{y}_t -y^*(\bar{x}_t)\|^2 - \frac{3\eta_t}{4} \|\hat{y}_{t+1}-\bar{y}_t\|^2
+ \frac{4\eta_t\lambda}{\mu b_t}\|\nabla_y g(\bar{x}_t,\bar{y}_t)-\bar{v}_t\|^2.
\end{align}

Next, we decompose the term $\|\bar{y}_{t+1} - y^*(\bar{x}_{t+1})\|^2$ as follows:
\begin{align} \label{eq:F9}
  \|\bar{y}_{t+1} - y^*(\bar{x}_{t+1})\|^2 & = \|\bar{y}_{t+1} - y^*(\bar{x}_t) + y^*(\bar{x}_t) - y^*(\bar{x}_{t+1})\|^2    \nonumber \\
  & = \|\bar{y}_{t+1} - y^*(\bar{x}_t)\|^2 + 2\langle \bar{y}_{t+1} - y^*(\bar{x}_t), y^*(\bar{x}_t) - y^*(\bar{x}_{t+1})\rangle  + \|y^*(\bar{x}_t) - y^*(\bar{x}_{t+1})\|^2  \nonumber \\
  & \leq (1+\frac{\eta_t\mu\lambda}{4b_t})\|\bar{y}_{t+1} - y^*(\bar{x}_t)\|^2  + (1+\frac{4b_t}{\eta_t\mu\lambda})\|y^*(\bar{x}_t) - y^*(\bar{x}_{t+1})\|^2 \nonumber \\
  & \leq (1+\frac{\eta_t\mu\lambda}{4b_t})\|\bar{y}_{t+1} - y^*(\bar{x}_t)\|^2  + (1+\frac{4b_t}{\eta_t\mu\lambda})\kappa^2\|\bar{x}_t - \bar{x}_{t+1}\|^2,
\end{align}
where the first inequality holds by Cauchy-Schwarz inequality and Young's inequality, and the second inequality is due to
Lemma \ref{lem:2}, and the last equality holds by $\bar{x}_{t+1}=\bar{x}_t + \eta_t(\hat{x}_{t+1}-\bar{x}_t)$.

By combining the above inequalities \eqref{eq:F8} and \eqref{eq:F9}, we have
\begin{align}
 \|\bar{y}_{t+1} - y^*(\bar{x}_{t+1})\|^2 & \leq (1+\frac{\eta_t\mu\lambda}{4b_t})( 1-\frac{\eta_t\mu\lambda}{2b_t})\|\bar{y}_t -y^*(\bar{x}_t)\|^2
 - (1+\frac{\eta_t\mu\lambda}{4b_t})\frac{3\eta_t}{4} \|\hat{y}_{t+1}-\bar{y}_t\|^2     \nonumber \\
 & \quad + (1+\frac{\eta_t\mu\lambda}{4b_t})\frac{4\eta_t\lambda}{\mu b_t}\|\nabla_y g(\bar{x}_t,\bar{y}_t)-\bar{v}_t\|^2
  + (1+\frac{4b_t}{\eta_t\mu\lambda})\kappa^2\|\bar{x}_t - \bar{x}_{t+1}\|^2.   \nonumber
\end{align}
Since $0 < \eta_t \leq 1$, $0< \lambda \leq \frac{\rho}{6L_g} \leq \frac{b_t}{6L_g}$ and $L_g\geq \mu$, we have $\lambda \leq \frac{b_t}{6L_g} \leq \frac{b_t}{6\mu}$
and $\eta_t\leq 1\leq \frac{b_t}{6\mu \lambda}$. Then we have
\begin{align}
  (1+\frac{\eta_t\mu\lambda}{4b_t})(1-\frac{\eta_t\mu\lambda}{2b_t})&= 1-\frac{\eta_t\mu\lambda}{2b_t} +\frac{\eta_t\mu\lambda}{4b_t}
  - \frac{\eta_t^2\mu^2\lambda^2}{8b^2_t} \leq 1-\frac{\eta_t\mu\lambda}{4b_t}, \nonumber  \\
 - (1+\frac{\eta_t\mu\lambda}{4b_t})\frac{3\eta_t}{4} &\leq -\frac{3\eta_t}{4}, \nonumber  \\
 (1+\frac{\eta_t\mu\lambda}{4b_t})\frac{4\eta_t\lambda}{\mu b_t} & \leq (1+\frac{1}{24})\frac{4\eta_t\lambda}{\mu b_t}
 = \frac{25\eta_t\lambda}{6\mu b_t}, \nonumber  \\
  (1+\frac{4b_t}{\eta_t\mu\lambda})\kappa^2 &
  \leq \frac{b_t\kappa^2}{6\eta_t\mu\lambda} +\frac{4b_t\kappa^2}{\eta_t\mu\lambda} = \frac{25b_t\kappa^2}{6\eta_t\mu\lambda}, \nonumber
\end{align}
where the second last inequality is due to $\frac{\eta_t\mu\lambda}{b_t}\leq \frac{1}{6}$ and the last inequality holds by
$\frac{b_t}{6\mu\lambda\eta_t} \geq 1$.
By using $\bar{x}_{t+1}=\bar{x}_t + \eta_t(\hat{x}_{t+1}-\bar{x}_t)$, then we have
\begin{align}
     \|\bar{y}_{t+1} - y^*(\bar{x}_{t+1})\|^2 &\leq (1-\frac{\eta_t\mu\lambda}{4b_t})\|\bar{y}_t -y^*(\bar{x}_t)\|^2 -\frac{3\eta_t}{4} \|\hat{y}_{t+1}-\bar{y}_t\|^2 \nonumber \\
     & \quad + \frac{25\eta_t\lambda}{6\mu b_t} \|\nabla_y g(\bar{x}_t,\bar{y}_t)-\bar{v}_t\|^2 + \frac{25\kappa^2\eta_tb_t}{6\mu\lambda}\|\hat{x}_{t+1} - \bar{x}_t\|^2.
\end{align}

\end{proof}

\begin{lemma} \label{lem:A7}
Under the above assumptions, and assume the stochastic gradient estimators $\big\{\bar{v}_t,\bar{w}_t\big\}_{t=1}^T$ be generated from Algorithm \ref{alg:1},
we have
 \begin{align} \label{eq:A49}
 \mathbb{E}\|\bar{\nabla}_y g(x_{t+1},y_{t+1}) - \bar{v}_{t+1}\|^2
 & \leq (1-\alpha_{t+1})\mathbb{E} \|\bar{\nabla}_y g(x_t,y_t) - \bar{v}_t\|^2 + \frac{2\alpha_{t+1}^2\sigma^2 }{M} \nonumber \\
 & \quad + \frac{4L_g^2\eta_t^2}{M^2}\sum_{m=1}^M\mathbb{E}\big(\|\hat{x}^m_{t+1} - x^m_t\|^2 + \|\hat{y}^m_{t+1} - y^m_t\|^2 \big),
 \end{align}
 \begin{align} \label{eq:A50}
\mathbb{E}\|\bar{w}_{t+1} - \bar{\nabla} f(x_{t+1},y_{t+1}) - R_{t+1}\|^2 & \leq (1-\beta_{t+1}) \mathbb{E}\|\bar{w}_t - \bar{\nabla} f(x_t,y_t) - R_t\|^2 + \frac{2\beta^2_{t+1}\sigma^2}{M} \nonumber \\
& \quad + \frac{4L^2_K\eta^2_t}{M^2}\sum_{m=1}^M\mathbb{E}\big(\|\hat{x}^m_{t+1}-x^m_t\|^2 + \|\hat{y}^m_{t+1}-y^m_t\|^2 \big),
\end{align}
where $L_K^2 = 2L^2_f + 6C^2_{gxy}L^2_f\frac{K}{2\mu L_g - \mu^2} + 6C^2_{fy}L^2_{gxy}\frac{K}{2\mu L_g - \mu^2} + 6C^2_{gxy}L^2_f\frac{K^3L^2_{gyy}}{(L_g-\mu)^2(2\mu L_g - \mu^2)}$.
\end{lemma}

\begin{proof}
Without loss of generality, we only prove the above inequality \eqref{eq:A50}, and it is similar to \eqref{eq:A49}.
Since $\bar{w}_{t+1} = \frac{1}{M}\sum_{m=1}^M\big(\hat{\nabla} f^m(x^m_{t+1},y^m_{t+1};\bar{\xi}^m_{t+1}) + (1-\beta_{t+1})\big(w^m_t
 - \hat{\nabla} f^m(x^m_t,y^m_t;\bar{\xi}^m_{t+1})\big)\big)$,
we have
\begin{align}
 &\mathbb{E}\|\bar{w}_{t+1} - \bar{\nabla} f(x_{t+1},y_{t+1}) - R_{t+1}\|^2  \\
 & = \mathbb{E}\|\frac{1}{M}\sum_{m=1}^M\big(w^m_{t+1} - \hat{\nabla} f^m(x^m_{t+1},y^m_{t+1}) - R^m_{t+1}\big)\|^2 \nonumber \\
 & = \mathbb{E}\|\frac{1}{M}\sum_{m=1}^M\big(\hat{\nabla} f^m(x^m_{t+1},y^m_{t+1};\bar{\xi}^m_{t+1}) + (1-\beta_{t+1})\big(w^m_t
 - \hat{\nabla} f^m(x^m_t,y^m_t;\bar{\xi}^m_{t+1})\big) - \hat{\nabla} f^m(x^m_{t+1},y^m_{t+1}) - R^m_{t+1}\big)\|^2 \nonumber \\
 & = \mathbb{E}\big\|\frac{1}{M}\sum_{m=1}^M\big(\hat{\nabla} f^m(x^m_{t+1},y^m_{t+1};\bar{\xi}^m_{t+1})-\hat{\nabla} f^m(x^m_{t+1},y^m_{t+1})- R^m_{t+1} \nonumber \\
 & \quad - (1-\beta_{t+1})\big(
 \hat{\nabla} f^m(x^m_t,y^m_t;\bar{\xi}^m_{t+1}) - \hat{\nabla} f^m(x^m_t,y^m_t) - R^m_t\big)\big) \nonumber \\
 & \quad + (1-\beta_{t+1})\frac{1}{M}\sum_{m=1}^M\big( w^m_t - \hat{\nabla} f^m(x^m_t,y^m_t) - R^m_t \big) \big\|^2 \nonumber \\
 & = \frac{1}{M^2}\sum_{m=1}^M\mathbb{E}\big\|\hat{\nabla} f^m(x^m_{t+1},y^m_{t+1};\bar{\xi}^m_{t+1})-\hat{\nabla} f^m(x^m_{t+1},y^m_{t+1})- R^m_{t+1} \nonumber \\
 & \quad - (1-\beta_{t+1})\big(
 \hat{\nabla} f^m(x^m_t,y^m_t;\bar{\xi}^m_{t+1}) - \hat{\nabla} f^m(x^m_t,y^m_t) - R^m_t\big)\|^2 + (1-\beta_{t+1})^2\mathbb{E}\|\bar{w}_t - \bar{\nabla} f(x_t,y_t) - R_t\big\|^2 \nonumber \\
 & \leq \frac{2(1-\beta_{t+1})^2}{M^2}\sum_{m=1}^M\mathbb{E}\|\hat{\nabla} f^m(x^m_{t+1},y^m_{t+1};\bar{\xi}^m_{t+1}) - \hat{\nabla} f^m(x^m_t,y^m_t;\bar{\xi}^m_{t+1}) -
 \big( \hat{\nabla} f^m(x^m_{t+1},y^m_{t+1})+ R^m_{t+1}- \hat{\nabla} f^m(x^m_t,y^m_t) - R^m_t\big) \|^2 \nonumber \\
 & \quad + \frac{2\beta^2_{t+1}}{M^2}\sum_{m=1}^M\mathbb{E}\|\hat{\nabla} f^m(x^m_{t+1},y^m_{t+1};\bar{\xi}^m_{t+1}) -
 \hat{\nabla} f^m(x^m_{t+1},y^m_{t+1})- R^m_{t+1} \|^2 + (1-\beta_{t+1})^2\mathbb{E}\|\bar{w}_t - \bar{\nabla} f(x_t,y_t) - R_t\|^2 \nonumber \\
 & \leq \frac{2(1-\beta_{t+1})^2}{M^2}\sum_{m=1}^M\mathbb{E}\|\hat{\nabla} f^m(x^m_{t+1},y^m_{t+1};\bar{\xi}^m_{t+1}) - \hat{\nabla} f^m(x^m_t,y^m_t;\bar{\xi}^m_{t+1}) \|^2 + \frac{2\beta^2_{t+1}\sigma^2}{M} \nonumber \\
 & \quad  + (1-\beta_{t+1})^2\mathbb{E}\|\bar{w}_t - \bar{\nabla} f(x_t,y_t) - R_t\|^2 \nonumber \\
 & \leq (1-\beta_{t+1})^2 \mathbb{E}\|\bar{w}_t - \bar{\nabla} f(x_t,y_t) - R_t\|^2 + \frac{2\beta^2_{t+1}\sigma^2}{M} + \frac{4(1-\beta_{t+1})^2L^2_K}{M^2}\sum_{m=1}^M\mathbb{E}\big(
  \|x^m_{t+1}-x^m_t\|^2 + \|y^m_{t+1}-y^m_t\|^2 \big) \nonumber \\
 & \leq (1-\beta_{t+1}) \mathbb{E}\|\bar{w}_t - \bar{\nabla} f(x_t,y_t) - R_t\|^2 + \frac{2\beta^2_{t+1}\sigma^2}{M} + \frac{4L^2_K\eta^2_t}{M^2}\sum_{m=1}^M\mathbb{E}\big(
  \|\hat{x}^m_{t+1}-x^m_t\|^2 + \|\hat{y}^m_{t+1}-y^m_t\|^2 \big) \nonumber,
\end{align}
where the forth equality holds by the following fact: for any $m\in [M]$,
\begin{align}
 \mathbb{E}_{\bar{\xi}^m_{t+1}} \big[\hat{\nabla} f^m(x^m_{t+1},y^m_{t+1};\bar{\xi}^m_{t+1})\big]=\hat{\nabla} f^m(x^m_{t+1},y^m_{t+1})+ R^m_{t+1}, \
 \mathbb{E}_{\bar{\xi}^m_{t+1}} \big[\hat{\nabla} f^m(x^m_t,y^m_t;\bar{\xi}^m_{t+1}))\big]= \hat{\nabla} f^m(x^m_t,y^m_t) + R^m_t, \nonumber
\end{align}
and for any $m\neq j\in [M]$, $\bar{\xi}^m_{t+1}$ and $\bar{\xi}^j_{t+1}$ are independent, i.e.,
\begin{align}
 \big\langle \hat{\nabla} f^m(x^m_{t+1},y^m_{t+1};\bar{\xi}^m_{t+1})-\hat{\nabla} f^m(x^m_{t+1},y^m_{t+1})- R^m_{t+1} - (1-\beta_{t+1})\big(
 \hat{\nabla} f^m(x^m_t,y^m_t;\bar{\xi}^m_{t+1}) - \hat{\nabla} f^m(x^m_t,y^m_t) - R^m_t\big) \nonumber \\
 ,\hat{\nabla} f^m(x^j_{t+1},y^j_{t+1};\bar{\xi}^j_{t+1})-\hat{\nabla} f^j(x^j_{t+1},y^j_{t+1})- R^j_{t+1} - (1-\beta_{t+1})\big(
 \hat{\nabla} f^j(x^j_t,y^j_t;\bar{\xi}^j_{t+1}) - \hat{\nabla} f^j(x^j_t,y^j_t) - R^j_t\big)\big\rangle =0; \nonumber
\end{align}
the second inequality holds by the inequality $\mathbb{E}\|\zeta-\mathbb{E}[\zeta]\|^2 \leq \mathbb{E}\|\zeta\|^2$ and Assumption \ref{ass:5};
the second last inequality is due to Lemma \ref{lem:3}; the last inequality holds by $0<\beta_{t+1} \leq 1$ and $x^m_{t+1}=x^m_t+\eta_t(\hat{x}^m_{t+1}-x^m_t)$,
$y^m_{t+1}=y^m_t+\eta_t(\hat{y}^m_{t+1}-y^m_t)$.

\end{proof}

\begin{lemma} \label{lem:A9}
(Restatement of Lemma \ref{lem:6})
Based on the above Assumptions \ref{ass:1}-\ref{ass:3} and \ref{ass:7}, for any $m,j\in[M]$, we have
\begin{align}
 \big\|\hat{\nabla}f^m(x,y)-\hat{\nabla}f^j(x,y)\big\|^2 \leq \hat{\delta}^2,
\end{align}
where $\hat{\delta}^2=4\delta^2_f + \frac{4C^2_{fy}\delta^2_{g}}{\mu^2} + \frac{4C^2_{gxy}C^2_{fy}\delta^2_{g}}{\mu^4} + \frac{4C^2_{gxy}\delta^2_{f}}{\mu^2}$.
\end{lemma}

\begin{proof}
\begin{align}
 & \big\|\hat{\nabla} f^m(x,y)-\hat{\nabla} f^j(x,y)\big\|^2  \nonumber \\
&= \bigg\|\nabla_x f^m(x,y) - \nabla^2_{xy}g^m(x, y) \Big(\nabla^2_{yy} g^m(x, y)\Big)^{-1}\nabla_{y} f^m(x, y) \nonumber \\
& \quad \quad  - \nabla_x f^j(x, y) + \nabla^2_{xy} g^j(x, y) \Big(\nabla^2_{yy} g^j(x, y)\Big)^{-1}\nabla_{y}f^j(x,y) \bigg\|  \nonumber \\
&= \bigg\|\nabla_x f^m(x, y)- \nabla_x f^j(x, y) - \nabla^2_{xy} g^m(x, y) \Big(\nabla^2_{yy} g^m(x, y)\Big)^{-1}\nabla_{y}f^m(x, y)
+ \nabla^2_{xy} g^j(x, y) \Big(\nabla^2_{yy} g^m(x, y)\Big)^{-1}\nabla_{y} f^m(x, y)\nonumber \\
& \quad \quad -\nabla^2_{xy} g^j(x, y) \Big(\nabla^2_{yy} g^m(x, y)\Big)^{-1}\nabla_{y} f^m(x, y) + \nabla^2_{xy} g^j(x, y) \Big(\nabla^2_{yy} g^j(x,y)\Big)^{-1}\nabla_{y}f^m(x, y) \nonumber \\
& \quad \quad - \nabla^2_{xy} g^j(x, y) \Big(\nabla^2_{yy} g^j(x, y)\Big)^{-1}\nabla_{y} f^m(x, y) + \nabla^2_{xy} g^j(x, y) \Big(\nabla^2_{yy}g^j(x,y)\Big)^{-1}\nabla_{y}f^j(x,y) \bigg\|  \nonumber \\
& \leq 4\bigg\|\nabla_x f^m(x, y) - \nabla_x f^j(x, y) \bigg\|^2 + 4\bigg\| \nabla^2_{xy}g^m(x, y) - \nabla^2_{xy} g^j(x, y) \bigg\|^2
\bigg\|\Big(\nabla^2_{yy} g^m(x, y)\Big)^{-1}\nabla_{y} f^m(x, y)\bigg\|^2 \nonumber \\
& \quad + 4\bigg\|\nabla^2_{yy} g^m(x, y)  - \nabla^2_{yy} g^j(x, y) \bigg\|^2\bigg\| \nabla^2_{xy}g^j(x,y)\nabla_{y}f^m(x,y)\Big(\nabla^2_{yy}g^m(x,y)\Big)^{-1}\Big(\nabla^2_{yy}g^j(x,y)\Big)^{-1} \bigg\|^2 \nonumber \\
& \quad + 4\bigg\|\nabla_{y}f^m(x,y) - \nabla_{y}f^j(x, y) \bigg\|^2\bigg\|\nabla^2_{xy}g^j(x, y) \Big(\nabla^2_{yy}g^j(x, y)\Big)^{-1}\bigg\|^2  \nonumber \\
& \leq 4\delta^2_f + \frac{4C^2_{fy}\delta^2_{g}}{\mu^2} + \frac{4C^2_{gxy}C^2_{fy}\delta^2_{g}}{\mu^4} + \frac{4C^2_{gxy}\delta^2_{f}}{\mu^2},
\end{align}
where the last inequality holds by the above Assumptions \ref{ass:1}-\ref{ass:3} and \ref{ass:7}.

\end{proof}

\begin{lemma} \label{lem:A8}
(Restatement of Lemma \ref{lem:7})
Based on the above Assumptions \ref{ass:1}-\ref{ass:3}, we have, for any $m\in[M]$, $x_1,x_2\in \mathbb{R}^d$ and $y_1,y_2\in \mathbb{R}^p$,
\begin{align}
\|\hat{\nabla}f^m(x_1,y_1)-\hat{\nabla}f^m(x_2,y_2)\|^2 \leq \hat{L}^2\big(\|x_1-x_2\|^2+\|y_1-y_2\|^2\big),
\end{align}
where $\hat{L}^2=8L^2_f + \frac{8L^2_{gxy}C^2_{fy}}{\mu^2} + \frac{8L^2_{gyy}C^2_{gxy}C^2_{fy}}{\mu^4} + \frac{8L^2_fC^2_{gxy}}{\mu^2}$.
\end{lemma}

\begin{proof}

\begin{align}
& \|\hat{\nabla}f^m(x_1,y_1)-\hat{\nabla}f^m(x_2,y_2)\|^2 \nonumber \\
&= \bigg\|\nabla_x f^m(x_1, y_1) - \nabla^2_{xy} g^m(x_1, y_1) \big(\nabla^2_{yy} g^m(x_1, y_1)\Big)^{-1}\nabla_{y} f^m(x_1, y_1) \nonumber \\
& \quad \quad  - \nabla_x f^m(x_2, y_2) + \nabla^2_{xy} g^m(x_2, y_2) \big(\nabla^2_{yy} g^m(x_2, y_2)\Big)^{-1}\nabla_{y} f^m(x_2, y_2) \bigg\|  \nonumber \\
&= \bigg\|\nabla_x f^m(x_1, y_1)- \nabla_x f^m(x_2, y_2) - \nabla^2_{xy} g^m(x_1, y_1) \big(\nabla^2_{yy} g^m(x_1, y_1)\Big)^{-1}\nabla_{y} f^m(x_1, y_1) \nonumber \\
& \quad \quad + \nabla^2_{xy} g^m(x_2, y_2) \big(\nabla^2_{yy} g^m(x_1, y_1)\Big)^{-1}\nabla_{y} f^m(x_1, y_1)\nonumber \\
& \quad \quad -\nabla^2_{xy} g^m(x_2, y_2) \big(\nabla^2_{yy} g^m(x_1, y_1)\Big)^{-1}\nabla_{y} f^m(x_1, y_1) + \nabla^2_{xy} g^m(x_2, y_2) \big(\nabla^2_{yy} g^m(x_2, y_2)\Big)^{-1}\nabla_{y} f^m(x_1, y_1) \nonumber \\
& \quad \quad - \nabla^2_{xy} g^m(x_2, y_2) \big(\nabla^2_{yy} g^m(x_2, y_2)\Big)^{-1}\nabla_{y} f^m(x_1, y_1) + \nabla^2_{xy} g^m(x_2, y_2) \big(\nabla^2_{yy} g^m(x_2, y_2)\Big)^{-1}\nabla_{y} f^m(x_2, y_2) \bigg\|  \nonumber \\
& \leq 4\bigg\|\nabla_x f^m(x_1, y_1) - \nabla_x f^m(x_2, y_2) \bigg\|^2 + 4\bigg\| \nabla^2_{xy} g^m(x_1, y_1) - \nabla^2_{xy} g^m(x_2, y_2) \bigg\|^2
\bigg\|\big(\nabla^2_{yy} g^m(x_1, y_1)\Big)^{-1}\nabla_{y} f^m(x_1, y_1)\bigg\|^2 \nonumber \\
& \quad + 4\bigg\| \nabla^2_{yy} g^m(x_1, y_1)  - \nabla^2_{yy} g^m(x_2, y_2) \bigg\|^2\bigg\| \nabla^2_{xy} g^m(x_2, y_2)\nabla_{y} f^m(x_1, y_1)\big(\nabla^2_{yy} g^m(x_1, y_1)\Big)^{-1}\big(\nabla^2_{yy}g^m(x_2, y_2)\Big)^{-1} \bigg\|^2 \nonumber \\
& \quad + 4\bigg\| \nabla_{y} f^m(x_1, y_1)  - \nabla_{y} f^m(x_2, y_2) \bigg\|^2\bigg\|\nabla^2_{xy} g^m(x_2, y_2) \big(\nabla^2_{yy} g^m(x_2, y_2)\Big)^{-1}\bigg\|^2  \nonumber \\
& \leq \big( 8L^2_f + \frac{8L^2_{gxy}C^2_{fy}}{\mu^2} + \frac{8L^2_{gyy}C^2_{gxy}C^2_{fy}}{\mu^4} + \frac{8L^2_fC^2_{gxy}}{\mu^2} \big) \big(\|x_1-x_2\|^2+\|y_1-y_2\|^2\big)  \nonumber \\
& = \hat{L}^2\big(\|x_1-x_2\|^2+\|y_1-y_2\|^2\big),
\end{align}
where the last inequality holds by the above Assumptions \ref{ass:1}-\ref{ass:3}.

\end{proof}

\begin{lemma} \label{lem:A10}
Based on the above Assumptions \ref{ass:1}-\ref{ass:3} and \ref{ass:7}, we have
\begin{align}
 & \sum_{m=1}^M\mathbb{E}\big\|\hat{\nabla}f^m(x^m_t,y^m_t)-\frac{1}{M}\sum_{j=1}^M\hat{\nabla}f^j(x^j_t,y^j_t)\big\|^2 \leq
 6\hat{L}^2\sum_{m=1}^M\big(\mathbb{E}\|x^m_t-\bar{x}_t\|^2 + \mathbb{E}\|y^m_t-\bar{y}_t\|^2\big) + 3M\hat{\delta}^2, \nonumber \\
 & \sum_{m=1}^M\mathbb{E}\big\|\nabla_y g^m(x^m_t,y^m_t)-\frac{1}{M}\sum_{j=1}^M\nabla_yg^j(x^j_t,y^j_t)\big\|^2 \leq 6L^2_g\sum_{m=1}^M\big(\mathbb{E}\|x^m_t-\bar{x}_t\|^2
 + \mathbb{E}\|y^m_t-\bar{y}_t\|^2\big) + 3M\delta_{g}^2. \nonumber
\end{align}

\end{lemma}

\begin{proof}
Consider the term $\sum_{m=1}^M\mathbb{E}\big\|\hat{\nabla}f^m(x^m_t,y^m_t)-\frac{1}{M}\sum_{j=1}^M\hat{\nabla}f^j(x^j_t,y^j_t)\big\|^2$, we have
\begin{align}
 & \sum_{m=1}^M\mathbb{E}\big\|\hat{\nabla}f^m(x^m_t,y^m_t)-\frac{1}{M}\sum_{j=1}^M\hat{\nabla}f^j(x^j_t,y^j_t)\big\|^2   \nonumber \\
 & = \sum_{m=1}^M\mathbb{E}\big\|\hat{\nabla}f^m(x^m_t,y^m_t)-\hat{\nabla}f^m(\bar{x}_t,\bar{y}_t) +\hat{\nabla}f^m(\bar{x}_t,\bar{y}_t) - \frac{1}{M}\sum_{j=1}^M\hat{\nabla}f^j(\bar{x}_t,\bar{y}_t) \nonumber \\
 & \quad + \frac{1}{M}\sum_{j=1}^M\hat{\nabla}f^j(\bar{x}_t,\bar{y}_t)-\frac{1}{M}\sum_{j=1}^M\hat{\nabla}f^j(x^j_t,y^j_t)\big\|^2   \nonumber \\
 & \leq \sum_{m=1}^M 3\mathbb{E}\big\|\hat{\nabla}f^m(x^m_t,y^m_t)-\hat{\nabla}f^m(\bar{x}_t,\bar{y}_t)\big\|^2 + \sum_{m=1}^M 3\mathbb{E}\big\|\hat{\nabla}f^m(\bar{x}_t,\bar{y}_t) - \frac{1}{M}\sum_{j=1}^M\hat{\nabla}f^j(\bar{x}_t,\bar{y}_t)\big\|^2 \nonumber \\
 & \quad + \sum_{m=1}^M 3\mathbb{E}\big\|\frac{1}{M}\sum_{j=1}^M\hat{\nabla}f^j(\bar{x}_t,\bar{y}_t)-\frac{1}{M}\sum_{j=1}^M\hat{\nabla}f^j(x^j_t,y^j_t)\big\|^2   \nonumber \\
 & \leq 3\hat{L}^2\sum_{m=1}^M\big(\mathbb{E}\|x^m_t-\bar{x}_t\|^2 + \mathbb{E}\|y^m_t-\bar{y}_t\|^2\big) + 3\sum_{m=1}^M \frac{1}{M}\sum_{j=1}^M \mathbb{E}\|\hat{\nabla}f^m(\bar{x}_t,\bar{y}_t) - \hat{\nabla}f^j(\bar{x}_t,\bar{y}_t)\|^2 \nonumber \\
 & \quad + 3\sum_{m=1}^M\frac{1}{M}\sum_{j=1}^M\big\|\hat{\nabla}f^j(\bar{x}_t,\bar{y}_t)-\hat{\nabla}f^j(x^j_t,y^j_t)\big\|^2 \nonumber \\
 & \leq 6\hat{L}^2\sum_{m=1}^M\big(\mathbb{E}\|x^m_t-\bar{x}_t\|^2 + \mathbb{E}\|y^m_t-\bar{y}_t\|^2\big) + 3M\hat{\delta}^2,
\end{align}
where the second last inequality holds by the above Lemma \ref{lem:A8}, and the last inequality holds by the above Lemma \ref{lem:A9}.

Next, we have
\begin{align}
 & \sum_{m=1}^M\mathbb{E}\big\|\nabla_y g^m(x^m_t,y^m_t)-\frac{1}{M}\sum_{j=1}^M\nabla_yg^j(x^j_t,y^j_t)\big\|^2   \nonumber \\
 & = \sum_{m=1}^M\mathbb{E}\big\|\nabla_yg^m(x^m_t,y^m_t)- \nabla g^m(\bar{x}_t,\bar{y}_t) + \nabla g^m(\bar{x}_t,\bar{y}_t) - \frac{1}{M}\sum_{j=1}^M\nabla_yg^j(\bar{x}_t,\bar{y}_t) \nonumber \\
 & \quad + \frac{1}{M}\sum_{j=1}^M\nabla_y g^j(\bar{x}_t,\bar{y}_t)-\frac{1}{M}\sum_{j=1}^M\nabla_y g^j(x^j_t,y^j_t)\big\|^2   \nonumber \\
 & \leq \sum_{m=1}^M 3\mathbb{E}\big\|\nabla_yg^m(x^m_t,y^m_t)-\nabla_yg^m(\bar{x}_t,\bar{y}_t)\big\|^2 + \sum_{m=1}^M 3\mathbb{E}\big\|\nabla_yg^m(\bar{x}_t,\bar{y}_t) - \frac{1}{M}\sum_{j=1}^M\nabla_yg^j(\bar{x}_t,\bar{y}_t)\big\|^2 \nonumber \\
 & \quad + \sum_{m=1}^M 3\mathbb{E}\big\|\frac{1}{M}\sum_{j=1}^M\nabla_yg^j(\bar{x}_t,\bar{y}_t)-\frac{1}{M}\sum_{j=1}^M\nabla_yg^j(x^j_t,y^j_t)\big\|^2   \nonumber \\
 & \leq 3L^2_g\sum_{m=1}^M\big(\mathbb{E}\|x^m_t-\bar{x}_t\|^2 + \mathbb{E}\|y^m_t-\bar{y}_t\|^2\big) + 3\sum_{m=1}^M \frac{1}{M}\sum_{j=1}^M \mathbb{E}\|\nabla_yg^m(\bar{x}_t,\bar{y}_t) - \nabla_yg^j(\bar{x}_t,\bar{y}_t)\|^2 \nonumber \\
 & \quad + 3\sum_{m=1}^M\frac{1}{M}\sum_{j=1}^M\big\|\nabla_yg^j(\bar{x}_t,\bar{y}_t)-\nabla_yg^j(x^j_t,y^j_t)\big\|^2 \nonumber \\
 & \leq 6L^2_g\sum_{m=1}^M\big(\mathbb{E}\|x^m_t-\bar{x}_t\|^2 + \mathbb{E}\|y^m_t-\bar{y}_t\|^2\big) + 3M\delta_{g}^2,
\end{align}
where the last inequality is due to the above Assumption \ref{ass:7}.

\end{proof}

\begin{lemma} \label{lem:A11}
Suppose the iterates $\{x^m_t,y^m_t\}_{t=1}^T$, for all $m \in [M]$ generated from Algorithm \ref{alg:1} satisfy:
\begin{align}
& \sum_{m=1}^M \mathbb{E}\|x^m_t- \bar{x}_t \|^2 \leq (q-1)\sum_{l = s_t+1}^{t-1} \gamma^2\eta_l^2 \sum_{m = 1}^M \mathbb{E}\|A^{-1}_l(w^m_l - \bar{w}_l)\|^2, \nonumber \\
& \sum_{m=1}^M \mathbb{E}\|y^m_t- \bar{y}_t \|^2 \leq (q-1)\sum_{l = s_t+1}^{t-1} \lambda^2\eta_l^2 \sum_{m = 1}^M \mathbb{E}\|B^{-1}_l(v^m_l - \bar{v}_l)\|^2. \nonumber
\end{align}
\end{lemma}

\begin{proof}
From Algorithm \ref{alg:1}, when $s_t=q\lfloor t/q \rfloor+1$, we have $t = s_t$ and $x^m_t = \bar{x}_t$, the above inequality holds trivially. When $t\in (s_t,s_t+q)$, we have
\begin{align*}
    x^m_t = x^m_{s_t+1} - \sum_{l=s_t+1}^{t-1}\gamma\eta_lA_l^{-1}w^m_l, \quad \text{and} \quad \bar{x}_{t}  = \bar{x}_{s_t+1}  - \sum_{l=s_t+1}^{t-1}\gamma\eta_l A_l^{-1}\bar{w}_l.
\end{align*}
Thus, we have
\begin{align*}
  \sum_{m=1}^M\mathbb{E}\|x^m_t - \bar{x}_t\|^2 & = \sum_{m =1}^M \mathbb{E}\Big\|x^m_{s_t+1} - \bar{x}_{s_t+1}
  - \Big(\sum_{l=s_t+1}^{t-1}\gamma\eta_lA_l^{-1}w^m_l - \sum_{l=s_t+1}^{t-1} \gamma\eta_l A_l^{-1} \bar{w}_l\Big) \Big\|^2 \\
  & = \sum_{m = 1}^M \mathbb{E}\Big\|\sum_{l=s_t+1}^{t-1}\big(\gamma\eta_lA_l^{-1}w^m_l - \gamma\eta_l A_l^{-1} \bar{w}_l\big) \Big\|^2
  \leq {(q -1)} \sum_{l = s_t+1}^{t-1} \gamma^2\eta_l^2 \sum_{m = 1}^M \mathbb{E}\|A^{-1}_l(w^m_l - \bar{w}_l)\|^2,
\end{align*}
where the above inequality is due to $t-s_t-1\leq q-1$.
Similarly, we can obtain
\begin{align}
\sum_{m=1}^M \mathbb{E}\|y^m_t- \bar{y}_t \|^2 \leq (q-1)\sum_{l = s_t+1}^{t-1} \lambda^2\eta_l^2 \sum_{m = 1}^M \mathbb{E}\|B^{-1}_l(v^m_l - \bar{v}_l)\|^2. \nonumber
\end{align}

\end{proof}

\begin{lemma} \label{lem:A12}
Let $\eta_t \leq \frac{\rho\theta}{12\lambda q\sqrt{M}\sqrt{L^2_K+L^2_g}} \ (0< \theta \leq 1)$ for all $t\geq 1$, $\gamma=\tau\lambda \ (0<\tau \leq 1)$, $\alpha_{t+1}=c_1\eta^2_t$, $\beta_{t+1}=c_2\eta^2_t$ and
$c_2=\vartheta c_1$ with $\vartheta>0$, and $c_1\leq \frac{72\lambda^2q(L^2_K+L^2_g)}{\rho^2\sqrt{\vartheta^2\hat{L}^2+L^2_g}}$.
Set $s_t = q\lfloor t/q \rfloor + 1$ and $t\in [s_t,s_t+q)$, we have
\begin{align}
  & \sum_{t=s_t}^{s_t+q-1}\eta_t\sum_{m=1}^M\mathbb{E}\big( \|A^{-1}_t(w^m_t - \bar{w}_t)\|^2 + \|B^{-1}_t(v^m_t - \bar{v}_t)\|^2 \big)  \nonumber \\
  & \leq \frac{8M}{15}\sum_{t=s_t}^{s_t+q-1}\eta_t\mathbb{E}\big(\tau^2\|A^{-1}_t\bar{w}_t\|^2+\|B^{-1}_t\bar{v}_t\|^2\big)
   + \frac{4M\hat{c}^2 }{15\lambda^2(L^2_K+L^2_g)}\sum_{t=s_t}^{s_t+q-1} \eta^3_t,  \nonumber
\end{align}
where $\hat{c}^2 = \frac{2c^2_2}{T^2} + 2c^2_2\sigma^2 + c^2_1\sigma^2 + 3c^2_2\hat{\delta}^2 + 3c^2_1\delta_{g}^2$.
\end{lemma}

\begin{proof}
When $t=s_t=q\lfloor t/q\rfloor+1$, we have $w^m_{t} = \bar{w}_{t}$ for all $m\in [M]$. So $\sum_{m=1}^M\mathbb{E}\|A^{-1}_{t}(w^m_{t} - \bar{w}_{t})\|=0$,
clearly, the about inequality in the lemma holds trivially.
When $t\in (s_t,s_t+q)$, we have
\begin{align} \label{eq:A58}
  & \sum_{m=1}^M\mathbb{E}\|A^{-1}_{t+1}(w^m_{t+1} - \bar{w}_{t+1})\|^2  \\
  & = \sum_{m=1}^M\mathbb{E}\big\|A^{-1}_{t+1}\bigg(\hat{\nabla} f^m(x^m_{t+1},y^m_{t+1};\bar{\xi}^m_{t+1}) + (1-\beta_{t+1})\big(w^m_t
 - \hat{\nabla} f^m(x^m_t,y^m_t;\bar{\xi}^m_{t+1})\big) \nonumber \\
  & \quad - \frac{1}{M}\sum_{m=1}^M\big(\hat{\nabla} f^m(x^m_{t+1},y^m_{t+1};\bar{\xi}^m_{t+1}) + (1-\beta_{t+1})\big(w^m_t
 - \hat{\nabla} f^m(x^m_t,y^m_t;\bar{\xi}^m_{t+1})\big)\big)\bigg)\big\|^2 \nonumber \\
  & = \sum_{m=1}^M\mathbb{E}\big\|A^{-1}_{t+1}\bigg( (1-\beta_{t+1})(w^m_t - \bar{w}_t) + \big(\hat{\nabla} f^m(x^m_{t+1},y^m_{t+1};\bar{\xi}^m_{t+1}) - \frac{1}{M}\sum_{m=1}^M\hat{\nabla} f^m(x^m_{t+1},y^m_{t+1};\bar{\xi}^m_{t+1})\big) \nonumber \\
  & \quad - (1-\beta_{t+1})\big(\hat{\nabla} f^m(x^m_t,y^m_t;\bar{\xi}^m_{t+1}) - \frac{1}{M}\sum_{m=1}^M\hat{\nabla} f^m(x^m_t,y^m_t;\bar{\xi}^m_{t+1}) \big) \bigg)\big\|^2 \nonumber \\
  & \leq (1+\nu)(1-\beta_{t+1})^2\sum_{m=1}^M\mathbb{E}\|A^{-1}_t(w^m_t - \bar{w}_t)\|^2 + (1+\frac{1}{\nu})\frac{1}{\rho^2}\sum_{m=1}^M\mathbb{E}\big\|\hat{\nabla} f^m(x^m_{t+1},y^m_{t+1};\bar{\xi}^m_{t+1})  \nonumber \\
  & \quad - \frac{1}{M}\sum_{m=1}^M\hat{\nabla} f^m(x^m_{t+1},y^m_{t+1};\bar{\xi}^m_{t+1}) - (1-\beta_{t+1})\big(\hat{\nabla} f^m(x^m_t,y^m_t;\bar{\xi}^m_{t+1}) - \frac{1}{M}\sum_{m=1}^M\hat{\nabla} f^m(x^m_t,y^m_t;\bar{\xi}^m_{t+1}) \big)\big\|^2 \nonumber
\end{align}
where the last inequality holds by $A_{t+1}=A_t$ for any $t\in [s_t,s_t+q-1)$ and $A_t\succeq \rho I_d$ for any $t\geq 1$.

Next, we have
\begin{align} \label{eq:A59}
  & \sum_{m=1}^M \mathbb{E}\big\|\hat{\nabla} f^m(x^m_{t+1},y^m_{t+1};\bar{\xi}^m_{t+1})
  - \frac{1}{M}\sum_{m=1}^M\hat{\nabla} f^m(x^m_{t+1},y^m_{t+1};\bar{\xi}^m_{t+1})  \\
  & \quad - (1-\beta_{t+1})\big(\hat{\nabla} f^m(x^m_t,y^m_t;\bar{\xi}^m_{t+1}) - \frac{1}{M}\sum_{m=1}^M\hat{\nabla} f^m(x^m_t,y^m_t;\bar{\xi}^m_{t+1}) \big)\big\|^2 \nonumber \\
  & = \sum_{m=1}^M \mathbb{E}\big\| \hat{\nabla} f^m(x^m_{t+1},y^m_{t+1};\bar{\xi}^m_{t+1})-\hat{\nabla} f^m(x^m_t,y^m_t;\bar{\xi}^m_{t+1})
  - \frac{1}{M}\sum_{m=1}^M\big(\hat{\nabla} f^m(x^m_{t+1},y^m_{t+1};\bar{\xi}^m_{t+1})-\hat{\nabla} f^m(x^m_t,y^m_t;\bar{\xi}^m_{t+1})\big)  \nonumber \\
  & \quad +\beta_{t+1}\big(\hat{\nabla} f^m(x^m_t,y^m_t;\bar{\xi}^m_{t+1}) - \frac{1}{M}\sum_{m=1}^M\hat{\nabla} f^m(x^m_t,y^m_t;\bar{\xi}^m_{t+1}) \big)\big\|^2 \nonumber \\
  & \leq 2\sum_{m=1}^M\mathbb{E}\big\|\hat{\nabla} f^m(x^m_{t+1},y^m_{t+1};\bar{\xi}^m_{t+1})-\hat{\nabla} f^m(x^m_t,y^m_t;\bar{\xi}^m_{t+1})\|^2 \nonumber \\
  & \quad + 2\beta^2_{t+1}\sum_{m=1}^M \mathbb{E}\big\|\hat{\nabla} f^m(x^m_t,y^m_t;\bar{\xi}^m_{t+1}) - \frac{1}{M}\sum_{m=1}^M\hat{\nabla} f^m(x^m_t,y^m_t;\bar{\xi}^m_{t+1}) \big\|^2 \nonumber \\
  & \leq 4L^2_K\sum_{m=1}^M\mathbb{E}\big(\|x^m_{t+1}-x^m_t\|^2+\|y^m_{t+1}-y^m_t\|^2\big)+ 2\beta^2_{t+1}\sum_{m=1}^M\mathbb{E}\big\|\hat{\nabla} f^m(x^m_t,y^m_t;\bar{\xi}^m_{t+1}) - \frac{1}{M}\sum_{m=1}^M\hat{\nabla} f^m(x^m_t,y^m_t;\bar{\xi}^m_{t+1}) \big\|^2,  \nonumber
\end{align}
where the second last inequality is due to Young inequality and the above Lemma \ref{lem:A2}, and
the last inequality holds by the above Lemma \ref{lem:3}.

Consider the term $\sum_{m=1}^M\big\|\hat{\nabla} f^m(x^m_t,y^m_t;\bar{\xi}^m_{t+1}) - \frac{1}{M}\sum_{m=1}^M\hat{\nabla} f^m(x^m_t,y^m_t;\bar{\xi}^m_{t+1}) \big\|$, we have
\begin{align} \label{eq:A60}
 & \sum_{m=1}^M \big\|\hat{\nabla} f^m(x^m_t,y^m_t;\bar{\xi}^m_{t+1}) - \frac{1}{M}\sum_{m=1}^M\hat{\nabla} f^m(x^m_t,y^m_t;\bar{\xi}^m_{t+1}) \big\|^2 \nonumber \\
 & = \sum_{m=1}^M \big\|\hat{\nabla} f^m(x^m_t,y^m_t;\bar{\xi}^m_{t+1}) - \hat{\nabla} f^m(x^m_t,y^m_t) - \frac{1}{M}\sum_{m=1}^M\big(\hat{\nabla} f^m(x^m_t,y^m_t;\bar{\xi}^m_{t+1}) - \hat{\nabla} f^m(x^m_t,y^m_t)\big) \nonumber \\
 & \quad \quad + \hat{\nabla} f^m(x^m_t,y^m_t) - \frac{1}{M}\sum_{m=1}^M\hat{\nabla} f^m(x^m_t,y^m_t) \big\|^2 \nonumber \\
 & \leq 2\sum_{m=1}^M\big\|\hat{\nabla} f^m(x^m_t,y^m_t;\bar{\xi}^m_{t+1}) - \hat{\nabla} f^m(x^m_t,y^m_t) - \frac{1}{M}\sum_{m=1}^M\big(\hat{\nabla} f^m(x^m_t,y^m_t;\bar{\xi}^m_{t+1}) - \hat{\nabla} f^m(x^m_t,y^m_t)\big) \big\| \nonumber \\
 & \quad \quad + 2\sum_{m=1}^M\big\|\hat{\nabla} f^m(x^m_t,y^m_t) - \frac{1}{M}\sum_{m=1}^M\hat{\nabla} f^m(x^m_t,y^m_t) \big\|^2 \nonumber \\
 & \leq 2\sum_{m=1}^M\big\|\hat{\nabla} f^m(x^m_t,y^m_t;\bar{\xi}^m_{t+1}) - \hat{\nabla} f^m(x^m_t,y^m_t)\big\| + 2\sum_{m=1}^M\big\|\hat{\nabla} f^m(x^m_t,y^m_t) - \frac{1}{M}\sum_{m=1}^M\hat{\nabla} f^m(x^m_t,y^m_t) \big\|^2 \nonumber \\
 & \leq 4\sum_{m=1}^M\big(\|R^m_t\|^2 + \sigma^2\big) + 12\hat{L}^2\sum_{m=1}^M\big(\mathbb{E}\|x^m_t-\bar{x}_t\|^2 + \mathbb{E}\|y^m_t-\bar{y}_t\|^2\big) + 6M\hat{\delta}^2,
\end{align}
where the last inequality holds by the above Lemma \ref{lem:A10}.

By combining the above inequalities \eqref{eq:A58}, \eqref{eq:A59} and \eqref{eq:A60}, we have
\begin{align} \label{eq:A61}
 & \sum_{m=1}^M\mathbb{E}\|A^{-1}_{t+1}(w^m_{t+1} - \bar{w}_{t+1})\|^2  \\
 & \leq (1+\nu)(1-\beta_{t+1})^2\sum_{m=1}^M\mathbb{E}\|A^{-1}_t(w^m_t - \bar{w}_t)\|^2 + (1+\frac{1}{\nu})\frac{1}{\rho^2}
 \bigg(4L^2_K\sum_{m=1}^M\mathbb{E}\big(\|x^m_{t+1}-x^m_t\|^2+\|y^m_{t+1}-y^m_t\|^2\big)  \nonumber \\
 & \quad + 8\beta^2_{t+1}\sum_{m=1}^M\big(\|R^m_t\|^2 + \sigma^2\big) + 24\beta^2_{t+1}\hat{L}^2\sum_{m=1}^M\big(\mathbb{E}\|x^m_t-\bar{x}_t\|^2
  + \mathbb{E}\|y^m_t-\bar{y}_t\|^2\big) + 12\beta^2_{t+1}M\hat{\delta}^2 \bigg)  \nonumber \\
 & \leq (1+\nu)(1-\beta_{t+1})^2\sum_{m=1}^M\mathbb{E}\|A^{-1}_t(w^m_t - \bar{w}_t)\|^2 + (1+\frac{1}{\nu})\frac{1}{\rho^2}
 \bigg(4L^2_K\sum_{m=1}^M\mathbb{E}\big(\gamma^2\eta^2_t\|A^{-1}_tw^m_t\|^2+\lambda^2\eta^2_t\|B^{-1}_tv^m_t\|^2\big)  \nonumber \\
 & \quad + 8\beta^2_{t+1}\sum_{m=1}^M\big(\|R^m_t\|^2 + \sigma^2\big) + 24\beta^2_{t+1}\hat{L}^2\big((q-1)\sum_{l = s_t+1}^{t-1} \gamma^2\eta_l^2 \sum_{m = 1}^M \mathbb{E}\|A^{-1}_l(w^m_l - \bar{w}_l)\|^2 \nonumber \\
 & \quad + (q-1)\sum_{l = s_t+1}^{t-1} \lambda^2\eta_l^2 \sum_{m = 1}^M \mathbb{E}\|B^{-1}_l(v^m_l - \bar{v}_l)\|^2\big) + 12\beta^2_{t+1}M\hat{\delta}^2 \bigg)  \nonumber \\
 & \leq (1+\nu)(1-\beta_{t+1})^2\sum_{m=1}^M\mathbb{E}\|A^{-1}_t(w^m_t - \bar{w}_t)\|^2 + (1+\frac{1}{\nu})\frac{1}{\rho^2}
 \bigg(8L^2_K\sum_{m=1}^M\mathbb{E}\big(\gamma^2\eta^2_t\|A^{-1}_t(w^m_t-\bar{w}_t)\|^2+\lambda^2\eta^2_t\|B^{-1}_t(v^m_t-\bar{v}_t)\|^2\big)  \nonumber \\
 & \quad + 8L^2_K\sum_{m=1}^M\mathbb{E}\big(\gamma^2\eta^2_t\|A^{-1}_t\bar{w}_t\|^2+\lambda^2\eta^2_t\|B^{-1}_t\bar{v}_t\|^2\big)
 + 8\beta^2_{t+1}\sum_{m=1}^M\big(\|R^m_t\|^2 + \sigma^2\big) \nonumber \\
 & \quad + 24\beta^2_{t+1}\hat{L}^2\big((q-1)\sum_{l = s_t+1}^{t-1} \gamma^2\eta_l^2 \sum_{m = 1}^M \mathbb{E}\|A^{-1}_l(w^m_l - \bar{w}_l)\|^2
 + (q-1)\sum_{l = s_t+1}^{t-1} \lambda^2\eta_l^2 \sum_{m = 1}^M \mathbb{E}\|B^{-1}_l(v^m_l - \bar{v}_l)\|^2\big) + 12\beta^2_{t+1}M\hat{\delta}^2 \bigg),  \nonumber
\end{align}
where the second inequality holds by the above Lemma \ref{lem:A11}.

Similarly, we can also obtain
\begin{align} \label{eq:A62}
 & \sum_{m=1}^M\mathbb{E}\|B^{-1}_{t+1}(v^m_{t+1} - \bar{v}_{t+1})\|^2  \\
 & \leq (1+\nu)(1-\alpha_{t+1})^2\sum_{m=1}^M\mathbb{E}\|B^{-1}_t(v^m_t - \bar{v}_t)\|^2 + (1+\frac{1}{\nu})\frac{1}{\rho^2}
 \bigg(4L^2_g\sum_{m=1}^M\mathbb{E}\big(\|x^m_{t+1}-x^m_t\|^2+\|y^m_{t+1}-y^m_t\|^2\big)  \nonumber \\
 & \quad + 4M\alpha^2_{t+1}\sigma^2 + 24\alpha^2_{t+1}L^2_g\sum_{m=1}^M\big(\mathbb{E}\|x^m_t-\bar{x}_t\|^2
  + \mathbb{E}\|y^m_t-\bar{y}_t\|^2\big) + 12\alpha^2_{t+1}M\delta_{g}^2 \bigg) \nonumber \\
 & \leq (1+\nu)(1-\alpha_{t+1})^2\sum_{m=1}^M\mathbb{E}\|B^{-1}_t(v^m_t - \bar{v}_t)\|^2 + (1+\frac{1}{\nu})\frac{1}{\rho^2}
 \bigg(8L^2_g\sum_{m=1}^M\mathbb{E}\big(\gamma^2\eta^2_t\|A^{-1}_t(w^m_t-\bar{w}_t)\|^2+\lambda^2\eta^2_t\|B^{-1}_t(v^m_t-\bar{v}_t)\|^2\big)  \nonumber \\
 & \quad + 8L^2_g\sum_{m=1}^M\mathbb{E}\big(\gamma^2\eta^2_t\|A^{-1}_t\bar{w}_t\|^2+\lambda^2\eta^2_t\|B^{-1}_t\bar{v}_t\|^2\big) + 4M\alpha^2_{t+1}\sigma^2 \nonumber\\
 & \quad + 24\alpha^2_{t+1}L^2_g\big( (q-1)\sum_{l = s_t+1}^{t-1} \gamma^2\eta_l^2 \sum_{m = 1}^M \mathbb{E}\|A^{-1}_l(w^m_l - \bar{w}_l)\|^2
 + (q-1)\sum_{l = s_t+1}^{t-1} \lambda^2\eta_l^2 \sum_{m = 1}^M \mathbb{E}\|B^{-1}_l(v^m_l - \bar{v}_l)\|^2 \big) + 12\alpha^2_{t+1}M\delta_{g}^2 \bigg). \nonumber
\end{align}

By combining the above inequalities \eqref{eq:A61} with \eqref{eq:A62}, we have
\begin{align} \label{eq:A63}
 & \sum_{m=1}^M\mathbb{E}\big( \|A^{-1}_{t+1}(w^m_{t+1} - \bar{w}_{t+1})\|^2 + \|B^{-1}_{t+1}(v^m_{t+1} - \bar{v}_{t+1})\|^2 \big) \\
 & \leq (1+\nu)(1-\beta_{t+1})^2\sum_{m=1}^M\mathbb{E}\|A^{-1}_t(w^m_t - \bar{w}_t)\|^2 + (1+\frac{1}{\nu})\frac{1}{\rho^2}
 \bigg(8L^2_K\sum_{m=1}^M\mathbb{E}\big(\gamma^2\eta^2_t\|A^{-1}_t(w^m_t-\bar{w}_t)\|^2+\lambda^2\eta^2_t\|B^{-1}_t(v^m_t-\bar{v}_t)\|^2\big)  \nonumber \\
 & \quad + 8L^2_K\sum_{m=1}^M\mathbb{E}\big(\gamma^2\eta^2_t\|A^{-1}_t\bar{w}_t\|^2+\lambda^2\eta^2_t\|B^{-1}_t\bar{v}_t\|^2\big)
 + 8\beta^2_{t+1}\sum_{m=1}^M\big(\|R^m_t\|^2 + \sigma^2\big) \nonumber \\
 & \quad + 24\beta^2_{t+1}\hat{L}^2\big((q-1)\sum_{l = s_t+1}^{t-1} \gamma^2\eta_l^2 \sum_{m = 1}^M \mathbb{E}\|A^{-1}_l(w^m_l - \bar{w}_l)\|^2
 + (q-1)\sum_{l = s_t+1}^{t-1} \lambda^2\eta_l^2 \sum_{m = 1}^M \mathbb{E}\|B^{-1}_l(v^m_l - \bar{v}_l)\|^2\big) + 12\beta^2_{t+1}M\hat{\delta}^2 \bigg)  \nonumber \\
 & \quad + (1+\nu)(1-\alpha_{t+1})^2\sum_{m=1}^M\mathbb{E}\|B^{-1}_t(v^m_t - \bar{v}_t)\|^2 + (1+\frac{1}{\nu})\frac{1}{\rho^2}
 \bigg(8L^2_g\sum_{m=1}^M\mathbb{E}\big(\gamma^2\eta^2_t\|A^{-1}_t(w^m_t-\bar{w}_t)\|^2+\lambda^2\eta^2_t\|B^{-1}_t(v^m_t-\bar{v}_t)\|^2\big)  \nonumber \\
 & \quad + 8L^2_g\sum_{m=1}^M\mathbb{E}\big(\gamma^2\eta^2_t\|A^{-1}_t\bar{w}_t\|^2+\lambda^2\eta^2_t\|B^{-1}_t\bar{v}_t\|^2\big) + 4M\alpha^2_{t+1}\sigma^2 \nonumber\\
 & \quad + 24\alpha^2_{t+1}L^2_g\big( (q-1)\sum_{l = s_t+1}^{t-1} \gamma^2\eta_l^2 \sum_{m = 1}^M \mathbb{E}\|A^{-1}_l(w^m_l - \bar{w}_l)\|^2
 + (q-1)\sum_{l = s_t+1}^{t-1} \lambda^2\eta_l^2 \sum_{m = 1}^M \mathbb{E}\|B^{-1}_l(v^m_l - \bar{v}_l)\|^2 \big) + 12\alpha^2_{t+1}M\delta_{g}^2 \bigg) \nonumber \\
 & \leq \max\bigg((1+\nu)(1-\beta_{t+1})^2+8\gamma^2\eta^2_t(1+\frac{1}{\nu})\frac{1}{\rho^2}(L^2_K+L^2_g), (1+\nu)(1-\alpha_{t+1})^2+8\lambda^2\eta^2_t(1+\frac{1}{\nu})\frac{1}{\rho^2}(L^2_K+L^2_g) \bigg) \nonumber \\
 & \quad \cdot\sum_{m=1}^M\mathbb{E}\big( \|A^{-1}_t(w^m_t - \bar{w}_t)\|^2 + \|B^{-1}_t(v^m_t - \bar{v}_t)\|^2 \big) \nonumber \\
 & \quad + 8\eta^2_t(L^2_K+L^2_g)(1+\frac{1}{\nu})\frac{1}{\rho^2}\sum_{m=1}^M\mathbb{E}\big(\gamma^2\|A^{-1}_t\bar{w}_t\|^2+\lambda^2\|B^{-1}_t\bar{v}_t\|^2\big) \nonumber \\
 & \quad + (1+\frac{1}{\nu})\frac{1}{\rho^2}\bigg(8\beta^2_{t+1}\sum_{m=1}^M\|R^m_t\|^2 + 8M\beta^2_{t+1}\sigma^2
 + 4M\alpha^2_{t+1}\sigma^2 + 12\beta^2_{t+1}M\hat{\delta}^2 + 12\alpha^2_{t+1}M\delta_{g}^2 \bigg) \nonumber \\
 & \quad + 24(q-1)(1+\frac{1}{\nu})\frac{1}{\rho^2}(\beta^2_{t+1}\hat{L}^2+\alpha^2_{t+1}L^2_g)\max(\gamma^2,\lambda^2)\sum_{l = s_t+1}^{t-1}\eta_l^2 \sum_{m = 1}^M
 \Big( \mathbb{E}\|A^{-1}_l(w^m_l - \bar{w}_l)\|^2 + \mathbb{E}\|B^{-1}_l(v^m_l - \bar{v}_l)\|^2 \Big).  \nonumber
\end{align}

Let $\nu=\frac{1}{q}$ and $\gamma = \tau\lambda \ (0<\tau \leq 1)$, i.e., $\gamma \leq \lambda$. Since $M\geq 1$ and $0<\theta\leq 1$,
we have $\eta_t \leq \frac{\rho\theta}{12\lambda q\sqrt{M}\sqrt{L^2_K+L^2_g}} \leq \frac{\rho}{12\lambda q\sqrt{L^2_K+L^2_g}}$ for all $t\geq 1$.
Since $\alpha_{t+1}\in (0,1)$ and $\beta_{t+1}\in (0,1)$ for all $t\geq 0$, we have
\begin{align}
 & (1+\nu)(1-\beta_{t+1})^2+8\gamma^2\eta^2_t(1+\frac{1}{\nu})\frac{1}{\rho^2}(L^2_K+L^2_g) \nonumber \\
 & \leq 1+ \frac{1}{q} + 8(1+q)\frac{\gamma^2}{\rho^2}(L^2_K+L^2_g)\frac{\rho^2}{144\lambda^2q^2(L^2_K+L^2_g)} \nonumber \\
 & \leq 1+ \frac{1}{q} + \frac{\gamma^2}{\lambda^2}\frac{1+q}{18q^2} \leq 1+ \frac{10}{9q}.
\end{align}
Similarly, we can also obtain $(1+\nu)(1-\alpha_{t+1})^2+8\lambda^2\eta^2_t(1+\frac{1}{\nu})\frac{1}{\rho^2}(L^2_K+L^2_g) \leq 1+ \frac{10}{9q}$.
Thus, we have
\begin{align} \label{eq:A64}
 & \sum_{m=1}^M\mathbb{E}\big( \|A^{-1}_{t+1}(w^m_{t+1} - \bar{w}_{t+1})\|^2 + \|B^{-1}_{t+1}(v^m_{t+1} - \bar{v}_{t+1})\|^2 \big) \\
 & \leq \big(1+ \frac{10}{9q} \big)\sum_{m=1}^M\mathbb{E}\big( \|A^{-1}_t(w^m_t - \bar{w}_t)\|^2 + \|B^{-1}_t(v^m_t - \bar{v}_t)\|^2 \big) \nonumber \\
 & \quad + 8\eta^2_t(L^2_K+L^2_g)(1+q)\frac{\lambda^2}{\rho^2}\sum_{m=1}^M\mathbb{E}\big(\tau^2\|A^{-1}_t\bar{w}_t\|^2+\|B^{-1}_t\bar{v}_t\|^2\big) \nonumber \\
 & \quad + (1+q)\frac{1}{\rho^2}\big(8\beta^2_{t+1}\sum_{m=1}^M\|R^m_t\|^2 + 8M\beta^2_{t+1}\sigma^2 + 4M\alpha^2_{t+1}\sigma^2 + 12\beta^2_{t+1}M\hat{\delta}^2 + 12\alpha^2_{t+1}M\delta_{g}^2 \big) \nonumber \\
 & \quad + 24q^2\frac{\lambda^2}{\rho^2}(\beta^2_{t+1}\hat{L}^2+\alpha^2_{t+1}L^2_g)\sum_{l = s_t+1}^{t-1}\eta_l^2 \sum_{m = 1}^M
 \big( \mathbb{E}\|A^{-1}_l(w^m_l - \bar{w}_l)\|^2 + \mathbb{E}\|B^{-1}_l(v^m_l - \bar{v}_l)\|^2 \big) \nonumber \\
 & \leq \big(1+ \frac{10}{9q} \big)\sum_{m=1}^M\mathbb{E}\big( \|A^{-1}_t(w^m_t - \bar{w}_t)\|^2 + \|B^{-1}_t(v^m_t - \bar{v}_t)\|^2 \big) + \frac{1}{9q}\sum_{m=1}^M\mathbb{E}\big(\tau^2\|A^{-1}_t\bar{w}_t\|^2+\|B^{-1}_t\bar{v}_t\|^2\big) \nonumber \\
 & \quad + \frac{2M\eta^3_t}{3\rho\lambda\sqrt{L^2_K+L^2_g}}\big(\frac{2c^2_2}{T^2} + 2c^2_2\sigma^2 + c^2_1\sigma^2 + 3c^2_2\hat{\delta}^2 + 3c^2_1\delta_{g}^2 \big) \nonumber \\
 & \quad + (c^2_2\hat{L}^2 + c^2_1L^2_g)\frac{\eta^2_t}{6}\sum_{l = s_t+1}^{t-1}\eta_l^2 \sum_{m = 1}^M
 \big( \mathbb{E}\|A^{-1}_l(w^m_l - \bar{w}_l)\|^2 + \mathbb{E}\|B^{-1}_l(v^m_l - \bar{v}_l)\|^2 \big),
\end{align}
where the first inequality is due to $\alpha_{t+1}=c_1\eta^2_t$ and $\beta_{t+1}=c_2\eta^2_t$, and
the last inequality holds by $8\eta^2_t(L^2_K+L^2_g)(1+q)\frac{\lambda^2}{\rho^2}\leq \frac{1}{9q}$ and $\|R^m_t\|\leq \frac{1}{T}$ for any $m\in[M]$, $t\geq 1$.

Then we have
\begin{align} \label{eq:A65}
 & \sum_{m=1}^M\mathbb{E}\big( \|A^{-1}_{t+1}(w^m_{t+1} - \bar{w}_{t+1})\|^2 + \|B^{-1}_{t+1}(v^m_{t+1} - \bar{v}_{t+1})\|^2 \big) \\
 & \leq \frac{1}{9q}\sum_{s=s_t}^{t}\big(1+ \frac{10}{9q} \big)^{t-s_t}\sum_{m=1}^M\mathbb{E}\big(\tau^2\|A^{-1}_s\bar{w}_s\|^2+\|B^{-1}_s\bar{v}_s\|^2\big) \nonumber \\
 & \quad + \frac{2M}{3\rho\lambda\sqrt{L^2_K+L^2_g}}\big(\frac{2c^2_2}{T^2} + 2c^2_2\sigma^2 + c^2_1\sigma^2 + 3c^2_2\hat{\delta}^2 + 3c^2_1\delta_{g}^2 \big)
 \sum_{s=s_t}^t\big(1+ \frac{10}{9q} \big)^{t-s_t} \eta^3_s\nonumber \\
 & \quad + \frac{c^2_2\hat{L}^2 + c^2_1L^2_g}{6}\sum_{s=s_t}^t\big(1+ \frac{10}{9q} \big)^{t-s_t}\eta^2_s\sum_{l=s_t}^{s}\eta_l^2 \sum_{m = 1}^M
 \big( \mathbb{E}\|A^{-1}_l(w^m_l - \bar{w}_l)\|^2 + \mathbb{E}\|B^{-1}_l(v^m_l - \bar{v}_l)\|^2 \big) \nonumber \\
 & \leq \frac{1}{9q}\sum_{s=s_t}^{t}\big(1+ \frac{10}{9q} \big)^{q}\sum_{m=1}^M\mathbb{E}\big(\tau^2\|A^{-1}_s\bar{w}_s\|^2+\|B^{-1}_s\bar{v}_s\|^2\big) \nonumber \\
 & \quad + \frac{2M}{3\rho\lambda\sqrt{L^2_K+L^2_g}}\big(\frac{2c^2_2}{T^2} + 2c^2_2\sigma^2 + c^2_1\sigma^2 + 3c^2_2\hat{\delta}^2 + 3c^2_1\delta_{g}^2 \big)
 \sum_{s=s_t}^t\big(1+ \frac{10}{9q} \big)^{q} \eta^3_s\nonumber \\
 & \quad + \frac{c^2_2\hat{L}^2 + c^2_1L^2_g}{6}\sum_{s=s_t}^t\big(1+ \frac{10}{9q} \big)^{q}\eta^2_s\sum_{l=s_t}^{s}\eta_l^2 \sum_{m = 1}^M
 \big( \mathbb{E}\|A^{-1}_l(w^m_l - \bar{w}_l)\|^2 + \mathbb{E}\|B^{-1}_l(v^m_l - \bar{v}_l)\|^2 \big) \nonumber \\
 & \leq \frac{4M}{9q}\sum_{s=s_t}^{t+1} \mathbb{E}\big(\tau^2\|A^{-1}_s\bar{w}_s\|^2+\|B^{-1}_s\bar{v}_s\|^2\big) \nonumber \\
 & \quad + \frac{8M}{3\rho\lambda\sqrt{L^2_K+L^2_g}}\big(\frac{2c^2_2}{T^2} + 2c^2_2\sigma^2 + c^2_1\sigma^2 + 3c^2_2\hat{\delta}^2 + 3c^2_1\delta_{g}^2 \big)
 \sum_{s=s_t}^{t+1} \eta^3_s\nonumber \\
 & \quad + \frac{\rho^3(c^2_2\hat{L}^2 + c^2_1L^2_g)}{18*(12)^2\lambda^3q^2(L^2_K+L^2_g)^{3/2}}\sum_{s=s_t}^{t+1} \eta_s\sum_{m = 1}^M
 \big( \mathbb{E}\|A^{-1}_s(w^m_s - \bar{w}_s)\|^2 + \mathbb{E}\|B^{-1}_s(v^m_s - \bar{v}_s)\|^2 \big),
\end{align}
where the last inequality holds by $\big(1+ \frac{10}{9q} \big)^{q} \leq e^{10/9}\leq 4$.

By multiplying both sides of \eqref{eq:A65} by $\eta_{t+1}$ and summing over $t=s_t-1$ to $s_t+q-2$, we have
\begin{align} \label{eq:A66}
 & \sum_{t=s_t}^{s_t+q-1}\eta_t\sum_{m=1}^M\mathbb{E}\big( \|A^{-1}_t(w^m_t - \bar{w}_t)\|^2 + \|B^{-1}_t(v^m_t - \bar{v}_t)\|^2 \big) \\
 & \leq \frac{4M}{9}\sum_{t=s_t}^{s_t+q-1}\eta_t\mathbb{E}\big(\tau^2\|A^{-1}_t\bar{w}_t\|^2+\|B^{-1}_t\bar{v}_t\|^2\big) \nonumber \\
 & \quad + \frac{2M}{9\lambda^2(L^2_K+L^2_g)}\big(\frac{2c^2_2}{T^2} + 2c^2_2\sigma^2 + c^2_1\sigma^2 + 3c^2_2\hat{\delta}^2 + 3c^2_1\delta_{g}^2 \big)
 \sum_{t=s_t}^{s_t+q-1} \eta^3_t\nonumber \\
 & \quad + \frac{\rho^4(c^2_2\hat{L}^2 + c^2_1L^2_g)}{18*(12)^3\lambda^4q^2(L^2_K+L^2_g)^2}\sum_{t=s_t}^{s_t+q-1} \eta_t\sum_{m = 1}^M
 \big( \mathbb{E}\|A^{-1}_t(w^m_t - \bar{w}_t)\|^2 + \mathbb{E}\|B^{-1}_t(v^m_t - \bar{v}_t)\|^2 \big),
\end{align}

For notational simplicity, let $c_2=\vartheta c_1$ with $\vartheta>0$. Given $c_1\leq \frac{72\lambda^2q(L^2_K+L^2_g)}{\rho^2\sqrt{\vartheta^2\hat{L}^2+L^2_g}}$, we have
$\frac{60}{72}\leq 1-\frac{\rho^4(c^2_2\hat{L}^2 + c^2_1L^2_g)}{18*(12)^3\lambda^4q^2(L^2_K+L^2_g)^2}$,
we have
\begin{align}
 & \sum_{t=s_t}^{s_t+q-1}\eta_t\sum_{m=1}^M\mathbb{E}\big( \|A^{-1}_t(w^m_t - \bar{w}_t)\|^2 + \|B^{-1}_t(v^m_t - \bar{v}_t)\|^2 \big) \\
 & \leq \frac{8M}{15}\sum_{t=s_t}^{s_t+q-1}\eta_t\mathbb{E}\big(\tau^2\|A^{-1}_t\bar{w}_t\|^2+\|B^{-1}_t\bar{v}_t\|^2\big) \nonumber \\
 & \quad + \frac{4M}{15\lambda^2(L^2_K+L^2_g)}\big(\frac{2c^2_2}{T^2} + 2c^2_2\sigma^2 + c^2_1\sigma^2 + 3c^2_2\hat{\delta}^2 + 3c^2_1\delta_{g}^2 \big)
 \sum_{t=s_t}^{s_t+q-1} \eta^3_t.
\end{align}

\end{proof}

\begin{theorem}  \label{th:A1}
(Restatement of Theorem 1)
Suppose the sequence $\{\bar{x}_t,\bar{y}_t\}_{t=1}^T$ be generated from Algorithm \ref{alg:1}.
 Under the above Assumptions, and let $\eta_t=\frac{kM^{1/3}}{(n+t)^{1/3}}$ for all $t\geq 0$, $\alpha_{t+1}=c_1\eta_t^2$, $\beta_{t+1}=c_2\eta_t^2$, $n \geq \max\big(2, Mk^3, M(c_1k)^3, M(c_2k)^3, \frac{(12k\lambda q)^3M^{5/2}(L^2_K+L^2_g)^{3/2}}{(\theta\rho)^3}\big)$, $k>0$, $c_2=\vartheta c_1 \ (\vartheta>0)$,
 $\frac{2}{3k^3} + \frac{1000\bar{L}^2}{3\mu^2} \leq c_1 \leq \frac{72\lambda^2q(L^2_K+L^2_g)}{\rho^2\sqrt{\vartheta^2\hat{L}^2+L^2_g}}$, $c_2 \geq \frac{2}{3k^3} + 34$, $\lambda=\tau\gamma$,
 $0<\tau \leq \min\Big(\frac{1}{8}\sqrt{\frac{15M\rho\gamma}{\Gamma}},1\Big)$, $0<\theta\leq \min\Big(9\bar{L}\sqrt{\frac{75\lambda(L^2_K+L^2_g)M\mu}{\rho\big(30\hat{L}^2\mu^2
 +1000\bar{L}^2L^2_g + 52\hat{L}^2\mu^2\big)}},1\Big)$,
 $0< \lambda \leq \frac{225M\rho\bar{L}^2}{184\mu(L^2_K+L^2_g)}$, $0< \gamma \leq \min\Big(\frac{\rho}{8}\sqrt{\frac{1}{(125\bar{L}^2\kappa^2\hat{b}^2)/(6\mu^2\lambda^2)+(L^2_g+L^2_K)/M}}, \frac{n^{1/3}\rho}{4LkM^{1/3}} \Big)$ and $K=\frac{L_g}{\mu}\log(C_{gxy}C_{fy}T/\mu)$, we have
\begin{align}
 \frac{1}{T}\sum_{t=1}^T\mathbb{E}\|\nabla F(\bar{x}_t)\| \leq \Big( \frac{\sqrt{3G}n^{1/6}}{M^{1/6}T^{1/2}} + \frac{\sqrt{3G}}{M^{1/6}T^{1/3}}\Big)\sqrt{\frac{1}{T}\sum_{t=1}^T\mathbb{E}\|A_t\|^2},
\end{align}
where $G = \frac{4(F(\bar{x}_1) - F^*)}{k\rho\gamma} + \frac{160b_1\bar{L}^2\Delta_0}{k\lambda\mu\rho^2} + \frac{8n^{1/3}\sigma^2}{qM^{4/3}k^2\rho^2} + 8k^2\Big(\frac{(c^2_1+c^2_2)\sigma^2}{\rho^2}
 + \frac{2\hat{c}^2\Gamma }{15\rho\gamma\lambda^2(L^2_K+L^2_g)}\Big)\ln(n+T) + \frac{54kM^{1/3}(n+T)^{2/3}}{T^2\rho^2}$, $\Delta_0 = \|\bar{y}_1-y^*(\bar{x}_1)\|^2$,
 $\hat{c}^2 = \frac{2c^2_2}{T^2} + 2c^2_2\sigma^2 + c^2_1\sigma^2 + 3c^2_2\hat{\delta}^2 + 3c^2_1\delta_{g}^2$, $\hat{\delta}^2=4\delta^2_f + \frac{4C^2_{fy}\delta^2_{g}}{\mu^2} + \frac{4C^2_{gxy}C^2_{fy}\delta^2_{g}}{\mu^4} + \frac{4C^2_{gxy}\delta^2_{f}}{\mu^2}$ and $\Gamma = \frac{5\theta^2\hat{L}^2\gamma\rho}{36(L^2_K+L^2_g)} + \frac{125\theta^2\bar{L}^2 L^2_g\gamma \rho}{27\mu^2(L^2_K+L^2_g)} + \frac{8(L^2_K+L^2_g)\lambda^2\gamma}{\rho} + \frac{17\theta^2\hat{L}^2\rho\gamma }{72(L^2_K+L^2_g)}$.
\end{theorem}

\begin{proof}
Since $\eta_t=\frac{kM^{1/3}}{(n+t)^{1/3}}$ on $t$ is decreasing and $n\geq k^3M$, we have $\eta_t \leq \eta_0 = \frac{kM^{1/3}}{n^{1/3}} \leq 1$ and $\gamma \leq \frac{n^{1/3}\rho}{4LkM^{1/3}}\leq \frac{\rho}{2L\eta_0} \leq \frac{\rho}{2L\eta_t}$ for any $t\geq 0$. Since $\eta_t \leq \frac{\theta\rho}{12\lambda q\sqrt{M}\sqrt{L^2_K+L^2_g}} \ (0 < \theta \leq 1)$ for all $t\geq 0$, we have
$\frac{kM^{1/3}}{n^{1/3}} =\eta_0\leq \eta_t \leq \frac{\theta\rho}{12\lambda q\sqrt{M}\sqrt{L^2_K+L^2_g}} \ (0 < \theta \leq 1)$, then we have $n \geq \frac{(12k\lambda q)^3M^{5/2}(L^2_K+L^2_g)^{3/2}}{(\theta\rho)^3}$.
 Due to $0 < \eta_t \leq 1$ and $n\geq M(c_1k)^3$, we have $\alpha_{t+1} = c_1\eta_t^2 \leq c_1\eta_t \leq \frac{c_1kM^{1/3}}{n^{1/3}}\leq 1$.
Similarly, due to $n\geq M(c_2k)^3$, we have $\beta_{t+1}\leq 1$.

 According to Lemma \ref{lem:A7}, we have
 \begin{align}
  & \frac{1}{\eta_t}\mathbb{E} \|\bar{\nabla}_y g(x_{t+1},y_{t+1}) - \bar{v}_{t+1}\|^2 - \frac{1}{\eta_{t-1}}\mathbb{E} \|\bar{\nabla}_y g(x_t,y_t) - \bar{v}_t\|^2  \\
  & \leq \big(\frac{1-\alpha_{t+1}}{\eta_t} - \frac{1}{\eta_{t-1}}\big)\mathbb{E} \|\bar{\nabla}_y g(x_t,y_t) - \bar{v}_t\|^2 + \frac{4L^2_g}{M^2}\eta_t\sum_{m=1}^M\big(\|\hat{x}^m_{t+1}-x^m_t\|^2 + \|\hat{y}^m_{t+1}-y^m_t\|^2\big) + \frac{2\alpha_{t+1}^2\sigma^2}{M\eta_t}\nonumber \\
  & = \big(\frac{1}{\eta_t} - \frac{1}{\eta_{t-1}} - c_1\eta_t\big)\mathbb{E} \|\bar{\nabla}_y g(x_t,y_t) - \bar{v}_t\|^2  + \frac{4L^2_g}{M^2}\eta_t\sum_{m=1}^M\big(\|\hat{x}^m_{t+1}-x^m_t\|^2 +\|\hat{y}^m_{t+1}-y^m_t\|^2\big) + \frac{2c_1^2\eta^3_t\sigma^2}{M}, \nonumber
 \end{align}
 where the second equality is due to $\alpha_{t+1}=c_1\eta^2_t$.
 Similarly, we have
 \begin{align}
 & \frac{1}{\eta_t}\mathbb{E}\|\bar{w}_{t+1} -\bar{\nabla} f(x_{t+1},y_{t+1}) - R_{t+1}\|^2 - \frac{1}{\eta_{t-1}}\mathbb{E}\|\bar{w}_t - \bar{\nabla} f(x_t,y_t) - R_t\|^2 \\
 & \leq \big(\frac{1-\beta_{t+1}}{\eta_t} -\frac{1}{\eta_{t-1}}\big) \mathbb{E}\|\bar{w}_t - \bar{\nabla}f(x_t,y_t) -R_t\|^2 + \frac{4L^2_K}{M^2}\eta_t\sum_{m=1}^M\big( \|\hat{x}^m_{t+1}-x^m_t\|^2 + \|\hat{y}^m_{t+1}-y^m_t\|^2 \big) + \frac{2\beta^2_{t+1}\sigma^2}{M\eta_t} \nonumber \\
 & =  \big(\frac{1}{\eta_t} -\frac{1}{\eta_{t-1}} -c_2\eta_t\big) \mathbb{E}\|\bar{w}_t - \bar{\nabla}f(x_t,y_t) -R_t\|^2 + \frac{4L^2_K}{M^2}\eta_t\sum_{m=1}^M\big( \|\hat{x}_{t+1}^m-x^m_t\|^2 + \|\hat{y}^m_{t+1}-y^m_t\|^2 \big) + \frac{2c^2_2\eta^3_t\sigma^2}{M}. \nonumber
 \end{align}
By $\eta_t = \frac{kM^{1/3}}{(n+t)^{1/3}}$, we have
 \begin{align}
  \frac{1}{\eta_t} - \frac{1}{\eta_{t-1}} &= \frac{1}{M^{1/3}k}\big( (n+t)^{\frac{1}{3}} - (n+t-1)^{\frac{1}{3}}\big) \leq \frac{1}{3M^{1/3}k(n+t-1)^{2/3}} \leq \frac{1}{3M^{1/3}k\big(n/2+t\big)^{2/3}} \nonumber \\
  & \leq \frac{2^{2/3}}{3M^{1/3}k(n+t)^{2/3}} \leq \frac{2^{2/3}M^{2/3}}{3k^3}\frac{k^2}{(n+t)^{2/3}} = \frac{2^{2/3}}{3k^3}\eta_t^2 \leq \frac{2}{3k^3}\eta_t,
 \end{align}
 where the first inequality holds by the concavity of function $f(x)=x^{1/3}$, \emph{i.e.}, $(x+y)^{1/3}\leq x^{1/3} + \frac{y}{3x^{2/3}}$; the second inequality is due to $n\geq 2$,  and
 the last inequality is due to $0<\eta_t\leq 1$.

Let $c_1 \geq \frac{2}{3k^3} + \frac{1000\bar{L}^2}{3\mu^2}$, we have
 \begin{align} \label{eq:W1}
  & \frac{1}{\eta_t}\mathbb{E} \|\bar{\nabla}_y g(x_{t+1},y_{t+1}) - \bar{v}_{t+1}\|^2 - \frac{1}{\eta_{t-1}}\mathbb{E} \|\bar{\nabla}_y g(x_t,y_t) - \bar{v}_t\|^2  \nonumber \\
  & \leq -\frac{1000\bar{L}^2\eta_t}{3\mu^2} \mathbb{E} \|\bar{\nabla}_y g(x_t,y_t) - \bar{v}_t\|^2 + \frac{4L^2_g}{M^2}\eta_t\sum_{m=1}^M\big(\|\hat{x}^m_{t+1}-x^m_t\|^2 +\|\hat{y}^m_{t+1}-y^m_t\|^2\big)
  + \frac{2c_1^2\eta^3_t\sigma^2}{M} \nonumber \\
  & = -\frac{1000\bar{L}^2\eta_t}{3\mu^2} \mathbb{E} \|\bar{\nabla}_y g(x_t,y_t) - \bar{v}_t\|^2 + \frac{4L^2_g}{M^2}\eta_t\sum_{m=1}^M\big(\gamma^2\|A^{-1}_t(w^m_t-\bar{w}_t+\bar{w}_t)\|^2 + \lambda^2\|B^{-1}_t(v^m_t-\bar{v}_t+\bar{v}_t)\|^2\big) \nonumber \\
  & \quad + \frac{2c_1^2\eta^3_t\sigma^2}{M} \nonumber \\
  & \leq -\frac{1000\bar{L}^2\eta_t}{3\mu^2} \mathbb{E} \|\bar{\nabla}_y g(x_t,y_t) - \bar{v}_t\|^2 + \frac{8L^2_g}{M^2}\eta_t\sum_{m=1}^M\big(\gamma^2\|A^{-1}_t(w^m_t-\bar{w}_t)\|^2 + \gamma^2\|A^{-1}_t\bar{w}_t\|^2 \nonumber \\
  & \quad + \lambda^2\|B^{-1}_t(v^m_t-\bar{v}_t)\|^2 + \lambda^2\|B^{-1}_t\bar{v}_t\|^2\big) + \frac{2c_1^2\eta^3_t\sigma^2}{M}.
 \end{align}
Let $c_2 \geq \frac{2}{3k^3} + 34$, we have
 \begin{align} \label{eq:W2}
  & \frac{1}{\eta_t}\mathbb{E}\|\bar{w}_{t+1} - \bar{\nabla} f(x_{t+1},y_{t+1}) - R_{t+1}\|^2 - \frac{1}{\eta_{t-1}}\mathbb{E}\|\bar{w}_t - \bar{\nabla} f(x_t,y_t) - R_t\|^2 \nonumber \\
  & \leq -34\eta_t \mathbb{E}\|\bar{w}_t - \bar{\nabla} f(x_t,y_t) - R_t\|^2 +
   \frac{4L^2_K}{M^2}\eta_t\sum_{m=1}^M\big(\|\hat{x}^m_{t+1}-x^m_t\|^2 +\|\hat{y}^m_{t+1}-y^m_t\|^2\big) + \frac{2c_2^2\eta_t^3\sigma^2}{M} \nonumber \\
  & = -34\eta_t \mathbb{E}\|\bar{w}_t - \bar{\nabla} f(x_t,y_t) - R_t\|^2 +
   \frac{4L^2_K}{M^2}\eta_t\sum_{m=1}^M\big(\gamma^2\|A^{-1}_t(w^m_t-\bar{w}_t+\bar{w}_t)\|^2 + \lambda^2\|B^{-1}_t(v^m_t-\bar{v}_t+\bar{v}_t)\|^2\big) \nonumber \\
  & \quad + \frac{2c_2^2\eta_t^3\sigma^2}{M} \nonumber \\
  & \leq -34\eta_t \mathbb{E}\|\bar{w}_t - \bar{\nabla} f(x_t,y_t) - R_t\|^2 + \frac{8L^2_K}{M^2}\eta_t\sum_{m=1}^M\big(\gamma^2\|A^{-1}_t(w^m_t-\bar{w}_t)\|^2 + \gamma^2\|A^{-1}_t\bar{w}_t\|^2 \nonumber \\
  & \quad + \lambda^2\|B^{-1}_t(v^m_t-\bar{v}_t)\|^2 + \lambda^2\|B^{-1}_t\bar{v}_t\|^2\big) + \frac{2c_2^2\eta_t^3\sigma^2}{M}.
 \end{align}

According to Lemmas \ref{lem:A5} and \ref{lem:A11}, we have
\begin{align} \label{eq:W3}
    F(\bar{x}_{t+1})-F(\bar{x}_t) & \leq \frac{8\bar{L}^2\gamma\eta_t}{\rho}\|y^*(\bar{x}_t)-\bar{y}_t\|^2 +\frac{2\gamma\eta_t}{\rho}\|\bar{w}_t-\hat{\nabla} f(\bar{x}_t,\bar{y}_t)\|^2 -\frac{\rho\eta_t}{2\gamma}\|\hat{x}_{t+1}-\bar{x}_t\|^2 \nonumber \\
    & = \frac{8\bar{L}^2\gamma\eta_t}{\rho}\|y^*(\bar{x}_t)-\bar{y}_t\|^2 +\frac{2\gamma\eta_t}{\rho}\|\bar{w}_t -\bar{\nabla} f(x_t,y_t) +
    \bar{\nabla} f(x_t,y_t)- \hat{\nabla} f(\bar{x}_t,\bar{y}_t) \|^2 \nonumber \\
    & \quad -\frac{\rho\eta_t}{2\gamma}\|\hat{x}_{t+1}-\bar{x}_t\|^2 \nonumber \\
    & \leq \frac{8\bar{L}^2\gamma\eta_t}{\rho}\|y^*(\bar{x}_t)-\bar{y}_t\|^2 + \frac{4\gamma\eta_t}{\rho}\|\bar{w}_t -\bar{\nabla} f(x_t,y_t)\|^2 +
     \frac{4\gamma\eta_t}{\rho}\|\bar{\nabla} f(x_t,y_t)- \hat{\nabla} f(\bar{x}_t,\bar{y}_t) \|^2 \nonumber \\
    & \quad -\frac{\rho\eta_t}{2\gamma}\|\hat{x}_{t+1}-\bar{x}_t\|^2 \nonumber \\
    & \leq \frac{8\bar{L}^2\gamma\eta_t}{\rho}\|y^*(\bar{x}_t)-\bar{y}_t\|^2 + \frac{8\gamma\eta_t}{\rho}\|\bar{w}_t -\bar{\nabla} f(x_t,y_t) - R_t\|^2 + \frac{8\gamma\eta_t}{\rho}\|R_t\|^2 \nonumber \\
    & \quad + \frac{4\gamma\eta_t}{\rho}\|\frac{1}{M}\sum_{m=1}^M (\hat{\nabla} f^m(x^m_t,y^m_t)- \hat{\nabla} f^m(\bar{x}_t,\bar{y}_t)) \|^2 - \frac{\rho\eta_t}{2\gamma}\|\hat{x}_{t+1}-\bar{x}_t\|^2  \nonumber\\
    & \leq \frac{8\bar{L}^2\gamma\eta_t}{\rho}\|y^*(\bar{x}_t)-\bar{y}_t\|^2 + \frac{8\gamma\eta_t}{\rho}\|\bar{w}_t -\bar{\nabla} f(x_t,y_t) - R_t\|^2 + \frac{8\gamma\eta_t}{\rho}\|R_t\|^2 \nonumber \\
    & \quad + \frac{4\gamma\eta_t\hat{L}^2}{M\rho}\sum_{m=1}^M\big(\|x^m_t-\bar{x}_t\|^2 + \|y^m_t-\bar{y}_t\|^2 \big) - \frac{\rho\eta_t}{2\gamma}\|\hat{x}_{t+1}-\bar{x}_t\|^2  \nonumber\\
    & \leq \frac{8\bar{L}^2\gamma\eta_t}{\rho}\|y^*(\bar{x}_t)-\bar{y}_t\|^2 + \frac{8\gamma\eta_t}{\rho}\|\bar{w}_t -\bar{\nabla} f(x_t,y_t) - R_t\|^2 + \frac{8\gamma\eta_t}{\rho}\|R_t\|^2 \nonumber \\
    & \quad + \frac{4(q-1)\gamma\eta_t\hat{L}^2}{M\rho}\Big( \sum_{l = s_t+1}^{t-1} \gamma^2\eta_l^2 \sum_{m = 1}^M \mathbb{E}\|A^{-1}_l(w^m_l - \bar{w}_l)\|^2 +
    \sum_{l = s_t+1}^{t-1} \lambda^2\eta_l^2 \sum_{m = 1}^M \mathbb{E}\|B^{-1}_l(v^m_l - \bar{v}_l)\|^2 \Big) \nonumber \\
    & \quad - \frac{\rho\eta_t\gamma}{2}\|A^{-1}_t\bar{w}_t\|^2,
\end{align}
where the last inequality holds by the above Lemma \ref{lem:A11}, and $\hat{x}_{t+1}=\frac{1}{M}\sum_{m=1}^M\hat{x}^m_{t+1}$, $\bar{x}_t = \frac{1}{M}\sum_{m=1}^Mx^m_t$
and $\hat{x}_{t+1}-\bar{x}_t = \frac{1}{M}\sum_{m=1}^M(\hat{x}^m_{t+1}-x^m_t) = \frac{1}{M}\sum_{m=1}^M(-\gamma A^{-1}_tw^m_t)=-\gamma A^{-1}_t\bar{w}_t$.

According to Lemma \ref{lem:A6}, we have
\begin{align} \label{eq:W4}
  & \|\bar{y}_{t+1} - y^*(\bar{x}_{t+1})\|^2-\|\bar{y}_t -y^*(\bar{x}_t)\|^2 \nonumber \\
  & \leq -\frac{\eta_t\mu\lambda}{4b_t}\|\bar{y}_t -y^*(\bar{x}_t)\|^2 -\frac{3\eta_t}{4} \|\hat{y}_{t+1}-\bar{y}_t\|^2
   + \frac{25\eta_t\lambda}{6\mu b_t} \|\nabla_y g(\bar{x}_t,\bar{y}_t)-\bar{v}_t\|^2 + \frac{25\kappa^2\eta_tb_t}{6\mu\lambda}\|\hat{x}_{t+1} - \bar{x}_t\|^2 \nonumber \\
  & = -\frac{\eta_t\mu\lambda}{4b_t}\|\bar{y}_t -y^*(\bar{x}_t)\|^2 -\frac{3\eta_t\lambda^2}{4} \|B^{-1}_t\bar{v}_t\|^2
   + \frac{25\eta_t\lambda}{6\mu b_t} \|\nabla_y g(\bar{x}_t,\bar{y}_t) - \bar{\nabla}_y g(x_t,y_t) + \bar{\nabla}_y g(x_t,y_t) -\bar{v}_t\|^2 \nonumber \\
  & \quad + \frac{25\kappa^2\eta_tb_t\gamma^2}{6\mu\lambda}\|A^{-1}_t\bar{w}_t\|^2 \nonumber \\
  & \leq -\frac{\eta_t\mu\lambda}{4b_t}\|\bar{y}_t -y^*(\bar{x}_t)\|^2 -\frac{3\eta_t\lambda^2}{4} \|B^{-1}_t\bar{v}_t\|^2
   + \frac{25\eta_t\lambda}{3\mu b_t} \|\frac{1}{M}\sum_{m=1}^M (\nabla_y g^m(\bar{x}_t,\bar{y}_t) - \nabla_y g^m(x^m_t,y^m_t))\|^2 \nonumber \\
  & \quad + \frac{25\eta_t\lambda}{3\mu b_t}\|\bar{\nabla}_y g(x_t,y_t)-\bar{v}_t\|^2 + \frac{25\kappa^2\eta_tb_t\gamma^2}{6\mu\lambda}\|A^{-1}_t\bar{w}_t\|^2 \nonumber \\
  & \leq -\frac{\eta_t\mu\lambda}{4b_t}\|\bar{y}_t -y^*(\bar{x}_t)\|^2 -\frac{3\eta_t\lambda^2}{4} \|B^{-1}_t\bar{v}_t\|^2
   + \frac{50\eta_t\lambda L^2_g}{3\mu b_tM}\sum_{m=1}^M \big( \|\bar{x}_t-x^m_t\|^2 + \|\bar{y}_t-y^m_t\|^2 \big) \nonumber \\
  & \quad + \frac{25\eta_t\lambda}{3\mu b_t}\|\bar{\nabla}_y g(x_t,y_t)-\bar{v}_t\|^2 + \frac{25\kappa^2\eta_tb_t\gamma^2}{6\mu\lambda}\|A^{-1}_t\bar{w}_t\|^2 \nonumber \\
  & \leq -\frac{\eta_t\mu\lambda}{4b_t}\|\bar{y}_t -y^*(\bar{x}_t)\|^2 -\frac{3\eta_t\lambda^2}{4} \|B^{-1}_t\bar{v}_t\|^2
   + (q-1)\frac{50\eta_t\lambda L^2_g}{3\mu b_tM} \bigg( \sum_{l=s_t}^{t-1} \gamma^2\eta_l^2\sum_{m=1}^M \mathbb{E}\|A^{-1}_l(w^m_l-\bar{w}_l)\|^2 \nonumber \\
  & \quad + \sum_{l=s_t}^{t-1} \lambda^2\eta_l^2\sum_{m=1}^M \mathbb{E}\|B^{-1}_l(v^m_l - \bar{v}_l)\|^2 \bigg) + \frac{25\eta_t\lambda}{3\mu b_t}\|\bar{\nabla}_y g(x_t,y_t)-\bar{v}_t\|^2 + \frac{25\kappa^2\eta_tb_t\gamma^2}{6\mu\lambda}\|A^{-1}_t\bar{w}_t\|^2,
\end{align}
where the last inequality holds by the above Lemma \ref{lem:A11}.

Next, we define a \emph{Lyapunov} function, for any $t\geq 1$
\begin{align}
 \Omega_t & = \mathbb{E}\Big [F(\bar{x}_t) + \frac{40b_t\bar{L}^2\gamma}{\lambda\mu\rho}\|\bar{y}_t-y^*(\bar{x}_t)\|^2 + \frac{\gamma}{\rho\eta_{t-1}} \big(\|\bar{v}_t - \bar{\nabla}_y g(x_t,y_t)\|^2
 + \|\bar{w}_t - \bar{\nabla}f(x_t,y_t)-R_t\|^2 \big) \Big]. \nonumber
\end{align}
Then we have
 \begin{align} \label{eq:W5}
 & \Omega_{t+1} - \Omega_t \nonumber \\
 & = F(\bar{x}_{t+1}) - F(\bar{x}_t) + \frac{40b_t\bar{L}^2\gamma}{\lambda\mu\rho}
 \big(\|\bar{y}_{t+1}-y^*(\bar{x}_{t+1})\|^2 - \|\bar{y}_t-y^*(\bar{x}_t)\|^2 \big)
 + \frac{\gamma}{\rho} \bigg( \frac{1}{\eta_t}\mathbb{E}\|\bar{v}_{t+1} - \bar{\nabla}_y g(x_{t+1},y_{t+1})\|^2 \nonumber \\
 & \quad - \frac{1}{\eta_{t-1}}\mathbb{E}\|\bar{v}_t - \bar{\nabla}_y g(x_t,y_t)\|^2 + \frac{1}{\eta_t}\mathbb{E}\|\bar{w}_{t+1} - \bar{\nabla} f(x_{t+1},y_{t+1})-R_{t+1}\|^2
 - \frac{1}{\eta_{t-1}}\mathbb{E}\|\bar{w}_t - \bar{\nabla} f(x_t,y_t)-R_t\|^2 \bigg) \nonumber \\
 & \leq \frac{8\bar{L}^2\gamma\eta_t}{\rho}\|\bar{y}_t - y^*(\bar{x}_t)\|^2 + \frac{8\gamma\eta_t}{\rho}\|\bar{w}_t -\bar{\nabla} f(x_t,y_t) - R_t\|^2 + \frac{8\gamma\eta_t}{\rho}\|R_t\|^2
 - \frac{\rho\eta_t\gamma}{2}\|A^{-1}_t\bar{w}_t\|^2 \nonumber \\
 & \quad + \frac{4(q-1)\gamma\eta_t\hat{L}^2}{M\rho}\bigg(\sum_{l=s_t}^{t-1} \gamma^2\eta_l^2 \sum_{m = 1}^M \mathbb{E}\|A^{-1}_l(w^m_l - \bar{w}_l)\|^2 +
    \sum_{l=s_t}^{t-1} \lambda^2\eta_l^2 \sum_{m = 1}^M \mathbb{E}\|B^{-1}_l(v^m_l - \bar{v}_l)\|^2\bigg) \nonumber \\
 & \quad + \frac{40b_t\bar{L}^2\gamma}{\lambda\mu\rho} \Bigg( -\frac{\eta_t\mu\lambda}{4b_t}\|\bar{y}_t -y^*(\bar{x}_t)\|^2 -\frac{3\eta_t\lambda^2}{4} \|B^{-1}_t\bar{v}_t\|^2
   + (q-1)\frac{50\eta_t\lambda L^2_g}{3\mu b_tM} \Big( \sum_{l=s_t}^{t-1} \gamma^2\eta_l^2 \sum_{m = 1}^M \mathbb{E}\|A^{-1}_l(w^m_l - \bar{w}_l)\|^2 \nonumber \\
  & \quad + \sum_{l=s_t}^{t-1} \lambda^2\eta_l^2 \sum_{m = 1}^M \mathbb{E}\|B^{-1}_l(v^m_l - \bar{v}_l)\|^2 \Big) + \frac{25\eta_t\lambda}{3\mu b_t}\|\bar{\nabla}_y g(x_t,y_t)-\bar{v}_t\|^2 + \frac{25\kappa^2\eta_tb_t\gamma^2}{6\mu\lambda}\|A^{-1}_t\bar{w}_t\|^2\Bigg)  \nonumber \\
 & \quad + \frac{\gamma}{\rho} \bigg( -\frac{1000\bar{L}^2\eta_t}{3\mu^2} \mathbb{E} \|\bar{\nabla}_y g(x_t,y_t) - \bar{v}_t\|^2 + \frac{8L^2_g}{M^2}\eta_t\sum_{m=1}^M\big(\gamma^2\|A^{-1}_t(w^m_t-\bar{w}_t)\|^2 + \gamma^2\|A^{-1}_t\bar{w}_t\|^2 \nonumber \\
  & \quad + \lambda^2\|B^{-1}_t(v^m_t-\bar{v}_t)\|^2 + \lambda^2\|B^{-1}_t\bar{v}_t\|^2\big) + \frac{2c_1^2\eta^3_t\sigma^2}{M} \nonumber \\
 & \quad  -34\eta_t \mathbb{E}\|\bar{w}_t - \bar{\nabla} f(x_t,y_t) - R_t\|^2 + \frac{8L^2_K}{M^2}\eta_t\sum_{m=1}^M \Big(\gamma^2\|A^{-1}_t(w^m_t-\bar{w}_t)\|^2 + \gamma^2\|A^{-1}_t\bar{w}_t\|^2 \nonumber \\
  & \quad + \lambda^2\|B^{-1}_t(v^m_t-\bar{v}_t)\|^2 + \lambda^2\|B^{-1}_t\bar{v}_t\|^2\Big) + \frac{2c_2^2\eta_t^3\sigma^2}{M} \bigg),
 \end{align}
where the above inequality holds by the above inequalities \eqref{eq:W1}, \eqref{eq:W2}, \eqref{eq:W3} and \eqref{eq:W4}.

Let $s_t=q\lfloor t/q\rfloor+1$, summing the above inequality \eqref{eq:W5} over $t=s_t$ to $s_t+q-1$,
we have
 \begin{align} \label{eq:W6}
 & \sum_{t=s_t}^{s_t+q-1}\big( \Omega_{t+1} - \Omega_t \big) \nonumber \\
 & \leq \sum_{t=s_t}^{s_t+q-1}\bigg( \frac{8\bar{L}^2\gamma\eta_t}{\rho}\|\bar{y}_t - y^*(\bar{x}_t)\|^2 + \frac{8\gamma\eta_t}{\rho}\|\bar{w}_t -\bar{\nabla} f(x_t,y_t) - R_t\|^2 + \frac{8\gamma\eta_t}{\rho}\|R_t\|^2
 - \frac{\rho\eta_t\gamma}{2}\|A^{-1}_t\bar{w}_t\|^2 \bigg) \nonumber \\
 & \quad + \frac{\theta^2\gamma\rho\hat{L}^2}{36M(L^2_K+L^2_g)}\sum_{t=s_t}^{s_t+q-1}\eta_t\sum_{m=1}^M \Big(\mathbb{E}\|A^{-1}_t(w^m_t - \bar{w}_t)\|^2 +
    \mathbb{E}\|B^{-1}_t(v^m_t - \bar{v}_t)\|^2\Big) \nonumber \\
 & \quad + \frac{40\bar{L}^2\gamma}{\lambda\mu\rho} \bigg( \sum_{t=s_t}^{s_t+q-1}\big(-\frac{\eta_t\mu\lambda}{4}\|\bar{y}_t -y^*(\bar{x}_t)\|^2 -\frac{3b_t\eta_t\lambda^2}{4} \|B^{-1}_t\bar{v}_t\|^2
    + \frac{25\eta_t\lambda}{3\mu}\|\bar{\nabla}_y g(x_t,y_t)-\bar{v}_t\|^2 + \frac{25\kappa^2\eta_tb^2_t\gamma^2}{6\mu\lambda}\|A^{-1}_t\bar{w}_t\|^2 \big)\nonumber \\
  & \quad + \frac{25\theta^2\rho^2\lambda L^2_g}{216\mu M(L^2_K+L^2_g)} \sum_{t=s_t}^{s_t+q-1}\eta_t\sum_{m=1}^M \big(\mathbb{E}\|A^{-1}_t(w^m_t - \bar{w}_t)\|^2 +
    \mathbb{E}\|B^{-1}_t(v^m_t - \bar{v}_t)\|^2\big) \bigg)  \nonumber \\
 & \quad + \frac{\gamma}{\rho} \bigg( \sum_{t=s_t}^{s_t+q-1}\big( -\frac{1000\bar{L}^2\eta_t}{3\mu^2} \mathbb{E}\|\bar{\nabla}_y g(x_t,y_t) - \bar{v}_t\|^2 + \frac{8L^2_g\gamma^2}{M}\eta_t\|A^{-1}_t\bar{w}_t\|^2
  + \frac{8L^2_g\lambda^2}{M}\eta_t\|B^{-1}_t\bar{v}_t\|^2 + \frac{2c_1^2\eta^3_t\sigma^2}{M} \big)  \nonumber \\
  & \quad + \frac{8(L^2_K+L^2_g)\lambda^2}{M^2}\sum_{t=s_t}^{s_t+q-1}\eta_t\sum_{m=1}^M \big(\|A^{-1}_t(w^m_t-\bar{w}_t)\|^2 + \|B^{-1}_t(v^m_t-\bar{v}_t)\|^2\big)  \nonumber \\
 & \quad + \sum_{t=s_t}^{s_t+q-1}\big( -34\eta_t \mathbb{E}\|\bar{w}_t - \bar{\nabla} f(x_t,y_t) - R_t\|^2 + \frac{8L^2_K\gamma^2}{M}\eta_t\|A^{-1}_t\bar{w}_t\|^2
 + \frac{8L^2_K\lambda^2}{M}\eta_t\|B^{-1}_t\bar{v}_t\|^2 + \frac{2c_2^2\eta_t^3\sigma^2}{M} \big) \bigg) \nonumber \\
 & \leq \sum_{t=s_t}^{s_t+q-1}\bigg( \frac{8\bar{L}^2\gamma\eta_t}{\rho}\|\bar{y}_t - y^*(\bar{x}_t)\|^2 + \frac{16\gamma\eta_t}{\rho}\|\bar{w}_t - \hat{\nabla} f(\bar{x}_t,\bar{y}_t) - R_t\|^2+\frac{16(q-1)\hat{L}^2\gamma\eta_t}{M\rho}\bigg(\sum_{l=s_t}^{t-1} \gamma^2\eta_l^2 \sum_{m = 1}^M \mathbb{E}\|A^{-1}_l(w^m_l - \bar{w}_l)\|^2  \nonumber \\
 & \quad + \sum_{l=s_t}^{t-1} \lambda^2\eta_l^2 \sum_{m = 1}^M \mathbb{E}\|B^{-1}_l(v^m_l - \bar{v}_l)\|^2 \bigg) + \frac{8\gamma\eta_t}{\rho}\|R_t\|^2
 - \frac{\rho\eta_t\gamma}{2}\|A^{-1}_t\bar{w}_t\|^2 \bigg) \nonumber \\
 & \quad + \frac{\theta^2\gamma\rho\hat{L}^2}{36M(L^2_K+L^2_g)}\sum_{t=s_t}^{s_t+q-1}\eta_t\sum_{m=1}^M \big(\mathbb{E}\|A^{-1}_t(w^m_t - \bar{w}_t)\|^2 +
    \mathbb{E}\|B^{-1}_t(v^m_t - \bar{v}_t)\|^2\big) \nonumber \\
 & \quad + \frac{40\bar{L}^2\gamma}{\lambda\mu\rho} \bigg( \sum_{t=s_t}^{s_t+q-1}\big(-\frac{\eta_t\mu\lambda}{4}\|\bar{y}_t -y^*(\bar{x}_t)\|^2 -\frac{3\eta_tb_t\lambda^2}{4} \|B^{-1}_t\bar{v}_t\|^2
    + \frac{25\eta_t\lambda}{3\mu}\|\bar{\nabla}_y g(x_t,y_t)-\bar{v}_t\|^2 + \frac{25\kappa^2\eta_tb^2_t\gamma^2}{6\mu\lambda}\|A^{-1}_t\bar{w}_t\|^2 \big)\nonumber \\
 & \quad + \frac{25\theta^2\rho^2\lambda L^2_g}{216\mu M(L^2_K+L^2_g)} \sum_{t=s_t}^{s_t+q-1}\eta_t\sum_{m=1}^M \big(\mathbb{E}\|A^{-1}_t(w^m_t - \bar{w}_t)\|^2 + \mathbb{E}\|B^{-1}_t(v^m_t - \bar{v}_t)\|^2\big) \bigg)  \nonumber \\
 & \quad + \frac{\gamma}{\rho} \Bigg( \sum_{t=s_t}^{s_t+q-1}\big( -\frac{1000\bar{L}^2\eta_t}{3\mu^2} \mathbb{E}\|\bar{\nabla}_y g(x_t,y_t) - \bar{v}_t\|^2 + \frac{8L^2_g\gamma^2}{M}\eta_t\|A^{-1}_t\bar{w}_t\|^2
  + \frac{8L^2_g\lambda^2}{M}\eta_t\|B^{-1}_t\bar{v}_t\|^2 + \frac{2c_1^2\eta^3_t\sigma^2}{M} \big)  \nonumber \\
  & \quad + \frac{8(L^2_K+L^2_g)\lambda^2}{M^2}\sum_{t=s_t}^{s_t+q-1}\eta_t\sum_{m=1}^M \big(\|A^{-1}_t(w^m_t-\bar{w}_t)\|^2 + \|B^{-1}_t(v^m_t-\bar{v}_t)\|^2\big)  \nonumber \\
 & \quad + \sum_{t=s_t}^{s_t+q-1}\bigg( -17\eta_t\|\bar{w}_t -\hat{\nabla} f(\bar{x}_t,\bar{y}_t) - R_t\|^2 + \frac{34(q-1)\hat{L}^2}{M}\eta_t\big(\sum_{l= s_t}^{t-1} \gamma^2\eta_l^2 \sum_{m = 1}^M \mathbb{E}\|A^{-1}_l(w^m_l - \bar{w}_l)\|^2 \nonumber \\
 & \quad + \sum_{l=s_t}^{t-1} \lambda^2\eta_l^2 \sum_{m = 1}^M \mathbb{E}\|B^{-1}_l(v^m_l - \bar{v}_l)\|^2 \big) + \frac{8L^2_K\gamma^2}{M}\eta_t\|A^{-1}_t\bar{w}_t\|^2
 + \frac{8L^2_K\lambda^2}{M}\eta_t\|B^{-1}_t\bar{v}_t\|^2 + \frac{2c_2^2\eta_t^3\sigma^2}{M} \bigg) \Bigg),
 \end{align}
where the first inequality is due to $\eta_t \leq \frac{\theta\rho}{12\lambda q\sqrt{L^2_K+L^2_g}} \ (0 < \theta \leq 1)$ for all $t\geq 1$, $b_t\geq \rho$ and $\lambda\geq \gamma$,
and the last inequality holds by the following inequalities \eqref{eq:W7} and \eqref{eq:W8}.

\begin{align} \label{eq:W7}
 \|\bar{w}_t -\bar{\nabla} f(x_t,y_t) - R_t\|^2 & \leq 2\|\bar{w}_t - \hat{\nabla} f(\bar{x}_t,\bar{y}_t) - R_t\|^2 + 2\|\hat{\nabla} f(\bar{x}_t,\bar{y}_t)-\bar{\nabla} f(x_t,y_t)\|^2 \nonumber \\
 & = 2\|\bar{w}_t - \hat{\nabla} f(\bar{x}_t,\bar{y}_t) - R_t\|^2 + 2\|\frac{1}{M}\sum_{m=1}^M\big(\hat{\nabla} f^m(\bar{x}_t,\bar{y}_t)-\hat{\nabla} f(x^m_t,y^m_t)\big)\|^2 \nonumber \\
 & \leq 2\|\bar{w}_t - \hat{\nabla} f(\bar{x}_t,\bar{y}_t) - R_t\|^2 + 2\frac{1}{M}\sum_{m=1}^M\hat{L}^2\big(\|\bar{x}_t - x^m_t\|^2 + \|\bar{y}_t - y^m_t\|^2 \big) \nonumber \\
 & \leq 2\|\bar{w}_t - \hat{\nabla} f(\bar{x}_t,\bar{y}_t) - R_t\|^2 +\frac{2(q-1)\hat{L}^2}{M}\bigg(\sum_{l = s_t+1}^{t-1} \gamma^2\eta_l^2 \sum_{m=1}^M \mathbb{E}\|A^{-1}_l(w^m_l - \bar{w}_l)\|^2 \nonumber \\
 & \quad + \sum_{l = s_t+1}^{t-1} \lambda^2\eta_l^2 \sum_{m=1}^M \mathbb{E}\|B^{-1}_l(v^m_l - \bar{v}_l)\|^2 \bigg),
\end{align}
where the first inequality holds by Lemma \ref{lem:A8}, and the last inequality is due to Lemma \ref{lem:A11}.

Similarly, we can obtain
\begin{align}
 \|\bar{w}_t -\hat{\nabla} f(\bar{x}_t,\bar{y}_t) - R_t\|^2 & \leq 2\|\bar{w}_t - \bar{\nabla} f(x_t,y_t) - R_t\|^2 + 2\|\hat{\nabla} f(\bar{x}_t,\bar{y}_t)-\bar{\nabla} f(x_t,y_t)\|^2 \nonumber \\
 & \leq 2\|\bar{w}_t - \bar{\nabla} f(x_t,y_t) - R_t\|^2 +\frac{2(q-1)\hat{L}^2}{M}\bigg(\sum_{l = s_t+1}^{t-1} \gamma^2\eta_l^2 \sum_{m=1}^M \mathbb{E}\|A^{-1}_l(w^m_l - \bar{w}_l)\|^2 \nonumber \\
 & \quad + \sum_{l = s_t+1}^{t-1} \lambda^2\eta_l^2 \sum_{m=1}^M \mathbb{E}\|B^{-1}_l(v^m_l - \bar{v}_l)\|^2 \bigg).
\end{align}
Based on the above inequality, then we have
\begin{align} \label{eq:W8}
 -\|\bar{w}_t - \bar{\nabla} f(x_t,y_t) - R_t\|^2
 & \leq -\frac{1}{2}\|\bar{w}_t -\hat{\nabla} f(\bar{x}_t,\bar{y}_t) - R_t\|^2 + \frac{(q-1)\hat{L}^2}{M}\bigg(\sum_{l = s_t+1}^{t-1} \gamma^2\eta_l^2 \sum_{m=1}^M \mathbb{E}\|A^{-1}_l(w^m_l - \bar{w}_l)\|^2 \nonumber \\
 & \quad + \sum_{l = s_t+1}^{t-1} \lambda^2\eta_l^2 \sum_{m=1}^M \mathbb{E}\|B^{-1}_l(v^m_l - \bar{v}_l)\|^2 \bigg).
\end{align}

Since $\eta_t \leq \frac{\theta\rho}{12\lambda q\sqrt{M}\sqrt{L^2_K+L^2_g}} \ (0<\theta\leq 1)$ for all $t\geq 1$, and $\rho \leq b_t \leq \hat{b}$, and $\lambda\geq \gamma$, we have
\begin{align} \label{eq:W9}
 & \sum_{t=s_t}^{s_t+q-1}\big( \Omega_{t+1} - \Omega_t \big) \nonumber \\
 & \leq \sum_{t=s_t}^{s_t+q-1}\bigg( \frac{8\bar{L}^2\gamma\eta_t}{\rho}\|\bar{y}_t - y^*(\bar{x}_t)\|^2 + \frac{16\gamma\eta_t}{\rho}\|\bar{w}_t - \hat{\nabla} f(\bar{x}_t,\bar{y}_t) - R_t\|^2 + \frac{8\gamma\eta_t}{\rho}\|R_t\|^2 - \frac{\rho\eta_t\gamma}{2}\|A^{-1}_t\bar{w}_t\|^2 \bigg) \nonumber \\
 & \quad + \frac{5\theta^2\gamma\rho\hat{L}^2}{36M^2(L^2_K+L^2_g)}\sum_{t=s_t}^{s_t+q-1}\eta_t\sum_{m=1}^M \big(\mathbb{E}\|A^{-1}_t(w^m_t - \bar{w}_t)\|^2 +
    \mathbb{E}\|B^{-1}_t(v^m_t - \bar{v}_t)\|^2\big) \nonumber \\
 & \quad + \frac{40\bar{L}^2\gamma}{\lambda\mu\rho} \bigg( \sum_{t=s_t}^{s_t+q-1}\big(-\frac{\eta_t\mu\lambda}{4}\|\bar{y}_t -y^*(\bar{x}_t)\|^2 -\frac{3\eta_tb_t\lambda^2}{4} \|B^{-1}_t\bar{v}_t\|^2
    + \frac{25\eta_t\lambda}{3\mu}\|\bar{\nabla}_y g(x_t,y_t)-\bar{v}_t\|^2 + \frac{25\kappa^2\eta_tb^2_t\gamma^2}{6\mu\lambda}\|A^{-1}_t\bar{w}_t\|^2 \big)\nonumber \\
  & \quad + \frac{25\theta^2\rho^2\lambda L^2_g}{216\mu M^2(L^2_K+L^2_g)} \sum_{t=s_t}^{s_t+q-1}\eta_t\sum_{m=1}^M \big(\mathbb{E}\|A^{-1}_t(w^m_t - \bar{w}_t)\|^2 +
    \mathbb{E}\|B^{-1}_t(v^m_t - \bar{v}_t)\|^2\big) \bigg)  \nonumber \\
 & \quad + \frac{\gamma}{\rho} \Bigg( \sum_{t=s_t}^{s_t+q-1}\big( -\frac{125\bar{L}^2\eta_t}{6\mu^2} \mathbb{E}\|\bar{\nabla}_y g(x_t,y_t) - \bar{v}_t\|^2 + \frac{8L^2_g\gamma^2}{M}\eta_t\|A^{-1}_t\bar{w}_t\|^2
  + \frac{8L^2_g\lambda^2}{M}\eta_t\|B^{-1}_t\bar{v}_t\|^2 + \frac{2c_1^2\eta^3_t\sigma^2}{M} \big)  \nonumber \\
  & \quad + \frac{8(L^2_K+L^2_g)\lambda^2}{M^2}\sum_{t=s_t}^{s_t+q-1}\eta_t\sum_{m=1}^M \big(\|A^{-1}_t(w^m_t-\bar{w}_t)\|^2 + \|B^{-1}_t(v^m_t-\bar{v}_t)\|^2\big)  \nonumber \\
 & \quad -17\sum_{t=s_t}^{s_t+q-1}\eta_t\|\bar{w}_t -\hat{\nabla} f(\bar{x}_t,\bar{y}_t) - R_t\|^2 + \frac{17\theta^2\rho^2\hat{L}^2}{72M^2(L^2_K+L^2_g)}\sum_{t=s_t}^{s_t+q-1}\eta_t \sum_{m = 1}^M
 \big( \mathbb{E}\|A^{-1}_t(w^m_t - \bar{w}_t)\|^2 + \mathbb{E}\|B^{-1}_t(v^m_t - \bar{v}_t)\|^2 \big)  \nonumber \\
 & \quad  + \sum_{t=s_t}^{s_t+q-1}\big( \frac{8L^2_K\gamma^2}{M}\eta_t\|A^{-1}_t\bar{w}_t\|^2
 + \frac{8L^2_K\lambda^2}{M}\eta_t\|B^{-1}_t\bar{v}_t\|^2 + \frac{2c_2^2\eta_t^3\sigma^2}{M} \big) \Bigg) \nonumber \\
 & \leq -\frac{\gamma}{\rho}\sum_{t=s_t}^{s_t+q-1}\eta_t\|\bar{w}_t -\hat{\nabla} f(\bar{x}_t,\bar{y}_t) - R_t\|^2 - \frac{2\bar{L}^2\gamma}{\rho}\sum_{t=s_t}^{s_t+q-1}\eta_t\|\bar{y}_t - y^*(\bar{x}_t)\|^2
 - \frac{\rho\gamma}{2}\sum_{t=s_t}^{s_t+q-1}\eta_t\|A^{-1}_t\bar{w}_t\|^2 \nonumber \\
 & \quad - \frac{30\bar{L}^2\lambda\gamma}{\mu} \sum_{t=s_t}^{s_t+q-1} \eta_t\|B^{-1}_t\bar{v}_t\|^2 + \frac{8\gamma \lambda^2(L^2_g+L^2_K)}{M\rho}\sum_{t=s_t}^{s_t+q-1}\eta_t\|B^{-1}_t\bar{v}_t\|^2
 + \sum_{t=s_t}^{s_t+q-1}\frac{8\gamma\eta_t}{\rho}\|R_t\|^2 \nonumber \\
 & \quad + \Big(\frac{500\bar{L}^2\kappa^2\gamma^3\hat{b}^2}{3\mu^2\lambda^2\rho} + \frac{8\gamma^3(L^2_g+L^2_K)}{M\rho}\Big)\sum_{t=s_t}^{s_t+q-1}\eta_t\|A^{-1}_t\bar{w}_t\|^2
 + \frac{2(c_1^2+c_2^2)\gamma\sigma^2}{M\rho}\sum_{t=s_t}^{s_t+q-1} \eta^3_t\nonumber \\
 & \quad + \frac{\Gamma}{M^2}\Big(\frac{8M}{15}\sum_{t=s_t}^{s_t+q-1}\eta_t\mathbb{E}\big(\tau^2\|A^{-1}_t\bar{w}_t\|^2+\|B^{-1}_t\bar{v}_t\|^2\big)
   + \frac{4M\hat{c}^2}{15\lambda^2(L^2_K+L^2_g)}
 \sum_{t=s_t}^{s_t+q-1} \eta^3_t \Big),
 \end{align}
where the last inequality holds by $\Gamma= \frac{5\theta^2\hat{L}^2\gamma\rho}{36(L^2_K+L^2_g)} + \frac{125\theta^2\bar{L}^2 L^2_g\gamma \rho}{27\mu^2(L^2_K+L^2_g)} + \frac{8(L^2_K+L^2_g)\lambda^2\gamma}{\rho} + \frac{17\theta^2\hat{L}^2\rho\gamma }{72(L^2_K+L^2_g)}$ and Lemma \ref{lem:A12}.

\begin{align} \label{eq:W90}
 & \sum_{t=s_t}^{s_t+q-1}\big( \Omega_{t+1} - \Omega_t \big) \nonumber \\
 & \leq -\frac{\gamma}{\rho}\sum_{t=s_t}^{s_t+q-1}\eta_t\|\bar{w}_t -\hat{\nabla} f(\bar{x}_t,\bar{y}_t) - R_t\|^2 - \frac{2\bar{L}^2\gamma}{\rho}\sum_{t=s_t}^{s_t+q-1}\eta_t\|\bar{y}_t - y^*(\bar{x}_t)\|^2
 - \frac{\rho\gamma}{4}\sum_{t=s_t}^{s_t+q-1}\eta_t\|A^{-1}_t\bar{w}_t\|^2 \nonumber \\
 & \quad - \Big(\frac{30\bar{L}^2\lambda\gamma}{\mu} - \frac{8\gamma \lambda^2(L^2_g+L^2_K)}{M\rho} - \frac{8\Gamma}{15M} \Big)\sum_{t=s_t}^{s_t+q-1}\eta_t\|B^{-1}_t\bar{v}_t\|^2
 + \sum_{t=s_t}^{s_t+q-1}\frac{8\gamma\eta_t}{\rho}\|R_t\|^2 \nonumber \\
 & \quad - \Big(\frac{\rho\gamma}{4} - \frac{500\bar{L}^2\kappa^2\gamma^3\hat{b}^2}{3\mu^2\lambda^2\rho} - \frac{8\gamma^3(L^2_g+L^2_K)}{M\rho} - \frac{8\Gamma\tau^2}{15M}\Big)\sum_{t=s_t}^{s_t+q-1}\eta_t\|A^{-1}_t\bar{w}_t\|^2 \nonumber \\
 & \quad  + \frac{2(c_1^2+c_2^2)\gamma\sigma^2}{M\rho}\sum_{t=s_t}^{s_t+q-1} \eta^3_t + \frac{4\Gamma\hat{c}^2}{15M\lambda^2(L^2_K+L^2_g)}
 \sum_{t=s_t}^{s_t+q-1} \eta^3_t  \nonumber \\
 & \leq -\frac{\gamma}{\rho}\sum_{t=s_t}^{s_t+q-1}\eta_t\|\bar{w}_t -\hat{\nabla} f(\bar{x}_t,\bar{y}_t) - R_t\|^2 - \frac{2\bar{L}^2\gamma}{\rho}\sum_{t=s_t}^{s_t+q-1}\eta_t\|\bar{y}_t - y^*(\bar{x}_t)\|^2
 - \frac{\rho\gamma}{4}\sum_{t=s_t}^{s_t+q-1}\eta_t\|A^{-1}_t\bar{w}_t\|^2 \nonumber \\
 & \quad + \frac{2(c_1^2+c_2^2)\gamma\sigma^2}{M\rho}\sum_{t=s_t}^{s_t+q-1} \eta^3_t + \frac{4\hat{c}^2\Gamma }{15M\lambda^2(L^2_K+L^2_g)}
 \sum_{t=s_t}^{s_t+q-1} \eta^3_t + \sum_{t=s_t}^{s_t+q-1}\frac{8\gamma\eta_t}{\rho}\|R_t\|^2,
 \end{align}
where the last inequality holds by the following inequalities \eqref{eq:W91} and \eqref{eq:W92}.

Due to $\lambda \leq \frac{225M\rho\bar{L}^2}{184\mu(L^2_K+L^2_g)}$, we can obtain
\begin{align}
\frac{15\bar{L}^2\lambda\gamma}{\mu} \geq \frac{8\gamma \lambda^2(L^2_g+L^2_K)}{M\rho} + \frac{64\gamma \lambda^2(L^2_g+L^2_K)}{15M\rho}.
\end{align}
Meanwhile, due to $\theta \leq 9\bar{L}\sqrt{\frac{75\lambda(L^2_K+L^2_g)M\mu}{\rho\big(30\hat{L}^2\mu^2
 +1000\bar{L}^2L^2_g + 52\hat{L}^2\mu^2\big)}}$, we can obtain
\begin{align}
\frac{15\bar{L}^2\lambda\gamma}{\mu} \geq \frac{2\theta^2\gamma\rho\hat{L}^2}{27M(L^2_g+L^2_K)} + \frac{200\theta^2\bar{L}^2 L^2_g\rho\gamma}{81M\mu^2(L^2_g+L^2_K)}
+ \frac{17\theta^2\hat{L}^2\rho\gamma}{135M(L^2_g+L^2_K)}.
\end{align}
Thus, we have
\begin{align} \label{eq:W91}
 \frac{30\bar{L}^2\lambda\gamma}{\mu} & \geq \frac{8\gamma \lambda^2(L^2_g+L^2_K)}{M\rho} + \frac{8\Gamma}{15M}  \\
 & = \frac{8\gamma \lambda^2(L^2_g+L^2_K)}{M\rho} + \frac{64\gamma \lambda^2(L^2_g+L^2_K)}{15M\rho} + \frac{2\theta^2\gamma\rho\hat{L}^2}{27M(L^2_g+L^2_K)}
 + \frac{200\theta^2\bar{L}^2 L^2_g\rho\gamma}{81M\mu^2(L^2_g+L^2_K)} \nonumber \\
 &\quad + \frac{17\theta^2\hat{L}^2\rho\gamma}{135M(L^2_g+L^2_K)}. \nonumber
\end{align}
Due to $\gamma \leq \frac{\rho}{8}\sqrt{\frac{1}{(125\bar{L}^2\kappa^2\hat{b}^2)/(6\mu^2\lambda^2)+(L^2_g+L^2_K)/M}}$, we can obtain
\begin{align}
 \frac{\rho\gamma}{8} \geq \frac{500\bar{L}^2\kappa^2\gamma^3\hat{b}^2}{3\mu^2\lambda^2\rho} + \frac{8\gamma^3(L^2_g+L^2_K)}{M\rho}.
\end{align}
Meanwhile, due to $\tau\leq \frac{1}{8}\sqrt{\frac{15M\rho\gamma}{\Gamma}}$, we have $\frac{\rho\gamma}{8} \geq \frac{8\Gamma\tau^2}{15M}$.
Thus, we have
\begin{align} \label{eq:W92}
 \frac{\rho\gamma}{4} \geq \frac{500\bar{L}^2\kappa^2\gamma^3\hat{b}^2}{3\mu^2\lambda^2\rho} + \frac{8\gamma^3(L^2_g+L^2_K)}{M\rho} + \frac{8\Gamma\tau^2}{15M}.
\end{align}

Since $\|\bar{w}_t -\hat{\nabla} f(\bar{x}_t,\bar{y}_t)\|^2 \leq 2\|\bar{w}_t -\hat{\nabla} f(\bar{x}_t,\bar{y}_t)-R_t\|^2 + 2\|R_t\|^2$, we have
\begin{align} \label{eq:W10}
 -\|\bar{w}_t -\hat{\nabla} f(\bar{x}_t,\bar{y}_t) - R_t\|^2 \leq -\frac{1}{2}\|\bar{w}_t -\hat{\nabla} f(\bar{x}_t,\bar{y}_t)\|^2 + \|R_t\|^2.
\end{align}
Plugging the above inequalities \eqref{eq:W10} into \eqref{eq:W9}, we can obtain
\begin{align} \label{eq:W11}
 & \sum_{t=s_t}^{s_t+q-1}\big( \Omega_{t+1} - \Omega_t \big) \nonumber \\
 & \leq -\frac{\gamma}{2\rho}\sum_{t=s_t}^{s_t+q-1}\eta_t\|\bar{w}_t -\hat{\nabla} f(\bar{x}_t,\bar{y}_t)\|^2 - \frac{2\bar{L}^2\gamma}{\rho}\sum_{t=s_t}^{s_t+q-1}\eta_t\|\bar{y}_t - y^*(\bar{x}_t)\|^2
 - \frac{\rho\gamma}{4}\sum_{t=s_t}^{s_t+q-1}\eta_t\|A^{-1}_t\bar{w}_t\|^2 \nonumber \\
 & \quad + \frac{2(c_1^2+c_2^2)\gamma\sigma^2}{M\rho}\sum_{t=s_t}^{s_t+q-1} \eta^3_t + \frac{4\hat{c}^2\Gamma }{15M\lambda^2(L^2_K+L^2_g)}
 \sum_{t=s_t}^{s_t+q-1} \eta^3_t + \sum_{t=s_t}^{s_t+q-1}\frac{9\gamma\eta_t}{\rho}\|R_t\|^2.
 \end{align}
Summing the above inequality \ref{eq:W11} from $t=1$ to $T$, then we have
\begin{align} \label{eq:W12}
 & \sum_{t=1}^{T}\big( \Omega_{t+1} - \Omega_t \big) \nonumber \\
 & \leq -\frac{\gamma}{2\rho}\sum_{t=1}^{T}\eta_t\|\bar{w}_t -\hat{\nabla} f(\bar{x}_t,\bar{y}_t)\|^2 - \frac{2\bar{L}^2\gamma}{\rho}\sum_{t=1}^{T}\eta_t\|\bar{y}_t - y^*(\bar{x}_t)\|^2
 - \frac{\rho\gamma}{4}\sum_{t=1}^{T}\eta_t\|A^{-1}_t\bar{w}_t\|^2 \nonumber \\
 & \quad + \frac{2(c_1^2+c_2^2)\gamma\sigma^2}{M\rho}\sum_{t=1}^{T} \eta^3_t + \frac{4\hat{c}^2\Gamma }{15M\lambda^2(L^2_K+L^2_g)}
 \sum_{t=1}^{T} \eta^3_t + \sum_{t=1}^{T}\frac{9\gamma\eta_t}{\rho}\|R_t\|^2.
 \end{align}

Let $\Delta_0 = \|\bar{y}_1-y^*(\bar{x}_1)\|^2$. Since $v^m_1 = \frac{1}{q}\sum_{i=1}^q\nabla_y g^m(x^m_1,y^m_1;\zeta^{m,i}_1)$, and $w^m_1 = \frac{1}{q}\sum_{i=1}^q\hat{\nabla}f^m(x^m_1,y^m_1;\bar{\xi}^{m,i}_1)$ for any $m\in [M]$, we have
\begin{align} \label{eq:W13}
 \Omega_1 &= \mathbb{E}\Big[ F(\bar{x}_1) + \frac{40b_1\bar{L}^2\gamma}{\lambda\mu\rho}\|\bar{y}_1-y^*(\bar{x}_1)\|^2 + \frac{\gamma}{\rho\eta_{0}} \big(\|\bar{v}_1 - \bar{\nabla}_y g(x_1,y_1)\|^2 + \|\bar{w}_1 - \bar{\nabla} f(x_1,y_1)-R_1\|^2 \big) \Big] \nonumber \\
 & \leq F(\bar{x}_1) + \frac{40b_1\bar{L}^2\gamma\Delta_0}{\lambda\mu\rho} + \frac{2\gamma\sigma^2}{Mq\rho\eta_0},
\end{align}
where the last inequality holds by Assumption \ref{ass:2}.

Since $\eta_t=\frac{kM^{1/3}}{(n+t)^{1/3}}$ is decreasing, i.e., $\eta_T^{-1} \geq \eta_t^{-1}$ for any $0\leq t\leq T$, we have
 \begin{align}  \label{eq:W14}
 & \frac{1}{T} \sum_{t=1}^T \mathbb{E}\Big[ \|A^{-1}_t\bar{w}_t\|^2 + \frac{2}{\rho^2}\|\bar{w}_t -\hat{\nabla} f(\bar{x}_t,\bar{y}_t)\|^2 + \frac{8\bar{L}^2}{\rho^2}\|\bar{y}_t - y^*(\bar{x}_t)\|^2 \Big]  \\
 & \leq  \frac{4}{T\rho\gamma\eta_T} \sum_{t=1}^T\big(\Omega_t - \Omega_{t+1}\big) + \frac{8(c^2_1+c^2_2)\sigma^2}{MT\rho^2\eta_T}\sum_{t=1}^T\eta^3_t + \frac{36}{T\rho^2\eta_T}\sum_{t=1}^T\eta_t\|R_t\|^2
 +  \frac{16\hat{c}^2\Gamma }{15M T\eta_T\rho\gamma\lambda^2(L^2_K+L^2_g)}\sum_{t=1}^{T} \eta^3_t\nonumber \\
 & \leq \frac{4}{T\rho\gamma\eta_T} \big( F(\bar{x}_1) + \frac{40b_1\bar{L}^2\gamma\Delta_0}{\lambda\mu\rho} + \frac{2\gamma\sigma^2}{qM\rho\eta_0} - F^* \big) + \Big(\frac{8(c^2_1+c^2_2)\sigma^2}{MT\rho^2\eta_T}
 + \frac{16\hat{c}^2\Gamma }{15MT\eta_T\rho\gamma\lambda^2(L^2_K+L^2_g)}\Big)\sum_{t=1}^T\eta^3_t \nonumber \\
 & \quad + \frac{36}{\eta_T\rho^2T^3}\sum_{t=1}^T\eta_t  \nonumber \\
 & \leq \frac{4}{T\rho\gamma\eta_T} \big( F(\bar{x}_1) + \frac{40b_1\bar{L}^2\gamma\Delta_0}{\lambda\mu\rho} + \frac{2\gamma\sigma^2}{qM\rho\eta_0} - F^* \big) + \Big(\frac{8(c^2_1+c^2_2)\sigma^2}{MT\rho^2\eta_T}
 + \frac{16\hat{c}^2\Gamma }{15MT\eta_T\rho\gamma\lambda^2(L^2_K+L^2_g)}\Big)\int^T_1\frac{Mk^3}{n+t} dt \nonumber \\
 & \quad + \frac{36}{\eta_T\rho^2T^3}\int_{1}^T\frac{kM^{1/3}}{(n+t)^{1/3}} dt\nonumber \\
 & \leq  \frac{4(F(\bar{x}_1) - F^*)}{T\rho\gamma\eta_T} + \frac{160b_1\bar{L}^2\Delta_0}{\lambda\mu\rho^2\eta_TT} + \frac{8\sigma^2}{qM\rho^2\eta_0\eta_TT}
 + \frac{k^3}{T\eta_T}\Big(\frac{8(c^2_1+c^2_2)\sigma^2}{\rho^2}
 + \frac{16\hat{c}^2\Gamma }{15\rho\gamma\lambda^2(L^2_K+L^2_g)}\Big)\ln(n+T) \nonumber \\
 & \quad + \frac{54kM^{1/3}}{T^3\eta_T\rho^2}(n+T)^{2/3} \nonumber \\
 & = \bigg( \frac{4(F(\bar{x}_1) - F^*)}{k\rho\gamma} + \frac{160b_1\bar{L}^2\Delta_0}{k\lambda\mu\rho^2} + \frac{8n^{1/3}\sigma^2}{qM^{4/3}k^2\rho^2} + 8k^2\Big(\frac{(c^2_1+c^2_2)\sigma^2}{\rho^2}
 + \frac{2\hat{c}^2\Gamma }{15\rho\gamma\lambda^2(L^2_K+L^2_g)}\Big)\ln(n+T) \nonumber \\
 & \qquad + \frac{54kM^{1/3}(n+T)^{2/3}}{T^2\rho^2} \bigg)\frac{(n+T)^{1/3}}{M^{1/3}T}, \nonumber
\end{align}
where the second inequality holds by the above inequality \eqref{eq:W13}, Assumption \ref{ass:8} and $\|R_t\|^2\leq \frac{1}{T^2}$ that is due to the following fact:
\begin{align}
 \|R_t\|^2 = \|\frac{1}{M}\sum_{m=1}^MR_t\|^2 \leq \frac{1}{M}\sum_{m=1}^M\|R_t\|^2 \leq \frac{1}{T^2},
\end{align}
where the last inequality holds by the above Lemma \ref{lem:2} and $K=\frac{L_g}{\mu}\log(C_{gxy}C_{fy}T/\mu)$.
Let $G = \frac{4(F(\bar{x}_1) - F^*)}{k\rho\gamma} + \frac{160b_1\bar{L}^2\Delta_0}{k\lambda\mu\rho^2} + \frac{8n^{1/3}\sigma^2}{qM^{4/3}k^2\rho^2} + 8k^2\Big(\frac{(c^2_1+c^2_2)\sigma^2}{\rho^2}
 + \frac{2\hat{c}^2\Gamma }{15\rho\gamma\lambda^2(L^2_K+L^2_g)}\Big)\ln(n+T) + \frac{54kM^{1/3}(n+T)^{2/3}}{T^2\rho^2} $,
we have
\begin{align} \label{eq:W15}
 \frac{1}{T} \sum_{t=1}^T \mathbb{E}\Big[ \|A^{-1}_t\bar{w}_t\|^2 + \frac{2}{\rho^2}\|\bar{w}_t -\hat{\nabla} f(\bar{x}_t,\bar{y}_t)\|^2 + \frac{8\bar{L}^2}{\rho^2}\|\bar{y}_t - y^*(\bar{x}_t)\|^2 \Big]  \leq \frac{G}{M^{1/3}T}(n+T)^{1/3}.
\end{align}

According to Lemma \ref{lem:A4} and $(a+b)^{1/2} \leq a^{1/2}+b^{1/2}$, we have
\begin{align}
\|\bar{w}_t-\nabla F(\bar{x}_t)\| \leq 2\sqrt{2}\bar{L}\|y^*(\bar{x}_t)-\bar{y}_t\| + \sqrt{2}\|\bar{w}_t-\hat{\nabla} f(\bar{x}_t,\bar{y}_t)\|.
\end{align}
Let $\mathcal{G}_t = \|A^{-1}_t\bar{w}_t\|
+ \frac{1}{\rho}\Big( 2\sqrt{2}\bar{L}\|y^*(\bar{x}_t)-\bar{y}_t\| + \sqrt{2}\|\bar{w}_t-\hat{\nabla} f(\bar{x}_t,\bar{y}_t)\| \Big)$, we have
\begin{align}
\mathcal{G}_t & = \|A^{-1}_t\bar{w}_t\|
+ \frac{1}{\rho}\Big( 2\sqrt{2}\bar{L}\|y^*(\bar{x}_t)-\bar{y}_t\| + \sqrt{2}\|\bar{w}_t-\hat{\nabla} f(\bar{x}_t,\bar{y}_t)\| \Big) \nonumber \\
& \geq \|A^{-1}_t\bar{w}_t\|
+ \frac{1}{\rho}\|\bar{w}_t-\nabla F(\bar{x}_t)\| \nonumber \\
& = \frac{1}{\|A_t\|}\|A_t\|\|A_t^{-1}\bar{w}_t\|
+ \frac{1}{\rho}\|\bar{w}_t - \nabla F(\bar{x}_t)\| \nonumber \\
& \geq \frac{1}{\|A_t\|}\|\bar{w}_t\|
+ \frac{1}{\rho}\|\bar{w}_t - \nabla F(\bar{x}_t)\| \nonumber \\
& \mathop{\geq}^{(i)} \frac{1}{\|A_t\|}\|\bar{w}_t\|
+ \frac{1}{\|A_t\|}\|\nabla F(\bar{x}_t)-\bar{w}_t\| \nonumber \\
& \geq \frac{1}{\|A_t\|}\|\nabla F(\bar{x}_t)\|,
\end{align}
where the inequality $(i)$ holds by $\|A_t\| \geq \rho$ for all $t\geq1$ due to Assumption \ref{ass:6}. Then we have
\begin{align}
 \|\nabla F(\bar{x}_t)\| \leq \|A_t\|\mathcal{G}_t.
\end{align}
According to Cauchy-Schwarz inequality, we have
\begin{align} \label{eq:W16}
\frac{1}{T}\sum_{t=1}^T\mathbb{E}\|\nabla F(\bar{x}_t)\| \leq \frac{1}{T}\sum_{t=1}^T\mathbb{E}\big[\mathcal{G}_t \|A_t\|\big] \leq \sqrt{\frac{1}{T}\sum_{t=1}^T\mathbb{E}[\mathcal{G}_t^2]} \sqrt{\frac{1}{T}\sum_{t=1}^T\mathbb{E}\|A_t\|^2}.
\end{align}
According to the above inequality \eqref{eq:W15}, we have
\begin{align} \label{eq:W17}
 \frac{1}{T} \sum_{t=1}^T \mathbb{E}[\mathcal{G}_t^2] & \leq \frac{1}{T} \sum_{t=1}^T \mathbb{E}\Big[ 3\|A^{-1}_t\bar{w}_t\|^2
+ \frac{3}{\rho^2}\Big( 8\bar{L}^2\|y^*(\bar{x}_t)-\bar{y}_t\|^2 + 2\|\bar{w}_t-\hat{\nabla} f(\bar{x}_t,\bar{y}_t)\|^2 \Big) \Big] \nonumber \\
 & \leq \frac{3G}{M^{1/3}T}(n+T)^{1/3}.
\end{align}

By combining the above inequalities \eqref{eq:W16} and \eqref{eq:W17}, we have
\begin{align}
\frac{1}{T}\sum_{t=1}^T\mathbb{E}\|\nabla F(\bar{x}_t)\| & \leq  \sqrt{\frac{1}{T}\sum_{t=1}^T\mathbb{E}[\mathcal{G}_t^2]} \sqrt{\frac{1}{T}\sum_{t=1}^T\mathbb{E}\|A_t\|^2} \nonumber \\
& \leq\Big( \frac{\sqrt{3G}n^{1/6}}{M^{1/6}T^{1/2}} + \frac{\sqrt{3G}}{M^{1/6}T^{1/3}}\Big)\sqrt{\frac{1}{T}\sum_{t=1}^T\mathbb{E}\|A_t\|^2}.
\end{align}

\end{proof}

\end{document}